\documentclass[10pt,twocolumn,letterpaper]{article}

\usepackage{wacv}              

\usepackage{graphicx}
\usepackage{amsmath}
\usepackage{amssymb}
\usepackage{booktabs}

\usepackage{amsmath}
\usepackage{multicol}
\usepackage{multirow}
\usepackage[dvipsnames]{xcolor}
\usepackage{colortbl}
\usepackage{arydshln}
\usepackage[normalem]{ulem}
\usepackage{amssymb}
\usepackage{amsmath}
\usepackage{pifont}
\usepackage{adjustbox}
\usepackage{caption}
\usepackage[symbol]{footmisc}
\usepackage[linesnumbered, ruled, vlined]{algorithm2e}
\usepackage{algpseudocode}
\usepackage{bbm}

\usepackage{amsmath,amsfonts,bm}

\def\1{\bm{1}}

\def\vg{{\bm{g}}}

\def\vv{{\bm{v}}}
\def\vw{{\bm{w}}}

\def\vz{{\bm{z}}}

\DeclareMathAlphabet{\mathsfit}{\encodingdefault}{\sfdefault}{m}{sl}
\SetMathAlphabet{\mathsfit}{bold}{\encodingdefault}{\sfdefault}{bx}{n}
\newcommand{\tens}[1]{\bm{\mathsfit{#1}}}

\def\tI{{\tens{I}}}

\def\sM{{\mathbb{M}}}

\DeclareMathOperator*{\argmin}{arg\,min}

\definecolor{wacvblue}{rgb}{0.21,0.49,0.74}
\usepackage[pagebackref,breaklinks,colorlinks,allcolors=wacvblue]{hyperref}

\def\wacvPaperID{1803} 
\def\confName{WACV}
\def\confYear{2026}

\title{\uline{QCFace}: Image \uline{Q}uality \uline{C}ontrol for boosting \uline{Face} Representation \& Recognition}

\author{Duc-Phuong Doan-Ngo$^{3}$, ~~Thanh-Dang Diep$^{3,*}$, ~~Thanh Nguyen-Duc$^{5}$,\\
~~Thanh-Sach LE$^{2,3,*}$, ~~Nam Thoai$^{1,3,4,*}$\\
$^{1}$High Performance Computing Laboratory and $^{2}$Data Science Laboratory,\\
$^{3}$Faculty of Computer Science and Engineering and\\
$^{4}$Advanced Institute of Interdisciplinary Science and Technology,\\
Ho Chi Minh City University of Technology (HCMUT), VNU-HCM, Ho Chi Minh City, Vietnam.\\
$^{5}$School of Clinical Sciences, Monash University and Department of Imaging, Monash Health,\\
Victoria, Australia.\\
{\tt \small \{dndphuong,dang,ltsach,namthoai\}@hcmut.edu.vn ~~~ thanh.nguyen5@monash.edu}
}

\begin{document}
\maketitle

\footnotetext[1]{Corresponding authors.}
\footnotetext[2]{These terms are defined at Sec. B in Supplementary Material (SM).}

\begin{abstract}
Recognizability, a key perceptual factor in human face processing, strongly affects the performance of face recognition (FR) systems in both verification and identification tasks. Effectively using recognizability to enhance feature representation remains challenging. In deep FR, the loss function plays a crucial role in shaping how features are embedded. However, current methods have two main drawbacks: (i) recognizability is only partially captured through soft margin constraints, resulting in weaker quality representation and lower discrimination, especially for low-quality or ambiguous faces; (ii) mutual overlapping gradients between feature direction and magnitude introduce undesirable interactions during optimization, causing instability and confusion in hypersphere planning, which may result in poor generalization, and entangled representations where recognizability and identity are not cleanly separated. To address these issues, we introduce a hard margin strategy -- Quality Control Face (QCFace), which overcomes the mutual overlapping gradient problem and enables the clear decoupling of recognizability from identity representation. Based on this strategy, a novel hard-margin-based loss function employs a guidance factor for hypersphere planning, simultaneously optimizing for recognition ability and explicit recognizability representation. Extensive experiments confirm that QCFace not only provides robust and quantifiable recognizability encoding but also achieves state-of-the-art performance in both verification and identification benchmarks compared to existing recognizability-based losses. The code is available at \href{https://github.com/hpcc-hcmut/QCFace}{QCFace Repository}.
\end{abstract}

\section{Introduction}
\label{sec:introduction}
Due to the demonstrated superiority of deep learning techniques over traditional image processing methods \cite{ahonen2006face} in computer vision, face recognition (FR) is increasingly supplanting other biometric modalities (\textit{e.g.}, fingerprint, iris, voice) owing to its non-intrusive nature, user convenience, and high reliability \cite{taskiran2020face}. In particular, FR models autonomously learn and extract identity-discriminative features from facial images, enabling accurate individual recognition using only facial data \cite{taskiran2020face}.

For deep-learning-based approaches, four primary aspects can be leveraged to enhance the recognition capability of an FR model: \textit{model architecture} \cite{taigman2014deepface,chen2018mobilefacenets,alansari2023ghostfacenets,wang2020hierarchical,duta2021improved,li2023bionet}, \textit{training strategy} \cite{li2021dynamic,li2021virtual,wang2022efficient,an2022killing,huang2022evaluation,xu2023probabilistic,li2023rethinking,yu2023icd,sevastopolskiy2023boost,chai2023recognizability}, \textit{training dataset} \cite{wu2018light,yang2019feature,hu2019noise,wang2019co,zhang2020global,zhao2020rdcface,al2020automated,qiu2021synface,zhang2021adaptive}, and \textit{objective (loss) function} \cite{wen2019comprehensive}. To evaluate their effectiveness, two criteria are considered: \textit{knowledge accumulation (aka learning capacity)}\footnotemark[2], measuring the model's feature synthesis extraction capability, and \textit{knowledge capacity}\footnotemark[2], assessing data quality and utility. However, the first three aspects cannot simultaneously improve both the criteria, as they lack explicit guidance for hypersphere planning\footnotemark[2], which optimizes identity representation in facial embeddings. Specifically, previous \textit{model architecture} and \textit{training strategy} methods mainly address learning capacity, while \textit{training dataset} methods primarily enhance knowledge capacity.

\begin{figure}[!t]
  \centering 
  \begin{subfigure}{0.9\linewidth}
    \includegraphics[width=\textwidth]{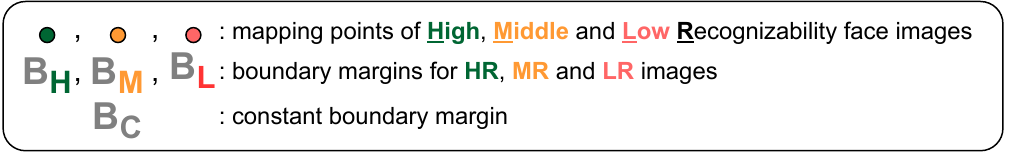}
  \end{subfigure}
  \begin{subfigure}{0.32\linewidth}
    \includegraphics[width=\textwidth]{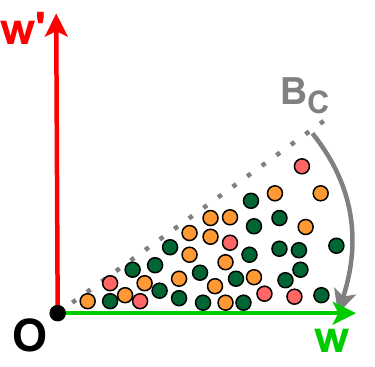}
    \caption{Constant margin}
    \label{fig:mov_other}
  \end{subfigure}
  \begin{subfigure}{0.32\linewidth}
    \includegraphics[width=\textwidth]{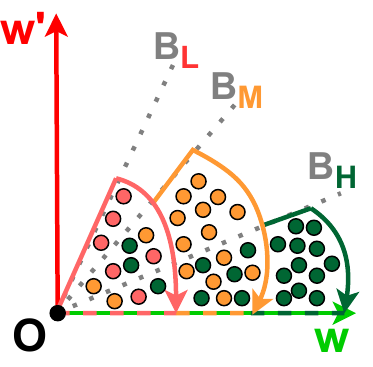}
    \caption{Soft margin}
    \label{fig:mov_mag}
  \end{subfigure}
  \begin{subfigure}{0.32\linewidth}
    \includegraphics[width=\textwidth]{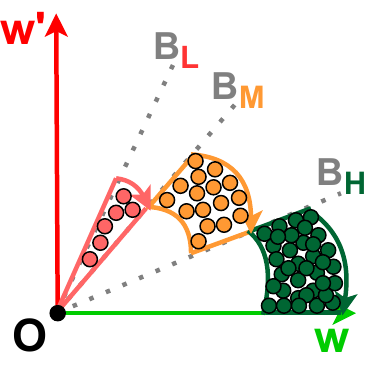}
    \caption{\textbf{\textit{Hard margin}}}
    \label{fig:mov_qc}
  \end{subfigure}
  \caption{Illustration of three types of hypersphere planning solutions for face representation learning. \textit{Constant margin} shows the most lenient constraint in hypersphere planning, causing the least informative feature magnitudes. Although \textit{soft margin} partially exploits the magnitude value, it possesses the entangled representation of recognizability. \textbf{\textit{Hard margin (ours)}} facilitates positive correlation between magnitude value and confidence score to completely mitigate confusion in recognizability representation.}
  \label{fig:motivation}
\end{figure}

In contrast, the \textit{loss function} is a promising approach capable of addressing both \textit{knowledge accumulation} and \textit{knowledge capacity} by explicitly defining mathematical constraints in hypersphere planning and modeling. A well-designed loss function improves feature representational capacity\footnotemark[2] across various backbone architectures, including general classification networks \cite{simonyan2014very, szegedy2015going, he2016deep, szegedy2016rethinking}. Although some studies tailor loss functions to data uncertainty \cite{shi2019probabilistic, chang2020data, shi2020towards} and the unique decision thresholds \cite{zhou2023uniface}, these methods neglect recognizability information from face images in their hypersphere planning to optimize the face representation learning (FRL). Meanwhile, \textit{sample-to-sample loss} and \textit{class center loss}, two primary branches of loss functions, offer flexibility for embedding recognizability constraints. However, \textit{sample-to-sample} methods \cite{chopra2005learning,schroff2015facenet} suffer from  the lack of global observation and computationally intensive pair/triplet mining procedures (\textit{i.e.}, locally optimizing for each pair of samples).
\newline \indent For \textit{class center loss}, boundary margin critically influences hypersphere planning by defining the region for each identity in the training dataset. However, to the best of our knowledge, categorizing margin adjustment has never been taken into account. Therefore, \textit{margin strategy} is provided as a new definition to categorize margin adjustments in prior methods (see \cref{fig:motivation}). Figs. \ref{fig:mov_other} and \ref{fig:mov_mag} depict two existing \textit{margin strategies}. The \textit{constant margin}, with minimal constraints and neglect of recognizability, yields less informative features. Although the \textit{soft margin} partially improves the expressive power of the embedded features by utilizing feature magnitude as a recognizability proxy, confusion in recognizability representation (\textit{i.e.}, overlapping regions in \cref{fig:mov_mag}) persists, as the correlation between recognizability level and confidence score is ignored in FRL.
\newline \indent With this new perspective, we propose \textbf{\textit{hard margin}} to resolve ambiguities in recognizability representation. Specifically, we enforce a consistent positive correlation between confidence score and recognizability level via a novel \textbf{\textit{hard constraint}} in hypersphere planning (see \cref{fig:mov_qc}). We formulate this constraint through our new loss function -- \textbf{\textit{QCFace}}, addressing the limitations in FRL of previous recognizability-based methods (see \cref{subsubsec:hard_margin}). 

\noindent Our key contributions are summarized as follows:
\begin{itemize}
    \item We first explore the strategy of margin adjustment implied in previous loss models to define a novel taxonomy in \textit{class center loss} -- \textit{margin strategy}, based on several lemmas and proofs. We then propose a new \textit{margin strategy} -- \textbf{\textit{hard margin}} to overcome the existed confusion in recognizability representation. (see \cref{subsec:margin_strategy}).
    \item Accordingly, \textbf{\textit{guidance factor}} is first-introduced to model the loss function -- \textbf{\textit{QCFace}} to fully exploit expressive power of embedded features (see \cref{subsec:qcface}).
    \item Our extensive experiments on several benchmark datasets successfully illustrate the effectiveness of \textbf{\textit{QCFace}} in recognizability representation; and show \textbf{\textit{QCFace}} as a new cutting-edge on recognition task (see \cref{sec:experiments}).
    
\end{itemize}

\section{Related Work}
\label{sec:related_works}
Motivated by the \textit{class center loss} (see \cref{sec:introduction}), this paper investigates this category, which targets two primary objectives: \textit{(i)} learning an optimal vector (proxy) representing each identity, and \textit{(ii)} enhancing intra-class compactness and inter-class discrepancy by optimizing the placement of embedded features on the hypersphere. A notable challenge is the reliance on extensive identity sets to construct discriminative hyperspherical spaces, an issue not met by several large-scale public datasets (\eg, \cite{deng2019lightweight,zhu2021webface260m,yi2014learning}). Here, we evaluate strengths and limitations in FRL of prior studies. 

\textit{Class center losses} are broadly categorized based on boundary adjustment methods into three groups: \textit{margin-based softmax}, \textit{mis-classified softmax}, and \textit{magnitude-based losses}. \textit{Margin-based softmax loss:} Typical methods include angular margin (\textit{i.e.}, SphereFace \cite{liu2017sphereface,wen2021sphereface2,liu2022sphereface}), additive margin (\textit{i.e.}, CosFace \cite{wang2018cosface}, AMSoftmax \cite{wang2018additive}), and additive angular margin (\textit{i.e.}, ArcFace \cite{deng2019arcface}), differing in geometric representation (see \cref{sec:prelimiary}). Some approaches (\textit{i.e.}, \cite{saadabadi2023quality, deng2021variational}) employ memory banks to store positive proxies for smoothing proxy optimization, but introduce memory overhead. \textit{Misclassified softmax loss:} To address limited attention on misclassified samples, MV-Softmax \cite{wang2020mis}, CurricularFace \cite{huang2020curricularface}, and UniFace \cite{zhou2023uniface} balance positive and negative sample-to-class impacts for structured embeddings. UniFace additionally sets a unique decision threshold, while UniTSFace \cite{jia2023unitsface} integrates local sample-to-sample similarity. \textit{Magnitude-based loss:} Recognizability significantly influences FRL optimization, a factor overlooked by both previous approaches. MagFace \cite{meng2021magface} and AdaFace \cite{kim2022adaface} use feature magnitude as a recognizability proxy but suffer drawbacks: \textit{(i)} the absence of meaningful feature magnitude in recognizability representation without any constraint for the magnitude value-driven planning, and \textit{(ii)} \textit{the mutual overlapping gradients} that cause confusion in magnitude encoding for recognizability representation.
\newline \indent Besides, TopoFR \cite{dan2024topofr} proposes a more special approach, which leverages topological features by applying persistent homology and emphasizes the learning-hardness samples by proposing structure damage estimation. \cref{tab:comparison_attribute} provides a comparison between previous studies based on five criteria, where \textit{margin strategy} is first-introduced to assess a strategy of margin adjustment in defining the constraint of representative region for each identity (see \cref{subsec:margin_strategy}).

\begin{table}[!t]
\centering
\caption{The comparison between the class center loss methods based on \textbf{B}oundary \textbf{M}argin for \textbf{A}ctual and \textbf{M}is\textbf{c}lassified \textbf{P}roxies (\textbf{BMAP} and \textbf{BMMcP}), \textbf{M}argin \textbf{S}trategy (\textbf{MS}), \textbf{R}ecognizability \textbf{R}epresentation \textbf{L}evel (\textbf{RRL}) and \textbf{M}emory \textbf{C}ost (\textbf{MC}). \textbf{AAa} is the utilization of both additive and additive angular margins. VPLFace and UniTSFace utilize a memory bank mechanism, which results in additional memory overhead. QAFace and TopoFR cache computationally expensive learnable modules, leading to more increased memory consumption. \textbf{\textit{QCFace (ours)}} is designed with \textbf{\textit{hard margin}} strategy to ignore \textit{the mutual overlapping gradient}, leading to more effectiveness in recognizability representation.}
\label{tab:comparison_attribute}
\begin{adjustbox}{width=\columnwidth,center}
\begin{tabular}{l;{1pt/1pt}c;{1pt/1pt}c;{1pt/1pt}c;{1pt/1pt}c;{1pt/1pt}c} 
\hline
\rowcolor[rgb]{0.784,0.784,0.784} \multicolumn{1}{c|}{\textbf{Method}} & \multicolumn{1}{c|}{\textbf{BMAP}} & \multicolumn{1}{c|}{\textbf{BMMcP}} & \multicolumn{1}{c|}{\textbf{MS}}                                   & \multicolumn{1}{c|}{\textbf{RRL}}                           & \textbf{MC}                                       \\ 
\hline\hline
SphereFace \cite{liu2017sphereface}                                                    & Angular                            & \ding{55}                                  & \textit{Constant margin}                                           & \ding{55}                                                          & Low                                                \\ 
\hdashline[1pt/1pt]
CosFace \cite{wang2018cosface}                                                       & Additive                           & \ding{55}                                  & \textit{Constant margin}                                           & \ding{55}                                                          & Low                                                \\ 
\hdashline[1pt/1pt]
ArcFace \cite{deng2019arcface}                                                       & Additive angular                   & \ding{55}                                  & \textit{Constant margin}                                           & \ding{55}                                                          & Low                                                \\ 
\hdashline[1pt/1pt]
MV-Softmax \cite{wang2020mis}                                                    & Additive angular                   & Constant                            & \textit{Constant margin}                                           & \ding{55}                                                          & Low                                                \\ 
\hdashline[1pt/1pt]
CurricularFace \cite{huang2020curricularface}                                                & Additive angular                   & Adaptive                            & \textit{Constant margin}                                           & \ding{55}                                                          & Low                                                \\ 
\hdashline[1pt/1pt]
MagFace \cite{meng2021magface}                                                       & Additive angular                   & \ding{55}                                  & \textit{Soft margin}                                               & Middle                                                      & Low                                                \\ 
\hdashline[1pt/1pt]
VPLFace \cite{deng2021variational}                                                       & Additive angular                   & Adaptive                            & \textit{Constant margin}                                           & \ding{55}                                                          & Middle                                            \\ 
\hdashline[1pt/1pt]
AdaFace \cite{kim2022adaface}                                                       & AAa                                & \ding{55}                                  & \textit{Soft margin}                                               & Low                                                         & Low                                                \\ 
\hdashline[1pt/1pt]
QAFace \cite{saadabadi2023quality}                                                        & Additive angular                   & Adaptive                            & \textit{Constant margin}                                           & \ding{55}                                                          & High                                              \\ 
\hdashline[1pt/1pt]
UniFace \cite{zhou2023uniface}                                                       & AAa                                & Constant                            & \textit{Constant margin}                                           & \ding{55}                                                          & Low                                                \\ 
\hdashline[1pt/1pt]
UniTSFace \cite{jia2023unitsface}                                                    & AAa                                & Constant                            & \textit{Constant margin}                                           & \ding{55}                                                          & Middle                                            \\ 
\hdashline[1pt/1pt]
TopoFR \cite{dan2024topofr}                                                        & Additive angular                   & \ding{55}                                  & \textit{Constant margin}                                           & \ding{55}                                                          & High                                              \\ 
\hline\hline
\textit{\textbf{QCFace (Ours)}}                                        & Additive angular                   & \textit{n/a}                        & \textbf{\textbf{\textit{\textcolor[rgb]{0,0.392,0}{Hard margin}}}} & \textbf{\textbf{\textit{\textcolor[rgb]{0,0.392,0}{High}}}} & \textbf{\textit{\textcolor[rgb]{0,0.392,0}{Low}}}  \\
\hline
\end{tabular}
\end{adjustbox}
\end{table}

\footnotetext[3]{$F$, $N$ and $\mathcal{L}_{reg}$ in previous methods are detailed at Sec. A in SM.}
\section{Preliminaries}
\label{sec:prelimiary}
In this section, we formulate the loss model of the \textit{class center loss} category. Let $\tI_i \in \mathbb{R}^{c \times h \times w}$ be an input face image $i$ where $c$, $h$ and $w$ are channels, height and width, respectively.  $\vz_i=\mathcal{F}(\tI_i)$ is embedded feature extracted by facial backbone $\mathcal{F}: \mathbb{R}^{C \times H \times W} \rightarrow \mathbb{R}^d$ and $d$ is embedded dimension. $\theta_{\vv_1,\vv_2}$ is the angular between two vectors $\vv_1, \vv_2$. Generally, the class center loss can be formulated as
\begin{equation}
    \resizebox{0.48\textwidth}{!}{$
        \begin{array}{l}
            \hspace{0.06cm} \mathcal{L}_i = \underbrace{-log\frac{e^{s.F(\sM, \theta_{\vw_{y_i},\vz_i})}}{e^{s.F(\sM, \theta_{\vw_{y_i},\vz_i})} + \sum_{j \neq y_i}^{C}e^{s.N(t, \theta_{\vw_j,\vz_i})}}}_{{\mathcal{L}_{sm}}_i} + \lambda_g.\underbrace{g(\vz_i)}_{{\mathcal{L}_{reg}}_i}
        \end{array}
    $}
  \label{eq:gen_cen_loss}
\end{equation}
where $\mathcal{L}_{sm}$ is softmax loss with the modulation of positive (with function $F$) and negative (with function $N$) cosine similarities. $\mathcal{L}_{reg}$ is the regularization loss to define the additional constraints in hypersphere planning, and $\lambda_g$ is its contribution adjustment. $y_i$ expresses the actual class of the input, and $\vw_j$ is the proxy of class $j$. $C$ and $s$ are the identity count and scale value accordingly. The adjustment function $F$ of the actual class can be generalized as
\begin{equation}
    \resizebox{0.433\textwidth}{!}{$
        F(\sM, \theta) = \text{cos}(m_1\theta+m_2)-m_3 ~~ \text{where} ~~ \sM = \{m_1, m_2, m_3\}
    $}
    \label{eq:pre_add_margin}
\end{equation}
\noindent where $m_1$, $m_2$ and $m_3$ are three \textit{margin types} with different geometric attributes, called angular, additive angular and additive boundary margin. $m_2$ may either be treated as a constant value, similar to $m_1$ and $m_3$, or defined as a function of $\|\vz_i\|$ (\textit{i.e.}, $m_2(\|\mathbf{z_i}\|)$) \cite{meng2021magface,kim2022adaface}. $F$, $N$ and $\mathcal{L}_{reg}$ all have a direct impact on the effectiveness in hypersphere planning for FRL \cite{zhang2023unifying}\footnotemark[3].

\section{Methodology}
\label{sec:methodology}
In this section, we first introduce \textbf{\textit{hard margin}} strategy for margin adjustment to optimize FRL (see \cref{subsec:margin_strategy}). Next, we propose \textbf{\textit{QCFace}} loss function obeying the \textbf{\textit{hard margin}} strategy to optimize recognizability representation by defining a \textbf{\textit{hard constraint}} (see \cref{subsec:qcface}).

\subsection{Margin Strategy}
\label{subsec:margin_strategy}

\textit{Margin strategy} is our first-defined taxonomy to classify various margin-adjustment strategies for each training sample, utilized to modify the decision boundary in hypersphere planning. There are two main categories of margin strategy: \textit{constant margin} and \textit{soft margin}. To analyze and differentiate between margin strategies, some mathematical interpretations are used. 

First, the derivatives of ${\mathcal{L}_{sm}}_i$ with respect to $\vw_k$ and $\vz_i$ are shown as follows.
\begin{equation}
    \resizebox{0.43\textwidth}{!}{$
        \begin{aligned}
            \frac{\partial {\mathcal{L}_{sm}}_i}{\partial \vw_k} & = s.\left[P^{(i)}_k-\1_\mathrm{k=y_i}\right].\frac{\partial \textit{Fnc}^{(i)}_k}{\partial \text{cos}(\theta_{\vw_k,\vz_i})}.\frac{\partial \text{cos}(\theta_{\vw_k,\vz_i})}{\partial \vw_k} \\
            & := 
            \begin{cases}
                \vg_{ac} & k = y_i \\
                \vg_{mc} & k \neq y_i
            \end{cases}
            ~~ \text{where} ~~ \textit{Fnc}^{(i)}_k = 
            \begin{cases}
                F(\sM, \theta_{\vw_{yi},\vz_i}) & k = y_i \\
                N(t, \theta_{\vw_k,\vz_i}) & k \neq y_i
            \end{cases}
        \end{aligned}
    $}
    \label{eq:met_proxy_derivative}
\end{equation}
\begin{equation}
    \resizebox{0.35\textwidth}{!}{$
        \begin{aligned}
            \frac{\partial {\mathcal{L}_{sm}}_i}{\partial \vz_i} & = s.\sum^{C}_{k=1} \left[P^{(i)}_k-\1_\mathrm{k=y_i}\right] \times \\ 
            & \left[\frac{\partial \textit{Fnc}^{(i)}_k}{\partial \text{cos}(\theta_{\vw_k,\vz_i})}.\frac{\partial \text{cos}(\theta_{\vw_k,\vz_i})}{\partial \vz_i}+\frac{\partial \textit{Fnc}^{(i)}_k}{\partial \|\vz_i\|}.\frac{\partial \|\vz_i\|}{\partial \vz_i}\right] \\
            & := \vg_\theta + \vg_{\|\vz\|^-}
        \end{aligned}
    $}
    \label{eq:met_sm_emb_derivative}
\end{equation}
where $P^{(i)}_k$ is the probability output at class $k$ of image $i$. $\vg_{ac}$, $\vg_{mc}$, $\vg_\theta$ and $\vg_{\|\vz\|^-}$ denote the gradients affecting the actual proxy, non-actual (misclassified) proxy, embedding direction and embedding magnitude, respectively.  

Next, we perform a similar analysis with ${\mathcal{L}_{reg}}_i$.
\begin{equation}
    \resizebox{0.4\textwidth}{!}{$
        \frac{\partial {\mathcal{L}_{reg}}_i}{\partial \vw_k} = 0 ~~~ \text{and} ~~~ \frac{\partial {\mathcal{L}_{reg}}_i}{\partial \vz_i} = \frac{\partial {\mathcal{L}_{reg}}_i}{\partial \|\vz_i\|}.\frac{\partial \|\vz_i\|}{\partial \vz_i} := \vg_{\|\vz\|^+}
    $}
    \label{eq:met_rc_emb_derivative}
\end{equation}
The magnitude gradient is synthesized as
\begin{equation}
    \resizebox{0.2\textwidth}{!}{$
        \vg_{\|\mathbf{z}\|} = \vg_{\|\mathbf{z}\|^-} + \vg_{\|\mathbf{z}\|^+}
    $}
    \label{eq:met_mag_gradient}
\end{equation}

Subsequently, we provide the definitions of the previous margin strategies and analyze their properties (see \cref{subsubsec:const_soft_margin}). Accordingly, we introduce \textbf{\textit{hard margin}} to resolve the existing limitations (see \cref{subsubsec:hard_margin}).

\subsubsection{Constant \& Soft Margins}
\label{subsubsec:const_soft_margin}

\begin{figure*}[!t]
  \centering
  \begin{subfigure}{0.24\linewidth}
    \includegraphics[width=\textwidth]{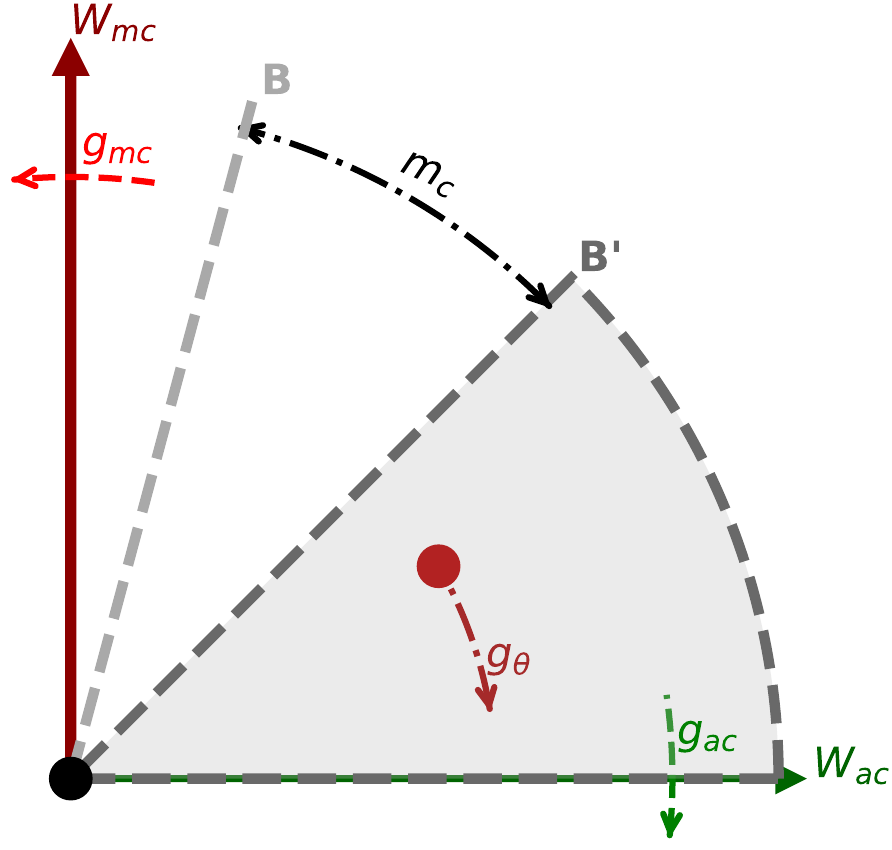}
    \caption{Constant Margin}
    \label{subfig:geo_const}
  \end{subfigure}
  \begin{subfigure}{0.24\linewidth}
    \includegraphics[width=\textwidth]{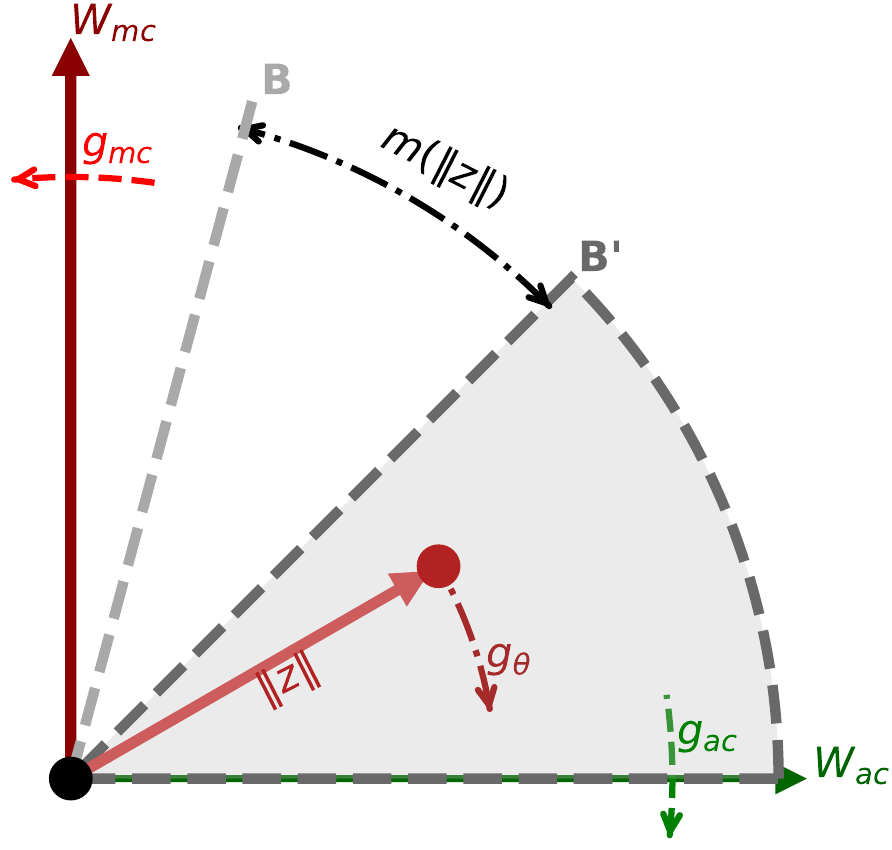}
    \caption{Soft Margin w/o MVP}
    \label{subfig:geo_soft_nvp}
  \end{subfigure}
  \begin{subfigure}{0.24\linewidth}
    \includegraphics[width=\textwidth]{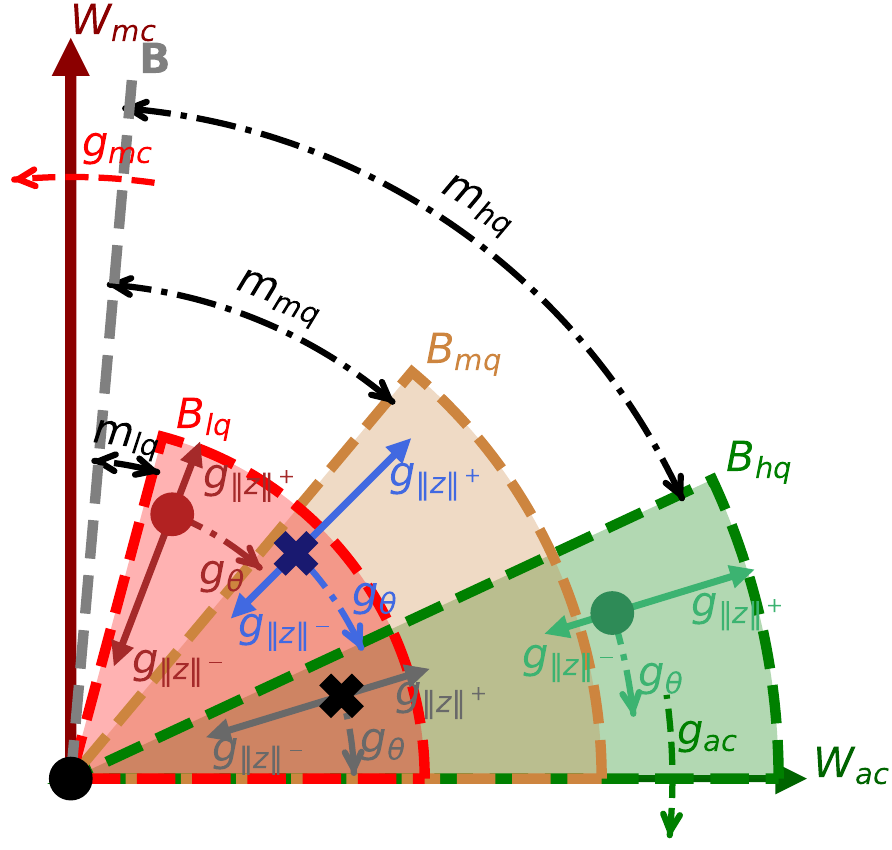}
    \caption{Soft Margin with MVP}
    \label{subfig:geo_soft_vp}
  \end{subfigure}
  \begin{subfigure}{0.24\linewidth}
    \includegraphics[width=\textwidth]{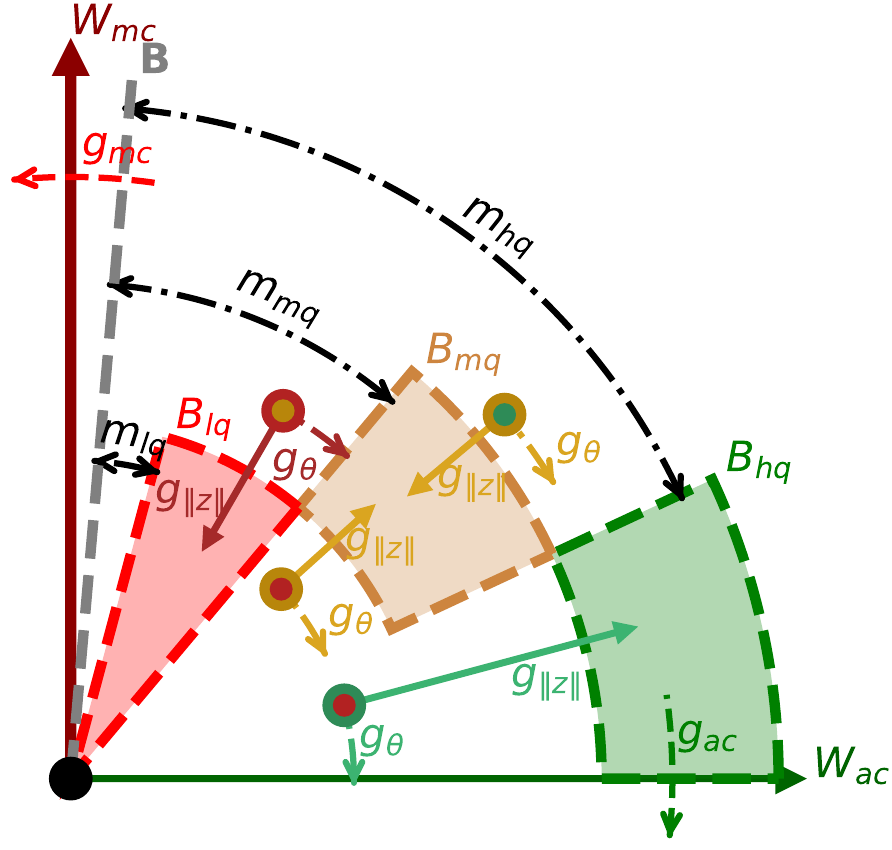}
    \caption{\textbf{\textit{Hard Margin (Ours)}}}
    \label{subfig:geo_hard}
  \end{subfigure}
  \caption{Illustration of the difference in geometric interpretation between margin strategies in hypersphere planning. \textbf{\textit{MVP}} is an abbreviation of \textbf{\textit{magnitude value-driven planning}}. \textbf{Circle} and \textbf{X} points express well-planned and unexpected-planned embedded features. In (c), the \textcolor{blue}{\textbf{blue feature}} has a large angular distance from its proxy, emphasizing $\vg_\theta$ over $\vg_{\|\vz\|^-}$ and yielding $\|\vg_{\|\vz\|^+}\| \gg \|\vg_{\|\vz\|^-}\|$. Conversely, the \textcolor{black}{\textbf{black feature}} shows the opposite trend due to its smaller angular distance. In (d), we apply a unified gradient (\textit{i.e.}, $\vg_{\|\vz\|}$) to influence the magnitude of the embedded features, thereby disregarding the issue of unregulated gradients typically encountered in the \textit{soft margin} approach. In (d), the feature's outline color indicates its expected region, while the center color represents its current region.}
  \label{fig:geo_interpre}
\end{figure*}

\noindent \textbf{Constant Margin.} It illustrates a margin strategy where the margin value is kept as a constant to adjust the decision boundary in hypersphere planning for all training samples, that is, $m_1$, $m_2$ and $m_3$ in \cref{eq:pre_add_margin} are constants. \Cref{subfig:geo_const} shows the geometric analysis of \textit{constant margin}, specifically, the location of original and adjusted decision boundaries (\textit{i.e.}, $\textbf{B}$ and  $\textbf{B}'$) and the direction of embedding gradient ($g_\theta$). However, there are two shortcomings in the \textit{constant margin} strategy.

\noindent \textbf{Lemma \textcolor{red}{1}.} \textit{Let $\vz_1, \vz_2 \in \mathbb{R}^d$ where $d$ denotes the embedding dimension. If $\|\vz_1\| > \|\vz_2\|$, then $\|\vg_\theta(\vz_1)\| < \|\vg_\theta(\vz_2)\|$.}\footnotemark[4]

First, the magnitude is not exploited in face representation due to the lack of constraint in \textit{magnitude value-driven planning (MVP)}. Second, based on Lemma \textcolor{red}{1}, samples initialized with low magnitude converge faster in training due to their high gradient. Consequently, the optimization path continuously favors small (magnitude) representations, which leads the magnitude values to be generally decreased. The magnitude shrinking has a negative impact on recognition ability enhancement when the low-recognizable images are initialized with low feature magnitude at the early training stage, causing suboptimal direction in the optimization process.

\vspace{1mm}

\noindent \textbf{Soft Margin.} The margin value in this category is calibrated based on the magnitude value of the embedded feature ($\|\vz_i\|$) representing each facial image. Lemma \textcolor{red}{1} shows that $\|\vz_i\|$ can partially express the recognizability level of an image. Specifically, a greater gradient magnitude ($\|\vg_\theta\|$) expresses the higher learning hardness, which reasonably reflects the lower recognizability of an image. Therefore, it is used as a recognizability proxy for each facial image to calibrate the boundary margin in hypersphere planning based on the recognizability level of that image. Within \textit{the soft margin strategy}, two approaches are considered based on the constraint of magnitude value: \textit{without} and \textit{with MVP}.

For the former approach (see \cref{subfig:geo_soft_nvp}), represented by AdaFace \cite{kim2022adaface}, the magnitude calculation does not join the backward process of $\mathcal{L}_{sm}$. Therefore, its value is planned in a free manner without any constraint. Accordingly, it leads to confusion in recognizability representation by feature magnitude, and also falls into small representation overfitting, similar to \textit{constant margin}.

The latter approach, represented by MagFace \cite{meng2021magface}, facilitates optimizing the direction and magnitude of embedded features simultaneously, specifically, $\vg_\theta \neq 0$ and $\vg_{\|z\|} \neq 0$. The notion of this approach is to emphasize $\vg_\theta$ for the easy learning sample and $\vg_{\|z\|^-}$ for the hard learning one (see well-planned points in \cref{subfig:geo_soft_vp}). Specifically, $m_2$ is a linear function with respect to $\|\mathbf{z_i}\|$. Subsequently, this design of $\mathcal{L}_{sm}$ establishes two optimal routes, \textit{i.e.} diminishing $\theta_{\vw_{y_i}}$ and reducing $\|\vz_i\|$. However, there also exists confusion in recognizability representation due to \textit{the mutual overlapping gradient} in hypersphere planning. The following property demonstrates the wrong-oriented path in the optimization process of FRL.

\noindent \textbf{Property \textcolor{red}{1}.}  \textit{In \cref{eq:gen_cen_loss}, if $m_2$ is a strictly increasing convex function of $\|\vz_i\|$ (\textit{i.e.}, $m_2(\|\vz_i\|)$), then the norm $\|\vz_i\|$ influences the gradient with respect to $\theta$ (\textit{i.e.}, $\vg_\theta$), while $\theta$ in turn affects the gradient with respect to $\|\vz_i\|$ (\textit{i.e.}, $\vg_{\|\vz_i\|}$). This manifests as a mutual overlapping gradient.}\footnotemark[4]

\footnotetext[4]{The proofs are detailed at Sec. C in SM.}

Based on Property \textcolor{red}{1}, instability in either the direction or magnitude of an embedded feature can lead to misdirection in hypersphere planning. The confusion appears in the overlapping representative region between low and high recognizability levels (see \cref{subfig:geo_soft_vp}). Specifically, the instability of \textcolor{blue}{\textbf{blue}} feature's direction makes $\|\vg_\theta\| \gg \|\vg_{\|\vz\|^-}\|$, thus, the magnitude is stretched ad hoc. Besides, there also exists a situation where a feature is planned with low angular distance and initialized with low magnitude (\textcolor{black}{\textbf{black}} point). In this situation, the convergence of the feature's direction at a small angular distance makes $\|\vg_{\|\vz\|^-}\| \gg \|\vg_\theta\|$, thus, the magnitude continuously shrinks unexpectedly.

\subsubsection{Hard Margin}
\label{subsubsec:hard_margin}

\noindent \textbf{Lemma \textcolor{red}{2}.} \textit{Suppose that $a_1, a_2 \in \mathbb{R}$ are two independent variables, and a function $f(a_1, a_2): \mathbb{R}^2 \rightarrow \mathbb{R}$. If $f$ can be described as a form of $f(a_1, a_2) = f_1(a_1) + f_2(a_2)$ where $f_1, f_2: \mathbb{R} \rightarrow \mathbb{R}$, then the mutual overlapping gradient calculated from the derivative of $f$ does not exist.}\footnotemark[4]

Based on Lemma \textcolor{red}{2}, the direction and magnitude, which are two independent properties of an embedded feature, must be optimized by the two loss components formed through linear aggregation to overcome the mutual overlapping problem existing in \textit{soft margin}. In \cref{eq:gen_cen_loss}, $\mathcal{L}_{sm}$ is hardly assigned to the optimization of feature directions due to its strength in enhancing intra-class compactness and inter-class discrepancy. Besides, $\mathcal{L}_{reg}$ is leveraged to quantify the recognizability level of an image by defining a \textbf{\textit{hard constraint}} in \textit{MVP}. Specifically, the guarantee of positive correlation between confidence score and magnitude value is facilitated by individually assigning optimization orientation in MVP for each sample. Accordingly, the planning strategy results in distinctive regions to eliminate confusion in recognizability representation (see \cref{subfig:geo_soft_vp,subfig:geo_hard}). 

\subsection{QCFace}
\label{subsec:qcface}
In this section, we devise \textbf{\textit{hard constraint}} for hypersphere planning to leverage the \textbf{\textit{hard margin}} strategy with two key objectives: recognition ability enhancement and quantifying recognizability, formulated to a novel loss function.

Based on the analysis in \cref{subsubsec:hard_margin}, the former is hardly assigned to $\mathcal{L}_{sm}$ to find the optimal direction of embedded features. For the design of $\mathcal{L}_{sm}$, we reuse ArcFace \cite{deng2019arcface} due to its best geometric attributes with a constant linear angular margin, and can guarantee the \textbf{\textit{frozen proxy}}, a condition for a well-orientated optimization (see \cref{subsec:ablation}). $\mathcal{L}_{sm}$ of image $i$ is shown as follows.
\begin{equation}
    \resizebox{0.43\textwidth}{!}{$
        \begin{aligned}
            {\mathcal{L}_{sm}}_i = -log\frac{e^{s.\text{cos}(\theta_{\vw_{y_i},\vz_i} + m_c)}}{e^{s.\text{cos}(\theta_{\vw_{y_i},\vz_i} + m_c)} + \sum_{j \neq y_i}^{C}e^{s.\text{cos}(\theta_{\vw_j,\vz_i})}} \\
            \text{where} ~~ m_c ~~ \text{is a constant.}
        \end{aligned}
    $}
    \label{eq:met_qcface_sm}
\end{equation}

\subsubsection{Regularization Loss}
\label{subsugsec:loss_reg}

Our main contribution lies in the design of the $\mathcal{L}_{reg}$ leveraging for \textit{MVP} to quantify the recognizability level of an image. Specifically, we embed the recognizability information into feature magnitudes, making them recognizability-measurable. However, instead of planning a magnitude value in a free manner like previous methods in \textit{soft margin strategy} (see \cref{subsubsec:const_soft_margin}), we define a \textbf{\textit{hard constraint}} to direct the optimization pathway for \textit{MVP}. Our design of $\mathcal{L}_{reg}$ of image $i$ is shown as follows.
\begin{equation}
    \resizebox{0.43\textwidth}{!}{$
        {\mathcal{L}_{reg}}_i = k.{p_d}_i. \left (\frac{1}{\|\vz_i\|}+\frac{\|\vz_i\|}{u_a^2} \right ) + (1-{p_d}_i).\left (\frac{1}{\|\vz_i\|}+\frac{\|\vz_i\|}{l_a^2} \right ) - b
    $}
    \label{eq:qcface_regular}
\end{equation}
where $l_a$ and $u_a$ are the defined lower and upper bounds of the magnitude value. $p_d$ is a guidance factor (value) to guide the optimization path for \textit{MVP} ($p_d \in \left[0, 1\right]$). $k$ is the linearizing coefficient for the correlation between $\|\vz_i\|$ and ${p_d}_i$. Offset $b$ is leveraged for converged tracking.

In \cref{eq:qcface_regular}, the aggregation of the two round bracket terms indicates that each training sample is assigned an expected value of feature magnitude $\|\vz_i\|^*$ (expected magnitude value), defined as the minimum of ${\mathcal{L}_{reg}}_i$ with respect to ${p_d}_i$. A higher value of $p_d$ of an image corresponds to a greater expected magnitude value $\vz$ and vice versa, demonstrated by the following properties.
\begin{equation}
    \resizebox{0.5\textwidth}{!}{$
        \begin{cases}
            \frac{\partial \mathbf{\|z\|^*}}{\partial {p_d}_i} > 0 ~~ \text{where} ~~ \mathbf{\|z\|^*} = \argmin_{\|\mathbf{z_i}\| \in \mathbb{R}^+} \mathcal{L}_{reg}  = \mathbf{f_z}({p_d}_i) \protect \footnotemark[4] \vspace{1mm}\\
            \argmin_{\|\mathbf{z_i}\| \in \mathbb{R}^+} \lim_{p_d \to 0} \mathcal{L}_{reg} = l_a \vspace{1mm}\\ 
            \argmin_{\|\mathbf{z_i}\| \in \mathbb{R}^+} \lim_{p_d \to 1} \mathcal{L}_{reg} = u_a
        \end{cases}
    $}
    \label{eq:met_limit}
\end{equation}

\subsubsection{Recognizability Guidance}
\label{subsugsec:reg_guide}
To the best of our knowledge, this is the first work to introduce the \textbf{\textit{guidance value}}, used for the \textbf{\textit{hard constraint}} to guide the learning of encoding recognizability by the feature magnitude. To orientate the optimization path towards the \textbf{\textit{hard margin}} strategy (see \cref{subfig:geo_hard}), the guidance value (\textit{i.e.}, $p_d$) has to satisfy two conditions: \textit{(i)} $p_d$ is a value that has lower and upper bounds; \textit{(ii)} $p_d$ has a positive correlation with the recognizability level. We explore that the probability of the actual label of an image predicted by a converged softmax-based model guarantees all conditions are well-suited to serve as a guidance value (see \cref{eq:qcface_guidance}).
\begin{equation}
    \resizebox{0.38\textwidth}{!}{$
        {p_d}_i = \frac{e^{s.\text{cos}(\theta_{\vw_{yi}, \vz_i})}}{e^{s.\text{cos}(\theta_{\vw_{yi}, \vz_i})} + \sum_{j \neq y_i}^{C}e^{s.\text{cos}(\theta_{\vw_j, \vz_i})}}
    $}
    \label{eq:qcface_guidance}
\end{equation}

\noindent From the definition of the guidance value $p_d$ of image $i$, the following properties can be easily established.

\noindent \textbf{Property \textcolor{red}{2}.} \textit{In \cref{eq:qcface_guidance}, $p_d$ is bounded in $\left[0, 1\right]$.}\footnotemark[4]

\noindent \textbf{Property \textcolor{red}{3}.} \textit{Suppose that $\vz_1, \vz_2 \in \mathbb{R}^d$ where $d$ denotes the embedding dimension, and let their corresponding guidance values (\textit{i.e.}, ${p_d}_1, {p_d}_2 \in [0, 1]$) be computed by \cref{eq:qcface_guidance}. If ${p_d}_1 > {p_d}_2$, then $\|\vg_\theta(\vz_1)\| < \|\vg_\theta(\vz_2)\| ~ \forall \vz_1, \vz_2$.}\footnotemark[4]

\footnotetext[4]{The proofs are detailed at Secs. C and D in SM.}

According to Property \textcolor{red}{3}, ${p_d}$ exhibits a negative correlation with $\|\vg_\theta\|$. Besides, $\|\vg_\theta\|$ can also be interpreted as the learning hardness, rationally expressing an inverse relationship with the recognizability of an image. Thus, ${p_d}$ definitely satisfies the last condition. Besides, \textbf{\textit{the high learning hardness is also expressed by the mislabeled samples in training data, occurring in most large-scale datasets}}. 

In addition, to guarantee the \textbf{\textit{hard margin}} strategy, namely, to eliminate \textit{mutual overlapping gradient} between $\mathcal{L}_{sm}$ and $\mathcal{L}_{reg}$, $p_d$ serves as an independent property of an image; that is, it does not participate in the optimization process. Therefore, it must be detached from the computational graph before calculating $\mathcal{L}_{reg}$. 

\subsubsection{Loss Constraints}
\label{subsubsec:loss_strategy}

For the success in training, $\mathcal{L}_{reg}$ has to be designed to guarantee that the following constraints are satisfied\footnotemark[4]: \\
\textbf{Constraint for Convergence.} $\mathcal{L}_{reg}$ is the strictly convex downward function and \resizebox{0.27\textwidth}{!}{$\mathcal{L}_{reg} \geq 0 ~~\forall \|\mathbf{z}\| \in \left[l_a, u_a\right], \forall p_d \in \left[0, 1\right]$}.\\
\textbf{Constraint for balancing encoding value.} The expected magnitude value $\|\mathbf{z}\|^*$ exhibits a linear dependence on $p_d$.\\
\textbf{Constraint for tracking regularization loss.} In \cref{eq:qcface_regular}, $\min_{\|\mathbf{z}\| \in \mathbb{R}^+} \mathcal{L}_{reg}(\|\mathbf{z}\|, p_d) = 0 ~ \forall ~ p_d \in \left[0, 1\right]$.

The first constraint is to guarantee the expected optimization pathway in hypersphere planning. Besides, the second is utilized to find the best $k$ for the best linear correlation between $p_d$ and $\|\mathbf{z}\|$ in \cref{eq:qcface_regular}, while the third is utilized to find the best bias $b$ for tracking the convergence point.

\footnotetext[4]{The proofs are detailed at Sec. D in SM.}

\section{Experiments}
\label{sec:experiments}

\subsection{Dataset}
\label{subsec:exp_dataset}
\noindent \textbf{Training data.} Beside some large-scale datasets (e.g., MS-Celeb-1M \cite{guo2016ms}, MS1MV2 \cite{deng2019arcface}, and WebFace4M \cite{zhu2021webface260m}), CASIA-WebFace \cite{deng2019lightweight} and  MS1MV3 \cite{deng2019lightweight} are chosen as our main training dataset due to the low-noise nature of the data. The datasets are used to train the model and provide a fair comparison of \textbf{\textit{QCFace}}’s performance with all reproduced reputable methods (see \cref{subsec:exp_comp}). 

\vspace{1mm}

\noindent \textbf{Evaluation data.} Our evaluation datasets\footnotemark[5] include high-quality benchmark datasets (\textit{i.e.}, AgeDB-30 \cite{moschoglou2017agedb}, CFP-FP \cite{sengupta2016frontal}, LFW \cite{huang2008labeled}, CALFW \cite{zheng2017cross}, CPLFW \cite{zheng2018cross} and XQLFW \cite{knoche2021cross}), mixed-quality ones (\textit{i.e.}, IJB-B \cite{whitelam2017iarpa} and IJB-C \cite{maze2018iarpa}), and a low-quality (\textit{i.e.}, TinyFace \cite{cheng2018low}). These are all popular and public benchmark datasets to provide a comprehensive evaluation. Moreover, \textbf{\textit{QCFace}} is also evaluated on MegaFace Challenge 1 \cite{kemelmacher2016megaface} by utilizing the official devkit.

\footnotetext[5]{Datasets, augmentations and hyperparameters are detailed at Secs. D and E in SM.}

\subsection{Experimental Setup}
\label{subsec:exp_imp}
\noindent \textbf{Training setting.} Our experiments are conducted on a GPU NVIDIA A100 40GB. SGD \cite{ruder2016overview} is used as our main optimizer. The epoch count is 25, applying for the training of all \textbf{\textit{QCFace}} variants and the reproduced reputable methods. The learning rate is initialized to $0.01$ with step scheduling at epochs 10, 18, and 22.

\vspace{1mm}

\noindent \textbf{Model setting.} The input images are transformed using multiple augmentations\footnotemark[5], resized to $112 \times 112$, and normalized. For the backbone, we use IResNet \cite{duta2021improved} as our main backbone for the evaluation due to its diversity in size and prevalence (\textit{i.e.}, IResNet18 and IResNet100). These settings are uniformly applied across all reproduced methods for a fair benchmark. Hyperparameters of \textbf{\textit{QCFace}}\footnotemark[5] (see \cref{eq:qcface_regular}) are chosen based on the loss constraints (see \cref{subsubsec:loss_strategy}).

\subsection{Evaluation Metrics}
\label{subsec:exp_metric}
Our extensive experiments demonstrate the effectiveness of \textbf{\textit{QCFace}} in terms of recognizability representation and recognition ability enhancement. First, the former is evaluated by analyzing the distribution of feature vectors on the hypersphere. Second, we compare \textbf{\textit{QCFace}} with the well-known FR methods based on $1:1$ verification and identification accuracy, and AUC-ROC to assess the latter aspect based on the two corresponding properties, \textit{i.e.} recognition accuracy and cognitive ability of an FR model.

\begin{figure*}[!t]
  \centering
  \begin{subfigure}{0.28\linewidth}
    \includegraphics[width=\textwidth]{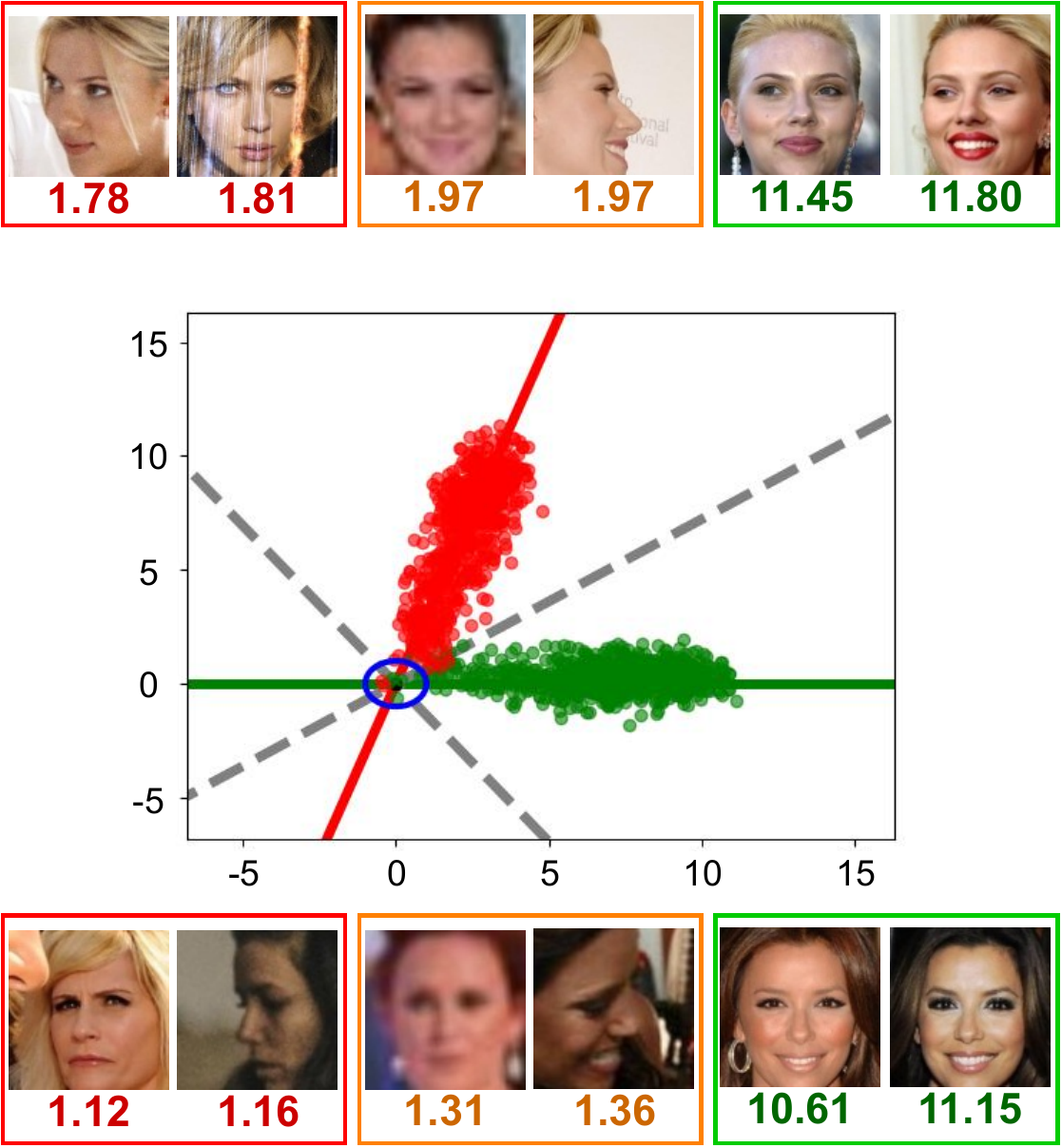}
    \caption{AdaFace \cite{kim2022adaface}}
    \label{fig:mp_ada}
  \end{subfigure}
  \hfill
  \begin{subfigure}{0.28\linewidth}
    \includegraphics[width=\textwidth]{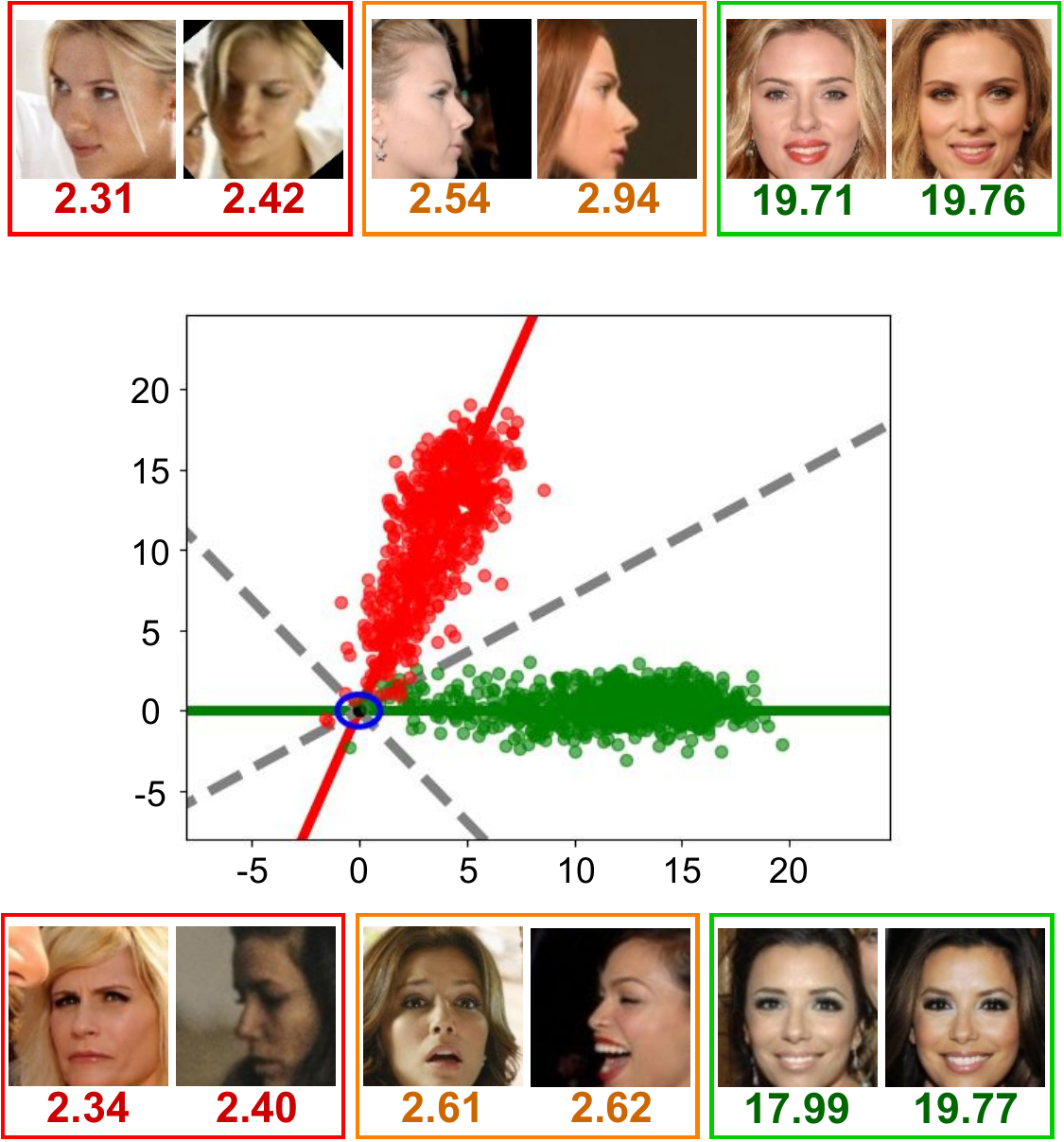}
    \caption{MagFace \cite{meng2021magface}}
    \label{fig:mp_mag}
  \end{subfigure}
  \hfill
  \begin{subfigure}{0.28\linewidth}
    \includegraphics[width=\textwidth]{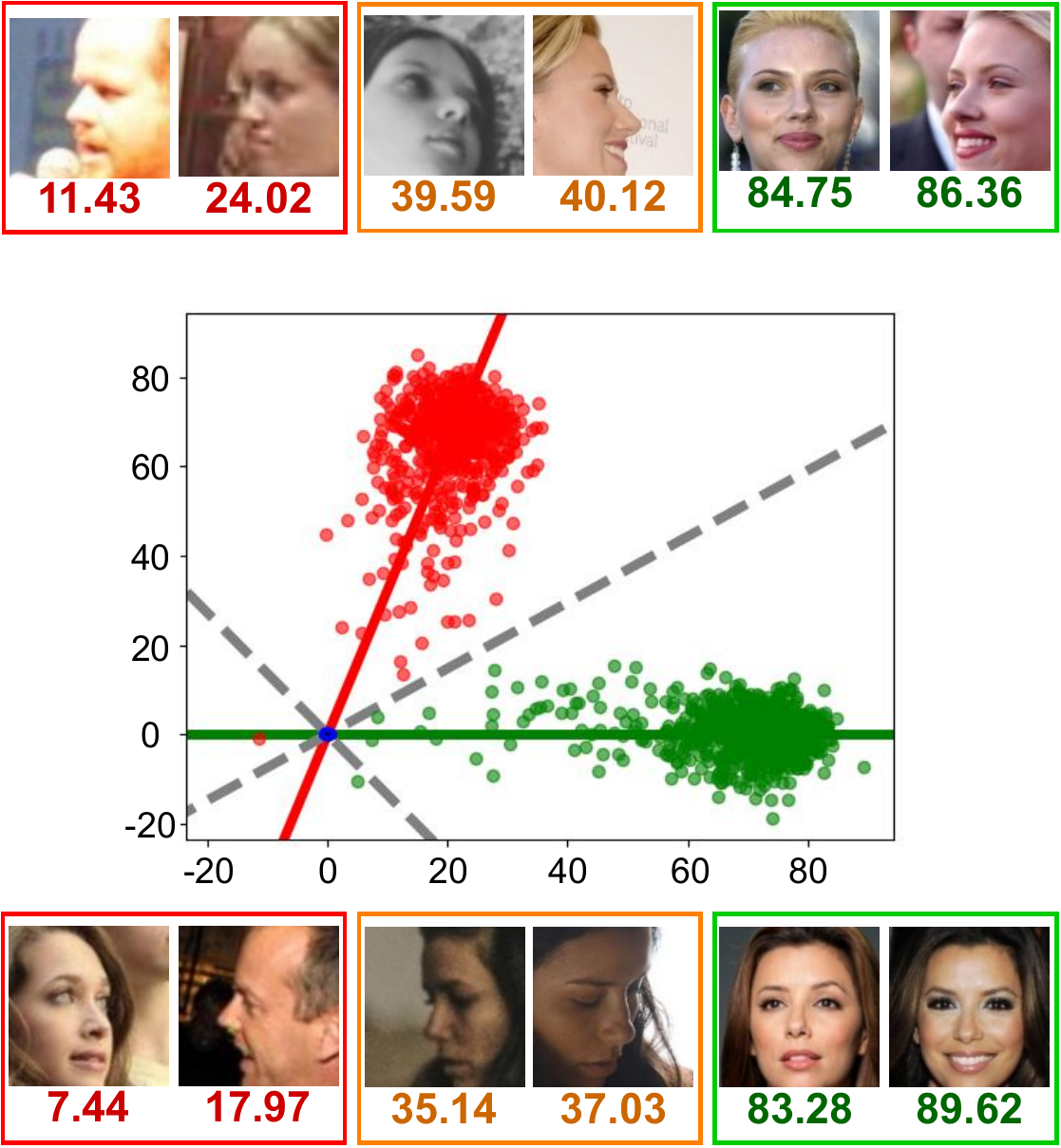}
    \caption{\textbf{\textit{QCFace-Arc (Ours)}}}
    \label{fig:mp_qc}
  \end{subfigure}
  \caption{The non-normalized geometrical representation of the feature space optimized by MagFace, AdaFace and \textbf{\textit{QCFace-Arc}}. The feature vectors of each image extracted by face backbones are projected to the hyperplane formed by two proxies of two identities in the CASIA-WebFace dataset by the Gram-Schmidt algorithm. The \textbf{border} \textcolor{red}{red} and \textcolor{OliveGreen}{green} lines express the proxies of two identities. The actual proxy of each mapping feature point has the same color as that of feature. The \textcolor{gray}{gray} dashed lines are the internal and external bisectors. The number below each image sample is the magnitude of the mapping feature vector.\protect\footnotemark[7]}
  \label{fig:mapping_point}
\end{figure*}

\begin{figure*}[!t]
  \centering
   \begin{subfigure}{0.42\linewidth}
    \includegraphics[width=\textwidth]{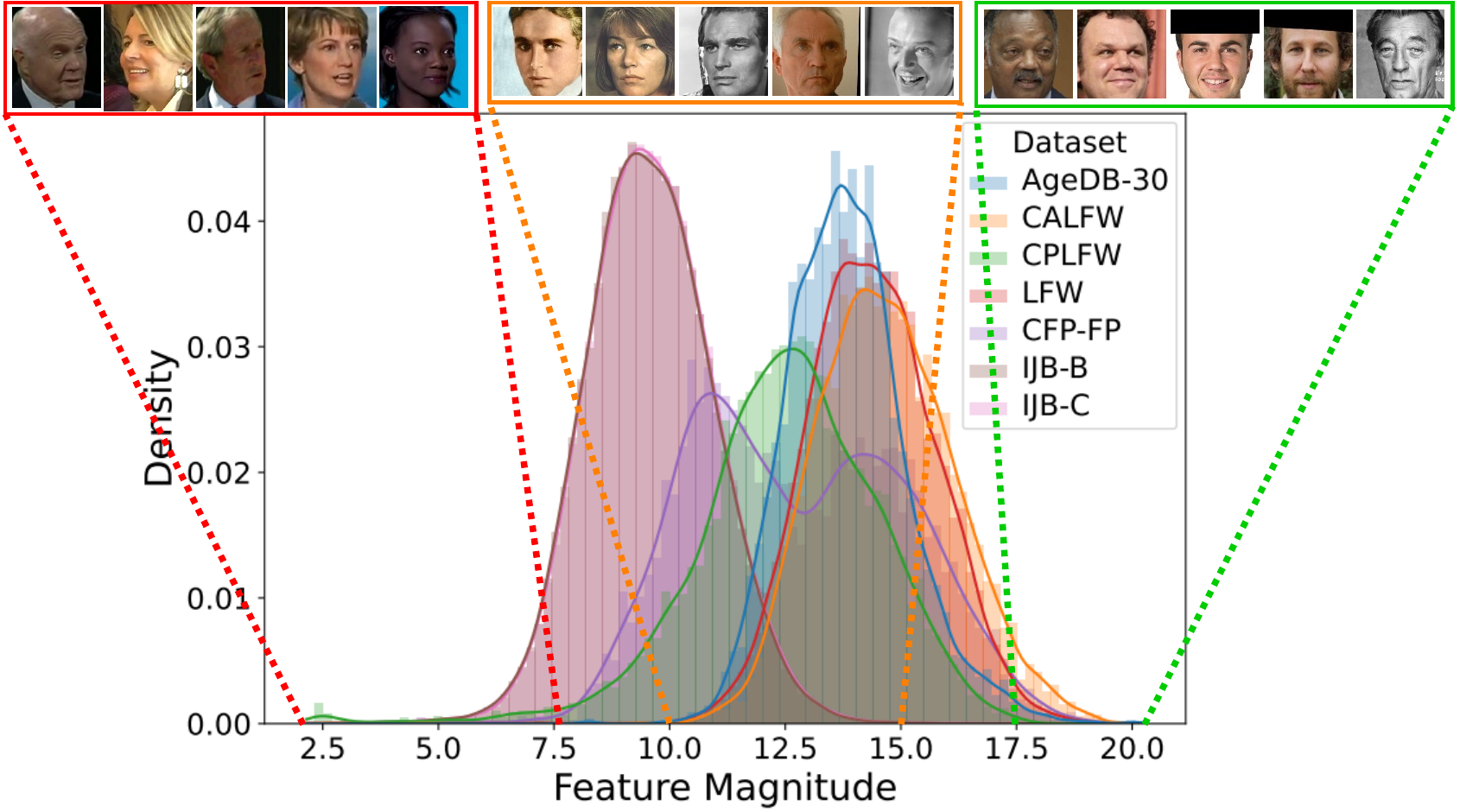}
    \caption{MagFace \cite{meng2021magface}}
    \label{fig:maghis_mag}
  \end{subfigure}
  \begin{subfigure}{0.42\linewidth}
    \includegraphics[width=\textwidth]{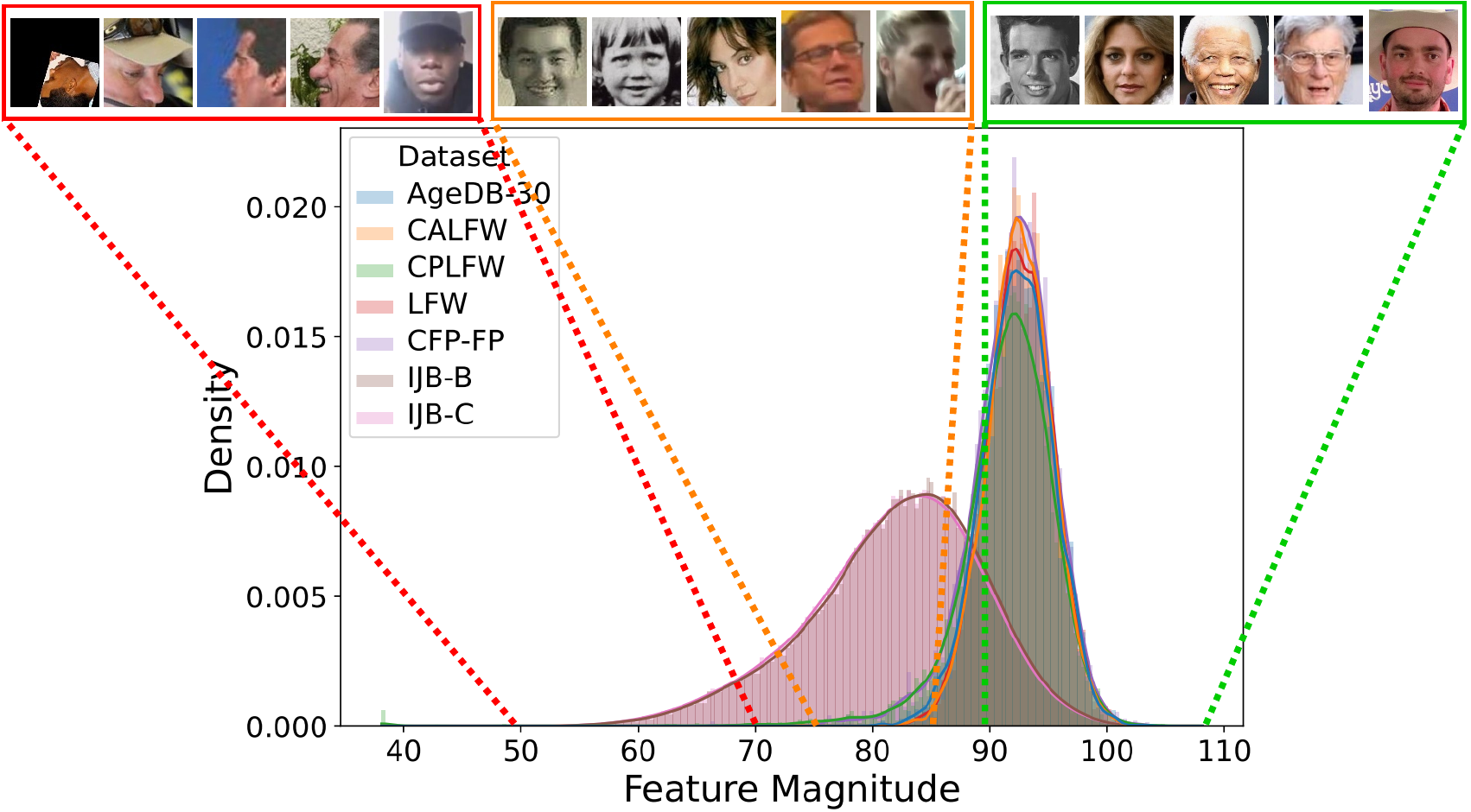}
    \caption{\textbf{\textit{QCFace (Ours)}}}
    \label{fig:maghis_qc}
  \end{subfigure}
  \caption{The magnitude distribution comparison between MagFace and \textbf{\textit{QCFace-Arc}}.}
  \label{fig:mag_hist}
\end{figure*}

\subsection{Results and Analyses}
\label{subsec:exp_comp}
In this section, we first show the recognizability representation capacity of \textbf{\textit{QCFace}} and the \textit{soft margin} methods \cite{meng2021magface, kim2022adaface}. Next, we analyze the recognition ability of \textbf{\textit{QCFace}} in comparison with the well-known approaches.

\subsubsection{Recognizability Representation}
\label{subsubsec:mag_encode}
In \cref{fig:mapping_point}, the efficiency in planning the placement of proxies for the three methods is the same. Specifically, they are clearly distinguished by the large angle (\textit{i.e.}, $>60^o$). Moreover, the geometrical representation of all three methods also shows that all mapping features are concentrated near their actual proxies, which is benefited from $\mathcal{L}_{sm}$. 

Based on Lemma \textcolor{red}{1}, \cref{fig:mp_ada,fig:mp_mag} show the consequence in the small representation of the optimization path in the experimental aspect. Specifically, the embedded features of MagFace \cite{meng2021magface} and AdaFace \cite{kim2022adaface} are uniformly distributed in the low range of magnitude value (\textit{i.e.,} $<15$ and $<20$, respectively). Therefore, although AdaFace and MagFace utilize \textit{soft margin} strategy to exploit the magnitude value of embedded features, they cannot overcome the small representation in the optimization (see \cref{subsubsec:const_soft_margin}). For MagFace, based on Property \textcolor{red}{1}, \textit{the mutual overlapping gradient} causes the confusion in recognizability representation demonstrated by the samples in the low range of magnitude value (see \textcolor{red}{red} and \textcolor{orange}{orange} boxes in \cref{fig:mp_mag,fig:maghis_mag}). For AdaFace, due to the lack of MVP, the magnitude value cannot fully represent the recognizability level of facial images (see \textcolor{red}{red} and \textcolor{orange}{orange} boxes in \cref{fig:mp_ada}). 

For \textbf{\textit{QCFace}}, the embedded features are planned in the widest range $\left[1, 100\right]$ (see \cref{fig:mp_qc}), which is consistent with the originally defined values of $l_a$ and $u_a$ (see \cref{eq:qcface_regular}). By designing the loss function obeying \textbf{\textit{hard margin}} strategy (see \cref{subsubsec:hard_margin}), the confusion in recognizability representation by magnitude values is significantly reduced (see the samples in  \cref{fig:mp_qc,fig:maghis_qc}). The positive correlation between recognizability level and magnitude value is successfully facilitated by the guidance value defined in \cref{eq:qcface_guidance} (see \cref{subsugsec:reg_guide}). This property enables low-magnitude values to indicate not only low recognizability but also mislabeled samples, occurring in large-scale datasets.

\footnotetext[7]{The extended version is shown at Sec. H in SM.}

On benchmark datasets, MagFace again expresses its confusion in recognizability representation (see \cref{fig:maghis_mag}), which is caused by small representation and \textit{mutual overlapping gradient} problems. Based on the magnitude distribution, the histogram of CFP-FP has two peaks, resulting from overfitting facial pose in MVP. It is also a consequence of small representation in optimization. In contrast, \textbf{\textit{QCFace}} shows its wide representative range of magnitude, demonstrating strong generalization capabilities of various recognizability levels in each benchmark dataset (see \cref{fig:maghis_qc}). The face-pose overfitting problem is significantly mitigated in recognizability representation. Moreover, \textbf{\textit{QCFace}} shows a marked distinction of the magnitude distribution between mixed-quality and high-quality datasets compared to MagFace (see \cref{fig:mag_hist}). Specifically, the mean and variance of the former are significantly lower and higher, respectively, than those of the latter. These results clearly underscore the distinction in image quality between the two dataset types in a logically consistent manner.

\subsubsection{Recognition ability enhancement}
\label{subsubsec:veri_comp}

In \cref{tab:comparison_hq}, top-1 and top-2 verification accuracies are all achieved by the versions of \textbf{\textit{QCFace}} with IResNet100 on all high-quality datasets. Specifically, \textbf{\textit{QQCFace-Arc}} viewed as the best version of \textbf{\textit{QCFace}}, surpasses the SoTAs on all high-quality datasets, achieving accuracy improvements of $0.701\%$, $2.544\%$, $0.050\%$, $0.227\%$, $2.107\%$ and $2.781\%$ on AgeDB-30, CFP-FP, LFW, CALFW, CPLFW and XQLFW, respectively. Consequently, \textbf{\textit{QCFace}} successfully improves its recognition ability by disregarding the small representation problem in previous approaches (see \cref{subsubsec:const_soft_margin}).

For the mixed-quality datasets, \textbf{\textit{(Q)QCFace-Arc}} shows performance improvements on IJB-B and IJB-C. Specifically, our approach improves $4.489\%$ and $2.012\%$ on IJB-B at FAR=$\{10^{-5},10^{-4}\}$, and $2.693\%$, $3.337\%$ and $1.601\%$ on IJB-C at FAR=$\{10^{-6},10^{-5},10^{-4}\}$, and achieves comparable performance in AUC-ROC (see \cref{tab:comparison_veri_IJB}). These results yield the improvement of \textbf{\textit{QCFace}} compared with other methods in cognitive ability. However, at FAR=$10^{-6}$, \textbf{\textit{QCFace}} cannot surpass the previous methods on IJB-B dataset. This observation reveals a limitation of \textbf{\textit{QCFace}} in handling very low-quality facial images, which stems from the insufficient contribution of the softmax loss (\textit{i.e.}, $\mathcal{L}_{sm}$) to the optimization of recognition performance. 

For the identification benchmark, \textbf{\textit{(Q)QCFace}} shows its superiority by outperforming all reputable methods. Specifically, \textbf{\textit{QQCFace}} improves $1.437\%$, $1.355\%$ and $6.746\%$ at rank-1 accuracy, and $0.882\%$, $0.995\%$ and $4.482\%$ at rank-5 accuracy, on IJB-B, IJB-C and TinyFace. On MegaFace Challenge, \textbf{\textit{QCFace}} improves $2.486\%$ and $1.868\%$ on identification and verification accuracies, respectively.

\begin{table}[!t]
\caption{Comparison in verification accuracy on high-quality datasets. The \textbf{bold}, \uline{underlined} and \textit{italic} numbers correspondingly express \textbf{top-1}, \uline{top-2} and \textit{top-3} accuracy. In notation \textbf{\textit{(Q)QCFace}-X}, \textbf{\textit{(Q)}} expresses the use of \cite{terhorst2023qmagface} as a similarity score calculation instead of cosine similarity in decision making, \textbf{\textit{X}} expresses the name of softmax loss (\textit{i.e.}, Arc - ArcFace, Cur - CurricularFace, MVS - MVSoftmax). \textcolor{YellowGreen}{Green} and \textcolor{Salmon}{red} backgrounds express the use of margin-based and misclassified softmax losses.\protect\footnotemark[7]}
\label{tab:comparison_hq}
\begin{adjustbox}{width=0.9\columnwidth,center}
\begin{tabular}{c|l|c;{1pt/1pt}c;{1pt/1pt}c;{1pt/1pt}c;{1pt/1pt}c;{1pt/1pt}c} 
\hline
\rowcolor[rgb]{0.784,0.784,0.784} \textbf{Year} & \multicolumn{1}{c|}{\textbf{Method}} & \multicolumn{1}{c|}{\textbf{AgeDB-30}} & \multicolumn{1}{c|}{\textbf{CFP-FP}} & \multicolumn{1}{c|}{\textbf{LFW}} & \multicolumn{1}{c|}{\textbf{CALFW}} & \multicolumn{1}{c|}{\textbf{CPLFW}} & \textbf{XQLFW}   \\ 
\hline\hline
\textit{2018}                                   & CosFace \cite{wang2018cosface}                             & 97.033                                 & 95.743                               & 99.750                            & 95.467                              & 90.317                              & 81.783           \\ 
\hdashline[1pt/1pt]
\textit{2019}                                   & ArcFace \cite{deng2019arcface}                             & 97.133                                 & 95.100                               & 99.633                            & 95.517                              & 90.217                              & 81.833           \\ 
\hdashline[1pt/1pt]
\textit{2020}                                   & MV-Arc-Softmax \cite{wang2020mis}                      & 97.100                                 & 95.757                               & 99.700                            & 95.617                              & 90.233                              & 83.150           \\ 
\hdashline[1pt/1pt]
\textit{2020}                                   & CurricularFace \cite{huang2020curricularface}                      & 97.100                                 & 94.686                               & 99.550                            & 95.700                              & 90.083                              & 82.650           \\ 
\hdashline[1pt/1pt]
\textit{2021}                                   & MagFace \cite{meng2021magface}                             & 96.867                                 & 95.357                               & 99.600                            & 95.517                              & 89.583                              & 81.233           \\ 
\hdashline[1pt/1pt]
\textit{2021}                                   & VPLFace \cite{deng2021variational}                             & 97.183                                 & 94.257                               & 99.700                            & 95.683                              & 89.617                              & 83.117           \\ 
\hdashline[1pt/1pt]
\textit{2022}                                   & SphereFace2 \cite{wen2021sphereface2}                         & 96.000                                 & 94.029                               & 99.583                            & 95.283                              & 90.100                              & 83.933           \\ 
\hdashline[1pt/1pt]
2022                                            & Elastic-Arc \cite{boutros2022elasticface}                         & 97.283                                 & 94.329                               & 99.467                            & 95.500                              & 89.800                              & 82.467           \\ 
\hdashline[1pt/1pt]
2022                                            & Elastic-Cos \cite{boutros2022elasticface}                         & 97.417                                 & 95.357                               & 99.700                            & 95.733                              & 89.400                              & 82.350           \\ 
\hdashline[1pt/1pt]
\textit{2022}                                   & AdaFace \cite{kim2022adaface}                             & 96.817                                 & 95.614                               & 99.717                            & 95.733                              & 90.917                              & 82.017           \\ 
\hdashline[1pt/1pt]
\textit{2023}                                   & QAFace \cite{saadabadi2023quality}                              & 96.783                                 & 95.243                               & 99.733                            & 95.400                              & 89.817                              & 83.550           \\ 
\hdashline[1pt/1pt]
\textit{2023}                                   & QMagFace \cite{terhorst2023qmagface}                            & 97.250                                 & 96.014                               & 99.700                            & 95.617                              & 90.683                              & 81.483           \\ 
\hdashline[1pt/1pt]
\textit{2023}                                   & UniFace \cite{zhou2023uniface}                             & 97.083                                 & 95.686                               & 99.617                            & 95.667                              & 90.683                              & 82.700           \\ 
\hdashline[1pt/1pt]
\textit{2023}                                   & UniTSFace \cite{jia2023unitsface}                           & 97.217                                 & 95.371                               & 99.667                            & 95.717                              & 90.450                              & 83.517           \\ 
\hdashline[1pt/1pt]
\textit{2024}                                   & TopoFR \cite{dan2024topofr}                              & 97.500                                 & 95.871                               & 99.750                            & 95.650                              & 90.383                              & 83.583           \\ 
\hline\hline
\rowcolor[rgb]{0.765,0.918,0.773} \textit{Now}  & \textbf{\textit{QCFace-Arc}}         & \uline{98.100}                         & \textit{98.200}                      & \textbf{99.800}                   & \textit{95.950}                     & \uline{92.483}                      & \uline{86.067}   \\ 
\hdashline[1pt/1pt]
\rowcolor[rgb]{0.765,0.918,0.773} \textit{Now}  & \textit{\textbf{QQCFace-Arc}}        & \textbf{98.183}                        & \textbf{98.457}                      & \textbf{99.800}                   & \textit{95.950}                     & \textbf{92.833}                     & \textbf{86.267}  \\ 
\hline\hline
\rowcolor[rgb]{0.957,0.741,0.741} \textit{Now}  & \textit{\textbf{QCFace-Cur}}         & 97.700                                 & \textit{98.200}                      & \uline{99.783}                    & \textbf{96.083}                     & 92.383                              & 83.283           \\ 
\hdashline[1pt/1pt]
\rowcolor[rgb]{0.957,0.741,0.741} \textit{Now}  & \textit{\textbf{QQCFace-Cur}}        & \textit{97.917}                        & \uline{98.271}                       & \textit{99.767}                   & \uline{96.067}                      & \textit{92.467}                     & 83.283           \\ 
\hline\hline
\rowcolor[rgb]{0.957,0.741,0.741} \textit{Now}  & \textit{\textbf{QCFace-MVS}}         & 93.200                                 & 96.229                               & 99.483                            & 91.683                              & 89.133                              & 82.833           \\ 
\hdashline[1pt/1pt]
\rowcolor[rgb]{0.957,0.741,0.741} \textit{Now}  & \textit{\textbf{QQCFace-MVS}}        & 93.517                                 & 96.186                               & 99.467                            & 91.833                              & 88.817                              & \textit{84.300}  \\
\hline
\end{tabular}
\end{adjustbox}
\end{table}

\begin{table}[!t]
\caption{Comparison in verification performance on IJB dataset.\protect\footnotemark[7]}
\label{tab:comparison_veri_IJB}
\begin{adjustbox}{width=0.9\columnwidth,center}
\begin{tabular}{c|l|c;{1pt/1pt}c;{1pt/1pt}c;{1pt/1pt}c|c;{1pt/1pt}c;{1pt/1pt}c;{1pt/1pt}c} 
\hline
\rowcolor[rgb]{0.784,0.784,0.784} {\cellcolor[rgb]{0.784,0.784,0.784}}                                & \multicolumn{1}{c|}{{\cellcolor[rgb]{0.784,0.784,0.784}}}                                  & \multicolumn{4}{c|}{\textbf{IJB-B (TAR@FAR)}}                                                                                                                                       & \multicolumn{4}{c}{\textbf{IJB-C (TAR@FAR)}}                                                                                                                                         \\ 
\cline{3-10}
\rowcolor[rgb]{0.784,0.784,0.784} \multirow{-2}{*}{{\cellcolor[rgb]{0.784,0.784,0.784}}\textbf{Year}} & \multicolumn{1}{c|}{\multirow{-2}{*}{{\cellcolor[rgb]{0.784,0.784,0.784}}\textbf{Method}}} & \multicolumn{1}{c|}{\textbf{10\textsuperscript{-6}}} & \multicolumn{1}{c|}{\textbf{10\textsuperscript{-5}}} & \multicolumn{1}{c|}{\textbf{10\textsuperscript{-4}}} & \textbf{AUC}   & \multicolumn{1}{c|}{\textbf{10\textsuperscript{-6}}} & \multicolumn{1}{c|}{\textbf{10\textsuperscript{-5}}} & \multicolumn{1}{c|}{\textbf{10\textsuperscript{-4}}} & \textbf{AUC}    \\ 
\hline\hline
\textit{2018}                                                                                         & CosFace \cite{wang2018cosface}                                                                                   & 37.81                                                & 84.36                                                & 91.82                                                & 99.44          & 82.44                                                & 90.02                                                & 94.07                                                & 99.57           \\ 
\hdashline[1pt/1pt]
\textit{2019}                                                                                         & ArcFace \cite{deng2019arcface}                                                                                   & 36.71                                                & 85.15                                                & 91.61                                                & 99.47          & 82.85                                                & 89.40                                                & 93.66                                                & 99.59           \\ 
\hdashline[1pt/1pt]
\textit{2020}                                                                                         & MV-Arc-Softmax \cite{wang2020mis}                                                                            & \textit{39.02}                                       & 80.41                                                & 90.87                                                & 99.44          & 79.61                                                & 87.82                                                & 93.26                                                & 99.59           \\ 
\hdashline[1pt/1pt]
\textit{2020}                                                                                         & CurricularFace \cite{huang2020curricularface}                                                                            & 37.63                                                & 84.99                                                & 92.27                                                & 99.40          & \textit{85.12}                                       & \textit{90.79}                                       & 94.20                                                & 99.53           \\ 
\hdashline[1pt/1pt]
\textit{2021}                                                                                         & MagFace \cite{meng2021magface}                                                                                   & 38.10                                                & 83.59                                                & 91.47                                                & 99.32          & 81.64                                                & 88.81                                                & 93.38                                                & 99.51           \\ 
\hdashline[1pt/1pt]
\textit{2021}                                                                                         & VPLFace \cite{deng2021variational}                                                                                   & 36.90                                                & \textit{85.55}                                       & 91.64                                                & 99.42          & \uline{86.52}                                        & 90.26                                                & 93.79                                                & 99.55           \\ 
\hdashline[1pt/1pt]
\textit{2022}                                                                                         & SphereFace2 \cite{wen2021sphereface2}                                                                               & 36.79                                                & 80.86                                                & 90.10                                                & \uline{99.50}  & 78.70                                                & 86.67                                                & 92.34                                                & \textit{99.63}  \\ 
\hdashline[1pt/1pt]
2022                                                                                                  & Elastic-Arc \cite{boutros2022elasticface}                                                                               & 36.48                                                & 83.97                                                & 91.79                                                & 99.45          & 80.43                                                & 89.95                                                & \textit{94.33}                                       & 99.56           \\ 
\hdashline[1pt/1pt]
2022                                                                                                  & Elastic-Cos \cite{boutros2022elasticface}                                                                               & 34.37                                                & 83.72                                                & 91.67                                                & \uline{99.50}  & 82.48                                                & 89.65                                                & 93.80                                                & 99.61           \\ 
\hdashline[1pt/1pt]
\textit{2022}                                                                                         & AdaFace \cite{kim2022adaface}                                                                                   & 35.01                                                & 85.07                                                & \textit{92.44}                                       & 99.38          & 80.43                                                & 89.95                                                & \textit{94.33}                                       & 99.56           \\ 
\hdashline[1pt/1pt]
\textit{2023}                                                                                         & QAFace \cite{saadabadi2023quality}                                                                                    & \uline{40.00}                                        & 80.19                                                & 89.49                                                & \textbf{99.54} & 79.02                                                & 87.17                                                & 92.27                                                & \textit{99.63}  \\ 
\hdashline[1pt/1pt]
\textit{2023}                                                                                         & QMagFace \cite{terhorst2023qmagface}                                                                                  & 36.80                                                & 83.42                                                & 88.25                                                & 99.38          & 81.24                                                & 87.28                                                & 91.23                                                & 99.61           \\ 
\hdashline[1pt/1pt]
\textit{2023}                                                                                         & UniFace \cite{zhou2023uniface}                                                                                   & 37.16                                                & 85.46                                                & 92.16                                                & 99.41          & 82.62                                                & 90.35                                                & 94.21                                                & 99.57           \\ 
\hdashline[1pt/1pt]
\textit{2023}                                                                                         & UniTSFace \cite{jia2023unitsface}                                                                                 & 34.35                                                & 82.71                                                & 91.64                                                & 99.47          & 79.78                                                & 88.66                                                & 93.64                                                & \uline{99.64}   \\ 
\hdashline[1pt/1pt]
\textit{2024}                                                                                         & TopoFR \cite{dan2024topofr}                                                                                    & \textbf{40.25}                                       & 82.88                                                & 91.25                                                & \textit{99.49} & 77.94                                                & 88.28                                                & 93.30                                                & \textit{99.63}  \\ 
\hline\hline
\rowcolor[rgb]{0.765,0.918,0.773} \textit{Now}                                                        & \textbf{\textit{QCFace-Arc}}                                                               & 34.42                                                & \textbf{89.39}                                       & \textbf{94.30}                                       & 99.44          & \textbf{88.85}                                       & \textbf{93.82}                                       & \textbf{95.84}                                       & 99.62           \\ 
\hdashline[1pt/1pt]
\rowcolor[rgb]{0.765,0.918,0.773} \textit{Now}                                                        & \textbf{\textit{QQCFace-Arc}}                                                              & 37.89                                                & \uline{88.80}                                        & \uline{93.93}                                        & \textit{99.48} & 80.57                                                & \uline{91.70}                                        & \uline{95.48}                                        & \textbf{99.66}  \\
\hline
\end{tabular}
\end{adjustbox}
\end{table}

\begin{table}[!t]
\caption{Comparison in identification performance on IJB, TinyFace and MegaFace datasets. In the MegaFace benchmark, \textbf{\textit{Iden}} expresses \textbf{\textit{Rank-1}} in identification and \textbf{\textit{Veri}} is TAR@FAR=$10^{-6}$. The similarity score calculation of \cite{terhorst2023qmagface} is not supported by MegaFace devkit, expressed by the notation ``-".\protect\footnotemark[7]}
\label{tab:comparison_iden_IJB}
\begin{adjustbox}{width=\columnwidth,center}
\begin{tabular}{c|l|c;{1pt/1pt}c|c;{1pt/1pt}c|c;{1pt/1pt}c|c;{1pt/1pt}c} 
\hline
\rowcolor[rgb]{0.784,0.784,0.784} {\cellcolor[rgb]{0.784,0.784,0.784}}                                & \multicolumn{1}{c|}{{\cellcolor[rgb]{0.784,0.784,0.784}}}                                  & \multicolumn{2}{c|}{\textbf{IJB-B~}}                                                       & \multicolumn{2}{c|}{\textbf{IJB-C~}}                                                                         & \multicolumn{2}{c|}{\textbf{TinyFace }}                                                    & \multicolumn{2}{c}{\textbf{MegaFace}}                          \\ 
\cline{3-10}
\rowcolor[rgb]{0.784,0.784,0.784} \multirow{-2}{*}{{\cellcolor[rgb]{0.784,0.784,0.784}}\textbf{Year}} & \multicolumn{1}{c|}{\multirow{-2}{*}{{\cellcolor[rgb]{0.784,0.784,0.784}}\textbf{Method}}} & \multicolumn{1}{c|}{\textbf{\textit{Rank-1}}} & \textbf{\textbf{\textit{Rank-}}\textit{5}} & \multicolumn{1}{c|}{\textbf{\textbf{\textit{Rank-}}\textit{1}}} & \textbf{\textbf{\textit{Rank-}}\textit{5}} & \multicolumn{1}{c|}{\textbf{\textbf{\textit{Rank-1}}}} & \textbf{\textbf{\textit{Rank-5}}} & \multicolumn{1}{c|}{\textbf{\textit{Iden}}} & \textbf{Veri}    \\ 
\hline\hline
\textit{2018}                                                                                         & CosFace \cite{wang2018cosface}                                                                                   & \textit{93.554}                               & 95.891                                     & 94.789                                                          & 96.392                                     & 60.381                                                 & 64.941                            & 95.617                                      & 96.260           \\ 
\hdashline[1pt/1pt]
\textit{2019}                                                                                         & ArcFace \cite{deng2019arcface}                                                                                   & 93.204                                        & 95.852                                     & 94.314                                                          & 96.193                                     & 57.672                                                 & 62.607                            & 95.242                                      & 95.895           \\ 
\hdashline[1pt/1pt]
\textit{2020}                                                                                         & MV-Arc-Softmax \cite{wang2020mis}                                                                            & 92.882                                        & 95.813                                     & 94.233                                                          & 96.172                                     & 59.120                                                 & 64.297                            & 94.188                                      & 94.944           \\ 
\hdashline[1pt/1pt]
\textit{2020}                                                                                         & CurricularFace \cite{huang2020curricularface}                                                                            & 93.476                                        & 95.794                                     & 94.794                                                          & 96.366                                     & 60.649                                                 & 64.941                            & 95.637                                      & 96.316           \\ 
\hdashline[1pt/1pt]
\textit{2021}                                                                                         & MagFace \cite{meng2021magface}                                                                                   & 92.911                                        & 95.803                                     & 94.187                                                          & 96.045                                     & 57.833                                                 & 62.527                            & 94.965                                      & 96.052           \\ 
\hdashline[1pt/1pt]
\textit{2021}                                                                                         & VPLFace \cite{deng2021variational}                                                                                   & 93.427                                        & \textit{96.076}                            & 94.666                                                          & 96.448                                     & 57.913                                                 & 62.446                            & 94.937                                      & 95.978           \\ 
\hdashline[1pt/1pt]
\textit{2022}                                                                                         & SphereFace2 \cite{wen2021sphereface2}                                                                               & 92.162                                        & 95.170                                     & 93.610                                                          & 95.743                                     & 57.994                                                 & 63.251                            & 92.595                                      & 94.224           \\ 
\hdashline[1pt/1pt]
2022                                                                                                  & Elastic-Arc \cite{boutros2022elasticface}                                                                               & 92.911                                        & 95.813                                     & 94.391                                                          & 96.203                                     & 57.779                                                 & 62.607                            & 95.674                                      & 96.429           \\ 
\hdashline[1pt/1pt]
2022                                                                                                  & Elastic-Cos \cite{boutros2022elasticface}                                                                               & 93.262                                        & 96.008                                     & 94.473                                                          & 96.315                                     & 60.086                                                 & 64.565                            & 94.989                                      & 95.602           \\ 
\hdashline[1pt/1pt]
\textit{2022}                                                                                         & AdaFace \cite{kim2022adaface}                                                                                   & 93.525                                        & 96.018                                     & 94.814                                                          & \textit{96.524}                            & \textit{60.837}                                        & 64.995                            & 95.888                                      & \uline{96.694}   \\ 
\hdashline[1pt/1pt]
\textit{2023}                                                                                         & QAFace \cite{saadabadi2023quality}                                                                                    & 92.687                                        & 95.764                                     & 94.018                                                          & 96.320                                     & 57.967                                                 & 62.983                            & 93.510                                      & 94.738           \\ 
\hdashline[1pt/1pt]
\textit{2023}                                                                                         & QMagFace \cite{terhorst2023qmagface}                                                                                  & 91.870                                        & 95.560                                     & 93.707                                                          & 96.213                                     & 57.833                                                 & 62.527                            & -                                           & -                \\ 
\hdashline[1pt/1pt]
\textit{2023}                                                                                         & UniFace \cite{zhou2023uniface}                                                                                   & 93.350                                        & 95.979                                     & \textit{94.886}                                                 & 96.402                                     & 60.327                                                 & 64.887                            & \uline{95.961}                              & 96.350           \\ 
\hdashline[1pt/1pt]
\textit{2023}                                                                                         & UniTSFace \cite{jia2023unitsface}                                                                                 & 93.155                                        & 95.891                                     & 94.442                                                          & 96.366                                     & 60.408                                                 & \textit{65.236}                   & 94.409                                      & 95.737           \\ 
\hdashline[1pt/1pt]
\textit{2024}                                                                                         & TopoFR \cite{dan2024topofr}                                                                                    & 93.019                                        & 95.813                                     & 94.243                                                          & 96.289                                     & 59.871                                                 & 64.029                            & \textit{95.913}                             & \textit{96.636}  \\ 
\hline\hline
\rowcolor[rgb]{0.765,0.918,0.773} \textit{Now}                                                        & \textbf{\textit{QCFace-Arc}}                                                               & \uline{94.761}                                & \uline{96.884}                             & \uline{96.157}                                                  & \uline{97.468}                             & \uline{62.634}                                         & \textbf{66.738}                   & \textbf{98.347}                             & \textbf{98.500}  \\ 
\hdashline[1pt/1pt]
\rowcolor[rgb]{0.765,0.918,0.773} \textit{Now}                                                        & \textbf{\textit{QQCFace-Arc}}                                                              & \textbf{94.898}                               & \textbf{96.923}                            & \textbf{96.172}                                                 & \textbf{97.484}                            & \textbf{64.941}                                        & \uline{68.160}                    & -                                           & -                \\
\hline
\end{tabular}
\end{adjustbox}
\end{table}

\subsection{Ablation Study - Frozen Proxy}
\label{subsec:ablation}

In \cref{tab:comparison_hq}, we evaluate the applicability of softmax loss in \textbf{\textit{QCFace}} by six variants of \textbf{\textit{QCFace}} (see the last six rows in \cref{tab:comparison_hq}), demonstrating the adverse impact of incorporating \textit{the misclassified loss} type as $\mathcal{L}_{sm}$. Their verification accuracy approximately drops $4.410\%$, $0.179\%$, $0.284\%$, $4.231\%$, $2.058\%$ and $1.311\%$ on AgeDB-30, CFP-FP, LFW, CALFW, CPLFW and XQLFW, respectively, when comparing with the SoTA on each benchmark dataset. This consequence arises from the requirement to utilize the guidance value, which is only optimally harnessed with immovable proxies. Specifically, the hypersphere planning strategy of MV-Softmax causes the oscillation of proxies due to its misclassified feature emphasis. Indeed, both MV-Softmax and CurricularFace, which are of \textit{misclassified softmax loss}, degrade the optimization process due to proxy oscillation\footnotemark[8]. However, \textbf{\textit{(Q)QCFace-Cur}} results suggest that this negative effect can be partly offset by a larger backbone and dataset.

\footnotetext[7]{These experimental results are done with the backbone IResNet100.}
\footnotetext[8]{The consequence is clearly demonstrated by the experimental results using IResNet18, as shown at Secs. F, G and H of the SM.}

\section{Conclusions \& Limitations}
\label{sec:conclusion_limitation}
\noindent \textbf{Conclusions.} In this paper, we proposed \textbf{\textit{QCFace}} loss to enhance the effectiveness of FRL for boosting face recognition with main contributions. First, we demonstrated the effectiveness of \textbf{\textit{hard margin}}, a novel \textit{margin strategy} on recognizability-based hypersphere planning. Second, this is the first time to introduce the \textbf{\textit{guidance value}} for MVP to guarantee a positive correlation between magnitude value and confidence score in FRL. Our extensive experiments reflect the superiority of \textbf{\textit{QCFace}} in recognizability representation and recognition ability over the SoTAs.

\noindent \textbf{Limitations.} Based on Property \textcolor{red}{3}, unexpected outcomes on low-recognizability images stem from their overlap with mislabeled samples in the low-magnitude region, while most data lie in the high-magnitude region due to higher recognizability. Therefore, gradients of low-recognizability samples fail to surpass those of mislabeled ones, hindering their optimization and leading to unstable convergence due to their large gradient magnitudes (see Lemma \textcolor{red}{1}). On the other hand, \textbf{\textit{QCFace}} effectively assigns low values of magnitude to these problematic samples, enabling their identification and exclusion.

\vspace{1mm}
\noindent \textbf{Acknowledgements.} We acknowledge Ho Chi Minh City University of Technology (HCMUT), VNU-HCM for supporting this study. We are also grateful to the anonymous reviewers for their thoughtful comments and suggestions.

{
    \small
    \bibliographystyle{ieeenat_fullname}
    \bibliography{references}
}

\appendix

\begin{center}
    \Large \textbf{Supplementary Material}
\end{center}

\section{The previous class center loss analysis}

\begin{figure*}[!t]
  \centering
    \includegraphics[width=0.9\linewidth]{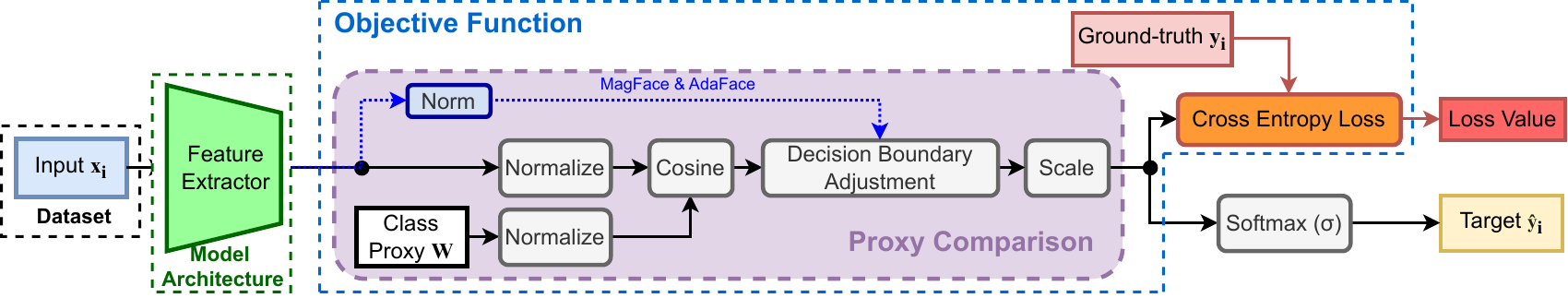}
    \caption{The training pipeline of face recognition models with class center loss.}
    \label{fig:appendix_training_pipeline}
\end{figure*} 

\cref{fig:appendix_training_pipeline} clearly illustrates about four components of a training pipeline with \textit{class-center-loss-based} methods, where we can target improvements to enhance face recognition performance, specifically, \textit{model architecture}, \textit{training dataset}, \textit{training strategy} and \textit{objective (loss) function}--the target of the following analysis. Based on the criteria mentioned in Sec. 1 in our main paper, the loss function shows its most potential approach. Recall that the general formula of \textit{the class center loss} is shown as follows.
\begin{equation}
    \resizebox{0.48\textwidth}{!}{$
        \begin{array}{l}
            \hspace{0.06cm} \mathcal{L}_i = \underbrace{-log\frac{e^{s.F(\sM, \theta_{\vw_{y_i},\vz_i})}}{e^{s.F(\sM, \theta_{\vw_{y_i},\vz_i})} + \sum_{j \neq y_i}^{C}e^{s.N(t, \theta_{\vw_j,\vz_i})}}}_{{\mathcal{L}_{sm}}_i} + \lambda_g.\underbrace{g(\vz_i)}_{{\mathcal{L}_{reg}}_i}
        \end{array}
    $}
  \label{eq:supp_gen_cen_loss}
\end{equation}
where $y_i$ is an identity label of image $i$, and
\begin{equation}
    \resizebox{0.433\textwidth}{!}{$
        F(\sM, \theta) = \text{cos}(m_1\theta+m_2)-m_3 ~~ \text{where} ~~ \sM = \{m_1, m_2, m_3\}
    $}
    \label{eq:supp_add_margin}
\end{equation}
where $m_1$, $m_2$ and $m_3$ express three margin types to adjust the decision boundary in $\mathcal{L}_{sm}$, \cref{tab:appendix_comparison_classcenterloss} details margin components ($\sM = \{m_1, m_2, m_3\}$) in positive modulation function (\textit{i.e.}, $F$), negative modulation function (\textit{i.e.}, $N$) and regularization loss (\textit{i.e.}, $\mathcal{L}_{reg}$) of previous studies (\textit{e.g.}, \cite{schroff2015facenet, liu2017sphereface, wang2018cosface, deng2019arcface, wang2020mis, huang2020curricularface}).

\begin{table*}[!t]
\centering
\caption{The detail of each component in \cref{eq:supp_gen_cen_loss} of the previous \textit{class center loss} solutions. $m$ expresses a constant coefficient, $u_m$, $l_m$, $u_a$ and $l_a$ are respectively upper and lower bounds of additive angular margin and feature magnitude $\|\vz\|$, $\widehat{\|\vz\|}$ is batch normalized value of $\|\vz\|$, and $t$ expresses mis-classified coefficient.}
\label{tab:appendix_comparison_classcenterloss}
\begin{adjustbox}{width=1.6\columnwidth,center}
    \begin{tabular}{l;{1pt/1pt}c;{1pt/1pt}c;{1pt/1pt}c;{1pt/1pt}c;{1pt/1pt}c} 
    \hline
    \rowcolor[rgb]{0.784,0.784,0.784} \multicolumn{1}{c|}{{\cellcolor[rgb]{0.784,0.784,0.784}}} & \multicolumn{3}{c|}{\textbf{Additive function components}} & \multicolumn{1}{c|}{{\cellcolor[rgb]{0.784,0.784,0.784}}} & \multicolumn{1}{c}{{\cellcolor[rgb]{0.784,0.784,0.784}}} \\ 
    \cline{2-4}
    \rowcolor[rgb]{0.784,0.784,0.784} \multicolumn{1}{c|}{\multirow{-2}{*}{{\cellcolor[rgb]{0.784,0.784,0.784}}\textbf{Method }}} & \multicolumn{1}{c|}{\textbf{$m_1$}} & \multicolumn{1}{c|}{\textbf{$m_2$}} & \multicolumn{1}{c|}{\textbf{$m_3$}} & \multicolumn{1}{c|}{\multirow{-2}{*}{{\cellcolor[rgb]{0.784,0.784,0.784}}\textbf{$N$}}} & \multicolumn{1}{c}{\multirow{-2}{*}{{\cellcolor[rgb]{0.784,0.784,0.784}}$g$}} \\ 
    \hline\hline
    \textbf{\textit{SphereFace}} \cite{liu2017sphereface}             & $m$ & 0                                                  & 0                               & $cos(\theta_j)$                   & 0                                  \\ 
    \hdashline[1pt/1pt]
    \textbf{\textit{CosFace}} \cite{wang2018cosface}                  & 0   & 0                                                  & $m$                             & $cos(\theta_j)$                   & 0                                  \\ 
    \hdashline[1pt/1pt]
    \textbf{\textit{ArcFace}} \cite{deng2019arcface}                  & 0   & $m$                                                & 0                               & $cos(\theta_j)$                   & 0                                  \\ 
    \hdashline[1pt/1pt]
    \textbf{\textit{MV-Arc-Softmax}} \cite{wang2020mis}               & 0   & $m$                                                & 0                               & $t.cos(\theta_j)+t-1$             & 0                                  \\ 
    \hdashline[1pt/1pt]
    \textbf{\textit{CurricularFace}} \cite{huang2020curricularface}   & 0   & $m$                                                & 0                               & $(t+cos(\theta_j)).cos(\theta_j)$ & 0                                  \\ 
    \hdashline[1pt/1pt]
    \textbf{\textit{MagFace}} \cite{meng2021magface}                  & 0   & $\frac{u_m-l_m}{u_a-l_a}.(\|\vz\|-l_a)+l_m$ & 0                               & $cos(\theta_j)$                   & $\frac{1}{\|\vz\|}+\frac{\|\vz\|}{u_a^2}$ \\ 
    \hdashline[1pt/1pt]
    \textbf{\textit{AdaFace}} \cite{kim2022adaface}                   & 0   & $-m.\widehat{\|\vz\|}$                      & $m.(\widehat{\|\vz\|}+1)$ & $cos(\theta_j)$                   & 0                                  \\ 
    \hdashline[1pt/1pt]
    \rowcolor[rgb]{0.765,0.918,0.773} \textbf{\textit{QCFace (Ours)}} & 0    & $m$                                               & 0                               & $cos(\theta_j)$                   & $g_{rc}(\|\vz\|, p_{rc})$ \\
    \hline
    \end{tabular}
\end{adjustbox}
\end{table*}

\section{Technical Terminology Definition}
The following definitions are common in the face recognition community and are mentioned in our main paper. 
\begin{itemize}
    \item \textbf{\textit{Knowledge accumulation (aka learning capacity)}} is a concept used to express the \textbf{model} property. It means how much knowledge (useful information) a deep learning model can extract (learn) from the data. 
    \item \textbf{\textit{Knowledge capacity}} is used to express the \textbf{data} quality. Specifically, it describes the degree of diversity and effect of noise (\textit{e.g.,} mislabeled samples) of the datasets.
    \item \textbf{\textit{Image recognizability}} is a human-perceptual variable used to express the \textbf{image} property, which quantifies the ease of recognizing an image. Specifically, it is a measurement of how easy an image can be recognized.
    \item \textbf{\textit{Recognizability capacity}} is used to express the usefulness of an embedded feature in representing a face image. 
    \item \textbf{\textit{Mislabeled sample}} refers to a sample with an incorrect label, which is a common problem in large-scale datasets.
    \item \textbf{\textit{Hypersphere}} is an embedded feature space, which is technically defined by choosing the number of dimensions of the embedded feature. Higher values can represent more data diversity, but they are easily prone to overfitting and cause high computational cost. The suitable value is frequently chosen based on the dataset scale and by conducting experiments. The value $512$ is the common embedded dimension in most well-known studies (\textit{e.g.}, \cite{schroff2015facenet, liu2017sphereface, wang2018cosface, deng2019arcface, wang2020mis, huang2020curricularface}).
\end{itemize}

\section{Lemma \& Property Proofs}
Recall that the formulas of the gradients affecting the actual-class proxy, the non-actual (misclassified) -class proxy, feature direction and magnitude are respectively shown as follows.
\begin{equation}
    \resizebox{0.4\textwidth}{!}{$
        \vg^i_{ac} := s . (P^{(i)}_{y_i}-1) . \frac{\partial F(\sM, \theta_{\vw_{y_i},\vz_i})}{\partial \text{cos}(\theta_{\vw_{y_i},\vz_i})}.\frac{\partial \text{cos}(\theta_{\vw_{y_i},\vz_i})}{\partial \vw_{y_i}}
    $}
    \label{eq:supp_gac}
\end{equation}
\begin{equation}
    \resizebox{0.43\textwidth}{!}{$
        \vg^i_{mc} := s.P^{(i)}_{k}.\frac{\partial N(t, \theta_{\vw_k,\vz_i})}{\partial \text{cos}(\theta_{\vw_k,\vz_i})}.\frac{\partial \text{cos}(\theta_{\vw_k,\vz_i})}{\partial \vw_k} ~~ \text{where} ~~ k \neq y_i
    $}
    \label{eq:supp_gmc}
\end{equation}
\begin{equation}
    \resizebox{0.43\textwidth}{!}{$
        \vg^i_\theta := s . \sum^{C}_{k=1} \left[P^{(i)}_k-\1_\mathrm{k=y_i}\right] . \frac{\partial \textit{Fnc}^{(i)}_k}{\partial \text{cos}(\theta_{\vw_k,\vz_i})}.\frac{\partial \text{cos}(\theta_{\vw_k,\vz_i})}{\partial \vz_i}
    $}
    \label{eq:supp_gtheta}
\end{equation}
\begin{equation}
    \resizebox{0.43\textwidth}{!}{$
        \vg^i_{\|\vz\|} := \underbrace{s . \sum^{C}_{k=1} \left[P^{(i)}_k-\1_\mathrm{k=y_i}\right] . \frac{\partial \textit{Fnc}^{(i)}_k}{\partial \|\vz_i\|}.\frac{\partial \|\vz_i\|}{\partial \vz_i}}_{\vg^i_{\|\vz\|^-}} + \underbrace{\frac{\partial {\mathcal{L}_{reg}}_i}{\partial \|\vz_i\|}.\frac{\partial \|\vz_i\|}{\partial \vz_i}}_{\vg^i_{\|\vz\|^+}}
    $}
    \label{eq:supp_gz}
\end{equation}
where 
\begin{equation}
    \resizebox{0.48\textwidth}{!}{$
        \textit{Fnc}^i_k = 
            \begin{cases}
                F(\sM, \theta_{\vw_{y_i},\vz_i}) & k = y_i \\
                N(t, \theta_{\vw_k,\vz_i}) & k \neq y_i
            \end{cases}
        ~~ \text{and} ~~
        \1_\mathrm{k=y_i} = 
        \begin{cases}
            1 & k = y_i \\
            0 & k \neq y_i
        \end{cases}
    $}
    \label{eq:supp_fnc}
\end{equation}
The probability output at class $k$ (\textit{i.e.}, $P^{(i)}_k$) and derivatives of $\textit{Fnc}^i_k$ with respect to direction (\textit{i.e.}, $\text{cos}(\theta_{\vw_k,\vz_i})$) and magnitude (\textit{i.e.}, $\|\vz_i\|$) are calculated as follows.
\begin{equation}
    \resizebox{0.15\textwidth}{!}{$
        \begin{aligned}
            P^{(i)}_k = \frac{e^{\textit{Fnc}^{(i)}_k}}{\sum_{j=1}^C e^{\textit{Fnc}^{(i)}_j}}
        \end{aligned}
    $}
    \label{eq:met_class_prob}
\end{equation}
\begin{equation}
    \resizebox{0.48\textwidth}{!}{$
        \begin{array}{ll}
            & \frac{\partial \textit{Fnc}^{(i)}_k}{\partial \text{cos}(\theta_{\vw_k,\vz_i})} = 
            \begin{cases}
                s.\frac{\partial \text{cos}(m_1.\theta_{\vw_k,\vz_i})}{\partial \text{cos}(\theta_{\vw_k,\vz_i})} \times f(\theta_{\vw_k,\vz_i}, \|\vz_i\|) & k = y_i \vspace{1mm}\\
                s.\frac{\partial N(t, \text{cos}(\theta_{\vw_k,\vz_i}))}{\partial \text{cos}(\theta_{\vw_k,\vz_i})} & k \neq y_i
            \end{cases} \\ 
            & \text{where} \vspace{1mm}\\
            & f(\theta_{\vw_k,\vz_i}, \|\vz_i\|) = \text{cos}(m_2(\|\vz_i\|))+\text{cot}(m_1.\theta_{\vw_k,\vz_i}).\text{sin}(m_2(\|\vz_i\|))
        \end{array}
    $}
    \label{eq:met_cos_derivative}
\end{equation}

\begin{equation}
    \resizebox{0.43\textwidth}{!}{$
        \begin{aligned}
            \frac{\partial \textit{Fnc}^{(i)}_k}{\partial \|\vz_i\|} = 
            \begin{cases}
                -s.\text{sin}(m_1.\theta_{\vw_k,\vz_i}+m_2(\|\vz_i\|)).\frac{\partial m_2(\mathbf{\|\vz_i\|})}{\partial \|\vz_i\|} & k = y_i \vspace{1mm}\\
                \hspace{3cm}0 & k \neq y_i
            \end{cases}
        \end{aligned}
    $}
    \label{eq:met_mag_derivative}
\end{equation}

\noindent \textbf{Lemma \textcolor{red}{1}.} \textit{Let $\vz_1, \vz_2 \in \mathbb{R}^d$ where $d$ denotes the embedding dimension. If $\|\vz_1\| > \|\vz_2\|$, then $\|\vg_\theta(\vz_1)\| < \|\vg_\theta(\vz_2)\|$.}

\noindent \textbf{Proof.} Let $f(\|\vz_i\|) = \text{cos}(\theta_{\vw_k,\vz_i})$. Subsequently, we have
\begin{equation*}
    \resizebox{0.47\textwidth}{!}{$
        \begin{aligned}
            f(\|\vz_i\|) = \frac{\vz_i^\top \vw_k}{\|\vz_i\| \|\vw_k\|} = \frac{1}{\|\vw_k\|}.\frac{\vz_i^\top\vw_k}{\|\vz_i\|} & = c . \frac{\vz_i^\top\vw_k}{\|\vz_i\|} \\
            & \text{where} ~~ c = \frac{1}{\|\vw_k\|} \text{ is constant.}
        \end{aligned}
    $}
\end{equation*}
Then, we calculate the derivative of $\text{cos}(\theta_{\vw_k,\vz_i})$ with respect to $\|\vz_i\|$ as follows.
\begin{equation*}
    \resizebox{0.47\textwidth}{!}{$
        \begin{aligned}
            \frac{\partial \text{cos}(\theta_{\vw_k,\vz_i})}{\partial \vz_i} = \nabla_{\vz_i}f(\vz_i) & = c.\nabla_{\vz_i}\left(\frac{\vz_i^\top\vw_k}{\|\vz_i\|}\right) \\ 
            & = c.\frac{\|\vz_i\| \vw_k - (\vz_i^\top\vw_k) . \frac{\vz_i}{\|\vz_i\|}}{\|\vz_i\|^2} \\
            & = c.\frac{\vw_k - \frac{\vz_i^\top \vw_k}{\|\vz_i\|^2}.\vz_i}{\|\vz_i\|}
        \end{aligned}
    $}
\end{equation*}
Now, we compute the norm of this gradient as follows to compare the magnitudes
\begin{equation*}
    \resizebox{0.3\textwidth}{!}{$
        \begin{aligned}
            \|\nabla_{\vz_i}f(\vz_i)\| = c.\frac{\|\vw_k - \frac{\vz_i^\top \vw_k}{\|\vz_i\|^2}.\vz_i\|}{\|\vz_i\|}
        \end{aligned}
    $}
\end{equation*}
Let us define the projection of $\vw_k$ onto $\vz_i$ as follows.
\begin{equation*}
    \resizebox{0.2\textwidth}{!}{$
        \begin{aligned}
            \text{proj}_{\vz_i}(\vw_k) = \frac{\vz_i^\top \vw_k}{\|\vz_i\|^2}.\vz_i
        \end{aligned}
    $}
\end{equation*}
Then the numerator becomes $\vw_k - \text{proj}_{\vz_i}(\vw_k)$, \textit{i.e.} the component of $\vw_k$ orthogonal to $\vz_i$. Thus, we have
\begin{equation}
    \resizebox{0.3\textwidth}{!}{$
        \begin{aligned}
            \|\nabla_{\vz_i}f(\vz_i)\| = c.\frac{\|{\vw_k}_\perp\|}{\|\vz_i\|} = \frac{1}{\|\vw_k\|} . \frac{\|{\vw_k}_\perp\|}{\|\vz_i\|}
        \end{aligned}
    $}
    \label{eq:supp_final_lemma1}
\end{equation}

\begin{figure}[!t]
  \centering
   \includegraphics[width=0.9\linewidth]{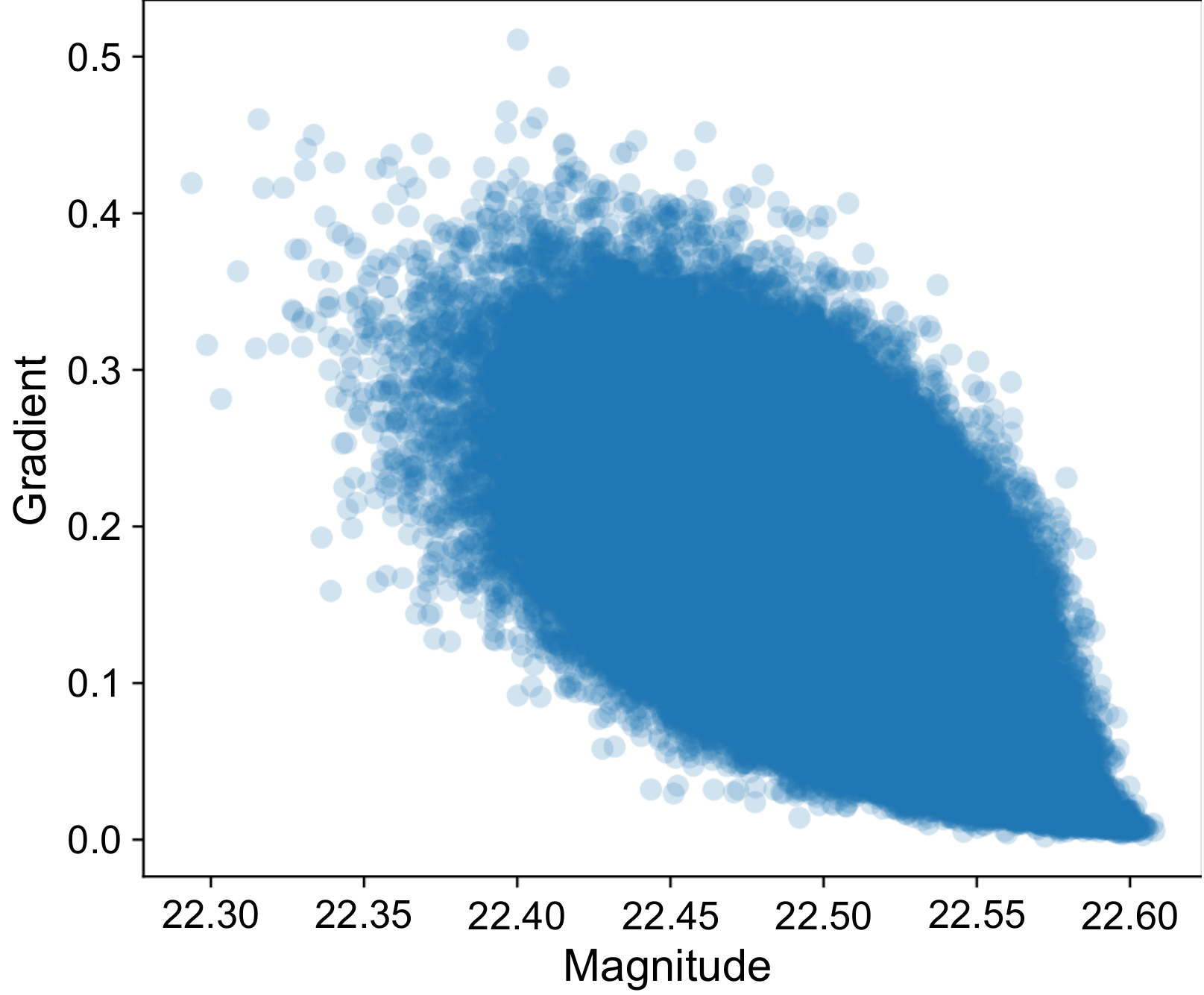}
   \caption{The plot of negative correlation between magnitude value of embedded features and mean values of gradient magnitude in an FR model trained with IResNet18 and CASIA-WebFace.}
   \label{fig:appendix_lemma1}
\end{figure}

\noindent Based on \eqref{eq:supp_final_lemma1}, we derive that $\|\nabla_{\vz_i}f(\vz_i)\| \propto \frac{1}{\|\vz_i\|}$. Besides, based on \eqref{eq:supp_gtheta}, we also have $\|\vg_\theta\| \propto \|\nabla_{\vz_i}f(\vz_i)\|$, thus, $\|\vg_\theta\| \propto \frac{1}{\|\vz_i\|}$. Therefore, we can derive that if $\|\vz_1\| > \|\vz_2\|$, then $\|\vg_\theta(\vz_1)\| < \|\vg_\theta(\vz_2)\|$, and Lemma \textcolor{red}{1} is completely demonstrated. The correctness of Lemma \textcolor{red}{1} is also validated by \cref{fig:appendix_lemma1} experimentally.

\vspace{1mm}

\noindent \textbf{Lemma \textcolor{red}{2}.} \textit{Suppose that $a_1, a_2 \in \mathbb{R}$ are two independent variables, and a function $f(a_1, a_2): \mathbb{R}^2 \rightarrow \mathbb{R}$. If $f$ can be described as a form of $f(a_1, a_2) = f_1(a_1) + f_2(a_2)$ where $f_1, f_2: \mathbb{R} \rightarrow \mathbb{R}$, then the mutual overlapping gradient calculated from the derivative of $f$ does not exist.}

\noindent \textbf{Proof.} We first calculate the derivatives with respect to $a_1$ and $a_2$ as follows.
\begin{equation*}
    \resizebox{0.3\textwidth}{!}{$
        \begin{cases}
            \frac{\partial f}{\partial a_1} = \frac{\partial f_1}{\partial a_1} + \frac{\partial f_2}{\partial a_1} = \frac{\partial f_1}{\partial a_1} & := \vg_{a_1} \vspace{2mm}\\
            \frac{\partial f}{\partial a_2} = \frac{\partial f_1}{\partial a_2} + \frac{\partial f_2}{\partial a_2} = \frac{\partial f_2}{\partial a_2} & := \vg_{a_2}
        \end{cases}
    $}
\end{equation*}
Applying gradient descent for $a_1$ and $a_2$ to minimize $f$, the optimization process is shown as follows.
\begin{equation}
    \resizebox{0.2\textwidth}{!}{$
        \begin{cases}
            a_1^{new} \gets \mathcal{O}_1(a_1^{old}, \vg_{a_1}) \vspace{2mm}\\
            a_2^{new} \gets \mathcal{O}_2(a_2^{old}, \vg_{a_2})
        \end{cases}
    $}
    \label{eq:supp_lemma2}
\end{equation}
where $\mathcal{O}_1$ and $\mathcal{O}_2$ are optimization (update) operations (\textit{e.g.}, SGD, Adam, etc.) for two variables, $a_1$ and $a_2$. We can easily observe that
\begin{equation}
    \resizebox{0.4\textwidth}{!}{$
        \begin{cases}
             \frac{\partial \vg_{a_1}}{\partial a_2} = \frac{\partial^2 f_1}{\partial a_1 \partial a_2} = 0  ~~ \forall a_2 \in \mathbb{R} & \implies \frac{\partial \mathcal{O}_1}{\partial a_2} = 0 \vspace{2mm}\\
            \frac{\partial \vg_{a_2}}{\partial a_1} = \frac{\partial^2 f_2}{\partial a_2 \partial a_1} = 0 ~~ \forall a_1 \in \mathbb{R} & \implies \frac{\partial \mathcal{O}_2}{\partial a_1} = 0
        \end{cases}
    $}
    \label{eq:supp_final_lemma2}
\end{equation}
Therefore, based on \eqref{eq:supp_final_lemma2}, the optimization processes of two variables (\textit{i.e.}, $a_1$ and $a_2$) are not affected by each other's previous values, and Lemma \textcolor{red}{2} is completely demonstrated.

\vspace{1mm}

\noindent \textbf{Property \textcolor{red}{1}.}  \textit{In \cref{eq:supp_gen_cen_loss}, if $m_2$ is a strictly increasing convex function of $\|\vz_i\|$ (\textit{i.e.}, $m_2(\|\vz_i\|)$), then the norm $\|\vz_i\|$ influences the gradient with respect to $\theta$ (\textit{i.e.}, $\vg_\theta$), while $\theta$ in turn affects the gradient with respect to $\|\vz_i\|$ (\textit{i.e.}, $\vg_{\|\vz_i\|}$). This manifests as a mutual overlapping gradient.}

\noindent \textbf{Proof.} Based on \eqref{eq:supp_gtheta} and \eqref{eq:supp_gz}, the component in the loss function that may lead to a mutual overlapping gradient problem between $\vg^i_\theta$ and $\vg^i_{\|\vz\|}$, is the margin function $F(\sM, \theta)$ (see \cref{eq:supp_add_margin}). Specifically, if $m_2$ is a strictly increasing convex function of $\|\vz_i\|$ (\textit{i.e.}, $m_2(\|\vz_i\|)$), based on \eqref{eq:met_cos_derivative} and \eqref{eq:met_mag_derivative}, the gradients influencing the updates of two independent attributes of the embedded feature, \textit{i.e.} direction ($\text{cos}(\theta_{\vw_k,\vz_i})$) and magnitude ($\|\vz_i\|$), depend on each other's previous values. In contrast to the findings of Lemma \textcolor{red}{2}, we have
\begin{equation*}
    \resizebox{0.45\textwidth}{!}{$
        \begin{cases}
            \exists ~ \|\vz_i\| \in \mathbb{R^+} \text{ such that } \frac{\partial \vg^i_\theta}{\partial \|\vz_i\|} \neq 0 & \implies \frac{\partial \mathcal{O}_\theta}{\partial \|\vz_i\|} \neq 0 \vspace{2mm}\\
            \exists ~ \theta_{\vw_k,\vz_i} \in \left[-\frac{\pi}{2}, \frac{\pi}{2}\right] \text{ such that } \frac{\partial \vg^i_{\|\vz\|}}{\partial \theta_{\vw_k,\vz_i}} \neq 0 & \implies \frac{\partial \mathcal{O}_{\|\vz\|}}{\partial \theta_{\vw_k,\vz_i}} \neq 0
        \end{cases}
    $}
\end{equation*}
where $\mathcal{O}_\theta$ and $\mathcal{O}_{\|\vz\|}$ are update functions for direction and magnitude of embedded feature. Consequently, the mutual overlapping gradient exists in this situation.

\vspace{1mm}

Let us recall the formulas of regularization loss ($\mathcal{L}_{reg}$) and \textbf{\textit{guidance value}} ($p_d$) of \textbf{\textit{QCFace}} for the proofs of the following properties.
\begin{equation}
    \resizebox{0.42\textwidth}{!}{$
        {\mathcal{L}_{reg}}_i = k.{p_d}_i. \left (\frac{1}{\|\vz_i\|}+\frac{\|\vz_i\|}{u_a^2} \right ) + (1-{p_d}_i).\left (\frac{1}{\|\vz_i\|}+\frac{\|\vz_i\|}{l_a^2} \right ) - b
    $}
    \label{eq:supp_qcface_regular}
\end{equation}
\begin{equation}
    \resizebox{0.35\textwidth}{!}{$
        {p_d}_i = \frac{e^{s.cos(\theta_{\vw_{y_i}, \vz_i})}}{e^{s.cos(\theta_{\vw_{y_i}, \vz_i})} + \sum_{j \neq y_i}^{n}e^{s.cos(\theta_{\mathbf{w_j}, \vz_i})}}
    $}
    \label{eq:supp_qcface_guidance}
\end{equation}

\noindent \textbf{Property \textcolor{red}{2}.} \textit{In \cref{eq:supp_qcface_guidance}, $p_d$ is bounded in $\left[0, 1\right]$.}

\noindent \textbf{Proof.} We rewrite \cref{eq:supp_qcface_guidance} as ${p_d}_i = \frac{e^{v_{y_i}}}{\sum_{j=1}^{n}e^{v_j}}$. Accordingly, we have
\begin{equation*}
    \resizebox{0.35\textwidth}{!}{$
        \begin{cases}
            e^{v_t} > 0 ~~ \forall v_t \in \mathbb{R} & \implies {p_d}_i > 0 \vspace{2mm}\\
            \sum_{j=1}^{n}e^{v_j} \geq e^{v_t} ~~ \forall v_t \in \mathbb{R} & \implies {p_d}_i \leq 1
        \end{cases}
    $}
\end{equation*}
Therefore, ${p_d}_i$ is a bounded value in $[0, 1]$.

\vspace{1mm}

\noindent \textbf{Property \textcolor{red}{3}.} \textit{Suppose that $\vz_1, \vz_2 \in \mathbb{R}^d$ where $d$ denotes the embedding dimension, and let their corresponding guidance values (\textit{i.e.}, ${p_d}_1, {p_d}_2 \in [0, 1]$) be computed by \cref{eq:supp_qcface_guidance}. If ${p_d}_1 > {p_d}_2$, then $\|\vg_\theta(\vz_1)\| < \|\vg_\theta(\vz_2)\| ~ \forall \vz_1, \vz_2$.}

\noindent \textbf{Proof.} We easily observe that both $p_d$ and $P^{(i)}_k$ are bounded in $[0, 1]$ (see Property \textcolor{red}{1}). Moreover, $p_d$ and $\|\sum^{n}_{k=1} \left[P^{(i)}_k-\mathbbm{1}(k=y_i)\right]\|$ exhibit opposite trends. Thus, we can derive as 
\begin{equation*}
    \resizebox{0.4\textwidth}{!}{$
        \begin{aligned}
            & {p_d}_1 > {p_d}_2 \implies \\
            & \left|\sum^{n}_{k=1} \left[P^{(i)}_k-\mathbbm{1}(k=y_i)\right]\right|_{\vz=\vz_1} < \left|\sum^{n}_{k=1} \left[P^{(i)}_k-\mathbbm{1}(k=y_i)\right]\right|_{\vz=\vz_2}
        \end{aligned}
    $}
\end{equation*}
On the other hand, $\|\vg_\theta\| \propto \left|\sum^{n}_{k=1} \left[P^{(i)}_k-\mathbbm{1}(k=y_i)\right]\right|$ when $\theta_{\vw_{y_i},\vz_i} \rightarrow 0$ and $\theta_{\vw_k,\vz_i} \rightarrow \pm\frac{\pi}{2} ~~ \forall k \neq y_i$. Consequently, if ${p_d}_1 > {p_d}_2$, then $\|\vg_\theta(\vz_1)\| < \|\vg_\theta(\vz_2)\| ~ \forall \vz_1, \vz_2$.


\section{Constraints for QCFace}
\label{sec:appendix_constraint}
In this section, we establish and prove three constraints for the convergence of optimization and hyperparameter selection in \textbf{\textit{QCFace}}. First, we demonstrate the convergence ability of \textbf{\textit{QCFace}} (see \cref{subsec:proof_convergence}). Second, we establish a condition to ensure the linearity of magnitude encoding (see \cref{subsec:proof_balanc}). Last, we add a constraint for bias value to observe whether $\mathcal{L}_{reg}$ converges (see \cref{subsec:proof_loss_track}).

\subsection{Constraint for convergence}
\label{subsec:proof_convergence}
In \cref{eq:supp_qcface_regular}, $l_a$ and $u_a$ are respectively set to $1$ and $100$, which are lower and upper bounds of the recognizability score of a face image (\textit{i.e.,} $score \in \left[1, 100\right]$), according to the work by Chen \textit{et al.} \cite{chen2014face}. To certainly reach the expected convergence point in the optimization process, $\mathcal{L}_{reg}$ is required to have the following three properties for all images: 
\begin{itemize}
    \item $\mathcal{L}_{reg}$ is a strictly convex downward function.
    \item $\mathcal{L}_{reg}$ has an optimal unique value in range $\left[l_a, u_a\right]$ (\textit{i.e.,} $\|\vz\|^* \in \left[l_a, u_a\right]$).
    \item The optimal magnitude value ($\|\vz\|^*$) of $\mathcal{L}_{reg}$ for each value $p_d$ is monotonically increasing as the recognizability \textbf{\textit{guidance value}} ($p_d$) increases.
\end{itemize}

For the demonstrations, we first introduce and prove three lemmas.

\noindent \textbf{Lemma \textcolor{red}{3}.} \textit{If $k>0$ and $p_d \in \left[0, 1\right]$, then $\mathcal{L}_{reg}$ is a strictly convex downward function.}

\noindent \textbf{Proof.}
We first calculate the first and second order derivatives of the loss $\mathcal{L}_{reg}$ with respect to $\|\vz\|$.
\begin{equation*}
     \resizebox{0.5\textwidth}{!}{$
        \begin{array}{cl}
            \frac{\partial \mathcal{L}_{reg}}{\partial \|\vz\|} & = k.p_d.\left(-\frac{1}{\|\vz\|^2} + \frac{1}{u_a^2}\right) + (1-p_d).\left(-\frac{1}{\|\vz\|^2} + \frac{1}{l_a^2}\right) \vspace{1mm}\\
            \implies \frac{\partial^2 \mathcal{L}_{reg}}{(\partial \|\vz\|)^2} & = \left[1+(k-1).p_d\right].\frac{2}{\|\vz\|^3}
        \end{array}
    $}
\end{equation*}
By analyzing each component in $\frac{\partial^2 \mathcal{L}_{reg}}{\partial \|\vz\|^2}$, we have 
\begin{equation*}
    \resizebox{0.3\textwidth}{!}{$
        \|\vz\| \geq 0 \hspace{3mm} \forall \vz \in \mathbb{R}^d \implies \frac{2}{\|\vz\|^3} > 0
    $}
\end{equation*}
and
\begin{equation*}
    \resizebox{0.3\textwidth}{!}{$
        1 + (k-1).p_d > 0 ~~~ \forall k > 0, ~~ \forall p_d \in \left[0, 1\right]
    $}
\end{equation*}
Subsequently, $\frac{\partial^2 \mathcal{L}_{reg}}{(\partial \|\vz\|)^2} > 0$, thus, $\mathcal{L}_{reg}$ is a strictly convex downward function.

\vspace{1mm}

\noindent \textbf{Lemma \textcolor{red}{4}.} \textit{If $l_a<u_a$ and $k>0$, then $\mathcal{L}_{reg}$ has an optimal unique magnitude value in range $\left[l_a, u_a\right]$ (\textit{i.e.,} $\|\vz\|^* \in \left[l_a, u_a\right]$).}

\noindent \textbf{Proof.}
According to Lemma \textcolor{red}{1}, $\mathcal{L}_{reg}$ is a strictly convex downward function if $k>0$, we have
\begin{equation*}
    \resizebox{0.35\textwidth}{!}{$
        \begin{aligned}
            \frac{\partial \mathcal{L}_{reg}}{\partial \|\vz\|^i} > \frac{\partial \mathcal{L}_{reg}}{\partial \|\vz\|^{ii}} \hspace{2mm}&\forall \|\vz\|^i, \|\vz\|^{ii} \in \left[l_a, u_a\right] \\
            &if\hspace{1mm} u_a \geq \|\vz\|^i > \|\vz\|^{ii} \geq l_a
        \end{aligned}
    $}
\end{equation*}
Therefore, if there exists an optimal solution $\|\vz\|^* \in \left[l_a, u_a\right]$, then it is an unique solution.

\noindent Let us consider the value of derivatives of $\mathcal{L}_{reg}$ of $l_a$, $u_a$
\begin{equation*}
    \resizebox{0.3\textwidth}{!}{$
        \begin{cases}
            \frac{\partial \mathcal{L}_{reg}}{\partial \|\vz\|}(u_a) &= (1-p_d).\left(-\frac{1}{u_a^2}+\frac{1}{l_a^2}\right) \vspace{1mm}\\
            \frac{\partial \mathcal{L}_{reg}}{\partial \|\vz\|}(l_a) &= k.p_d.\left(-\frac{1}{l_a^2}+\frac{1}{u_a^2}\right)
        \end{cases}
    $}
\end{equation*}
We also have $u_a > l_a$ since they are upper and lower bounds of $\|\vz\|$, respectively. Thus, we can derive 
\begin{equation*}
    \resizebox{0.15\textwidth}{!}{$
        \begin{cases}
            \frac{\partial \mathcal{L}_{reg}}{\partial \|\vz\|}(u_a) &> 0 \vspace{1mm}\\
            \frac{\partial \mathcal{L}_{reg}}{\partial \|\vz\|}(l_a) &< 0
        \end{cases}
    $}
\end{equation*}

As $\frac{\partial \mathcal{L}_{reg}}{\partial \|\vz\|}$ is monotonically and strictly increasing, there must exist an optimal unique value $\|\vz\|^* \in \left[l_a, u_a\right]$ which has a 0 derivative.

\vspace{1mm}

\noindent \textbf{Lemma \textcolor{red}{5}.} \textit{If $k>0$ and $l_a<u_a$, then the optimal magnitude value (\textit{i.e.}, $\|\vz\|^*$) of $\mathcal{L}_{reg}$ for each $p_d$ value is monotonically increasing as $p_d$ increases.}

\noindent \textbf{Proof.}
Assume that $k$, $l_a$ and $u_a$ have already been chosen. To find the formula of the optimal magnitude value $\|\vz\|^*$ depending on $p_d$, we resolve $\frac{\partial \mathcal{L}_{reg}}{\partial \|\vz\|}=0$
\begin{equation}
    \resizebox{0.35\textwidth}{!}{$
        \begin{array}{lccl}
            \implies &\|\vz\|^* &= &\sqrt{\frac{\left[1+(k-1).p_d\right].u_a^2.l_a^2}{u_a^2+(k.l_a^2-u_a^2).p_d}} \\
                     & &  &\resizebox{0.12\textwidth}{!}{$(\|\vz\|^* > 0 ~~~ \forall z \in \mathbb{R}^d)$} \\
                     & &= &\resizebox{0.04\textwidth}{!}{$f_{\vz}(p_d)$}
        \end{array}
    $}
    \label{eq:appendix_z}
\end{equation}
To make $\|\vz\|^*$ be monotonically increasing as $p_d$ increases, we have to satisfy
\begin{equation}
    \resizebox{0.25\textwidth}{!}{$
        \begin{array}{lcc}
                 &\frac{\partial f_{\vz}}{\partial p_d} &>0 \vspace{1mm}\\
            \iff &k.(u_a^2-l_a^2)                 &>0
        \end{array}
    $}
    \label{eq:appendix_lemma3}
\end{equation}
We also have $u_a^2-l_a^2>0$ since $u_a>l_a$, and $k>0$ by the constraint of convergence (see Lemma \textcolor{red}{3}), thus, the Proposition \eqref{eq:appendix_lemma3} is always true. Therefore, $\|\vz\|^*$ is monotonically increasing as $p_d$ increases.

\vspace{1mm}

\noindent \textbf{Theorem \textcolor{red}{1}.} \textit{The regularization loss function $\mathcal{L}_{reg}$ certainly converges at the expected point in training when $k>0$, $l_a<u_a$ and $0 \leq p_d \leq 1$.}

\noindent \textbf{Proof.} Immediate from Lemmas \textcolor{red}{3}, \textcolor{red}{4} and \textcolor{red}{5}.

\subsection{Constraint for balancing encoding value}
\label{subsec:proof_balanc}

\begin{figure}[!t]
  \centering
   \includegraphics[width=0.9\linewidth]{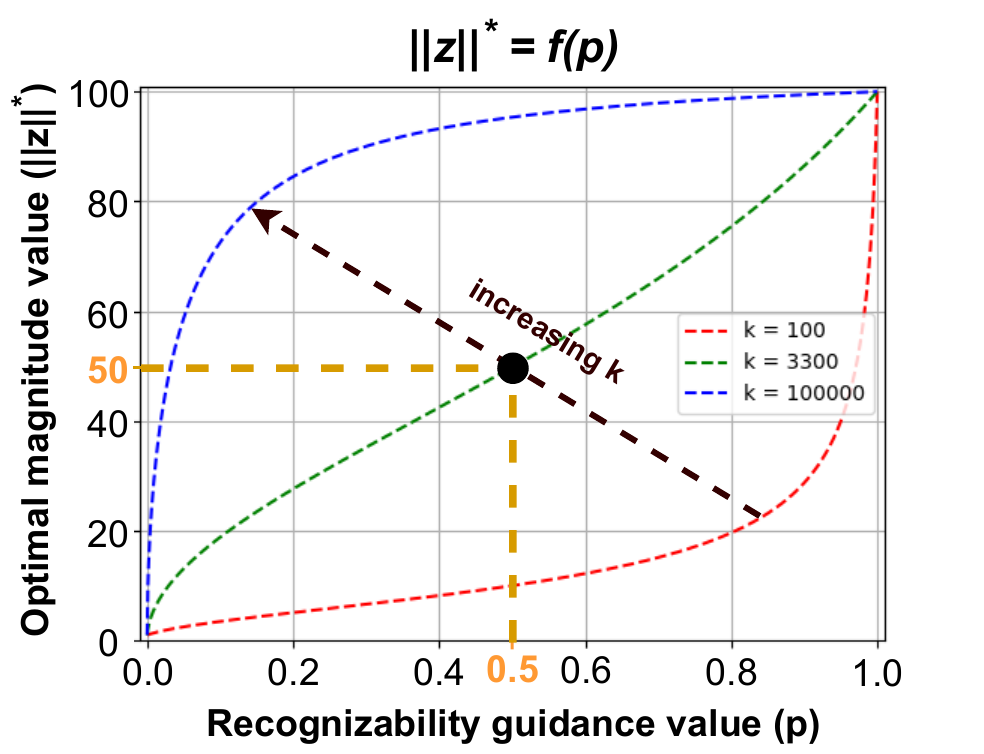}
   \caption{The visualization of function analysis of $f(p_d) \\ \forall p_d \in \left[0, 1\right]$.}
   \label{fig:appendix_functionanalysis}
\end{figure} 

To enhance the ability of recognizability representation, the $\|\vz\|^*$ value must be guaranteed to be linearly increasing as $p_d$ increases. This constraint makes the value of feature magnitude linearly distributed in the range value of $p_d$ ($p_d \in \left[0, 1\right]$). \cref{fig:appendix_functionanalysis} shows the effect of the value $k$ on the linearity of $f_{\vz}(p_d)$ (see \cref{eq:appendix_z}).

\noindent Therefore, to balance the range of encoded values, $k$ must be chosen to satisfy
\begin{equation*}
    \resizebox{0.25\textwidth}{!}{$
        \begin{cases}
            \begin{array}{lcc}
                f(p_d=0)   &= &l_a \\ 
                f(p_d=1)   &= &u_a \\
                f(p_d=0.5) &= &\frac{l_a+u_a}{2} 
            \end{array}
        \end{cases}
    $}
\end{equation*}
The first two constraints are always fulfilled when replacing $p_d=0$ and $p_d=1$ in $f(p_d)$. Thus, the third equation is a main condition for $k$ selection. After resolving it, we have
\begin{equation}
    \resizebox{0.25\textwidth}{!}{$
        \begin{aligned}
            k = \frac{u_a^2.\left[(u_a+l_a)^2-4.u_a^2.l_a^2\right]}{l_a^2.\left[4.u_a^2-(u_a+l_a)^2\right]}
        \end{aligned}
    $}
    \label{eq:appendix_k}
\end{equation}
However, the choice of $k$ must satisfy the constraint of convergence in training (see Lemma \textcolor{red}{3}), specifically, $k>0$. Let us consider the numerator and denominator of \cref{eq:appendix_k}
\begin{equation}
    \resizebox{0.45\textwidth}{!}{$
        \begin{cases}
            \begin{array}{lc}
                u_a^2.\left[(u_a+l_a)^2-4l_a^2\right] &>0 \vspace{1mm}\\
                l_a^2.\left[4u_a^2-(u_a+l_a)^2\right] &>0    
            \end{array}
        \end{cases}
        \iff 2l_a<ua+l_a<2u_a
    $}
    \label{eq:appendix_kcondition}
\end{equation}
Proposition \eqref{eq:appendix_kcondition} is always true since $u_a>l_a$. Therefore, the chosen $k$ following \cref{eq:appendix_k} still satisfies the convergence constraint.

\subsection{Constraint for tracking regularization loss}
\label{subsec:proof_loss_track}

\begin{figure}[!t]
  \centering
   \includegraphics[width=0.9\linewidth]{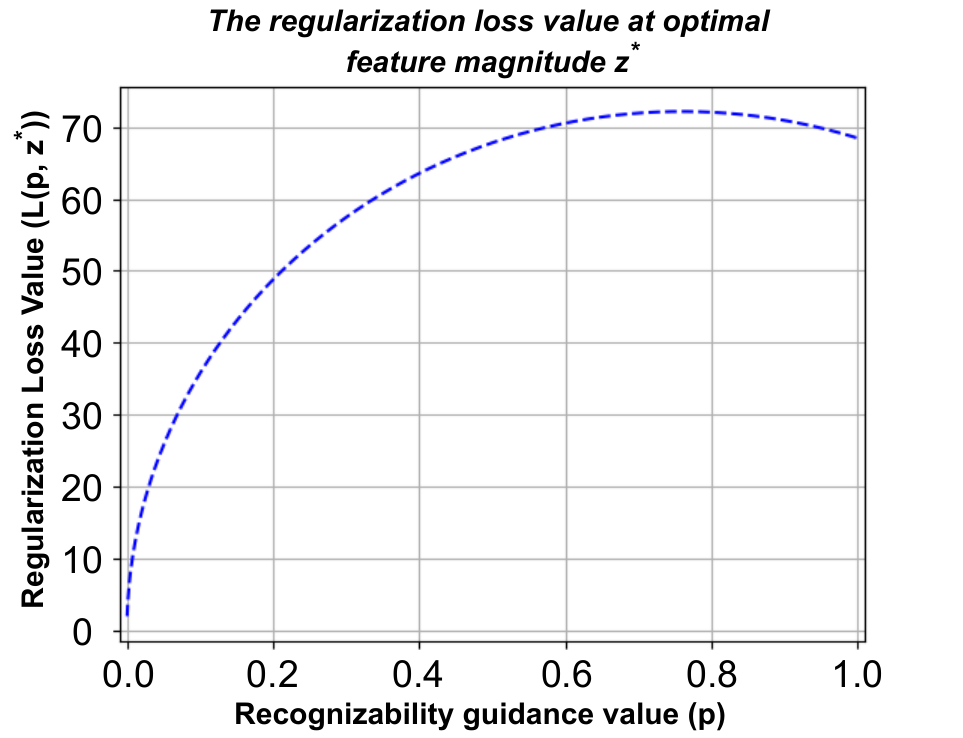}
   \caption{The visualization of loss value at optimal magnitude value $\|\vz\|^*$, $\mathcal{L}_{reg}(\|\vz\|^*)$.}
   \label{fig:appendix_lossval}
\end{figure}

By replacing $\|\vz\|$ in \cref{eq:supp_qcface_regular} by $\|\vz\|^*$ in \cref{eq:appendix_z}, setting value $b$ to $0$ and choosing $l_a$, $u_a$ and $k$ followed the two above constraints (\textit{i.e.,} \cref{subsec:proof_convergence,subsec:proof_balanc}), the optimal values of $\mathcal{L}_{reg}$ are plotted by \cref{fig:appendix_lossval}. However, this value makes them difficult to track the convergence point of $\mathcal{L}_{reg}$ since each value of $p_d$ yields a distinct value of optimal loss value $\mathcal{L}_{reg}(p_d, \|\vz\|^*)$, where $\|\vz\|^*$ is an optimal magnitude value for each $p_d$. To get over this problem, $b$ is used as a (reduction) offset value to make $\mathcal{L}_{reg}(p_d, \|\vz\|^*)=0$. 
\begin{equation*}
    \resizebox{0.45\textwidth}{!}{$
    \implies b=k.p_{d}. \left (\frac{1}{\|\vz\|^*}+\frac{\|\vz\|^*}{u_a^2} \right ) + (1-p_{d}).\left (\frac{1}{\|\vz\|^*}+\frac{\|\vz\|^*}{l_a^2} \right)
    $}
\end{equation*}

\begin{figure}[!t]
  \centering
   \includegraphics[width=\linewidth]{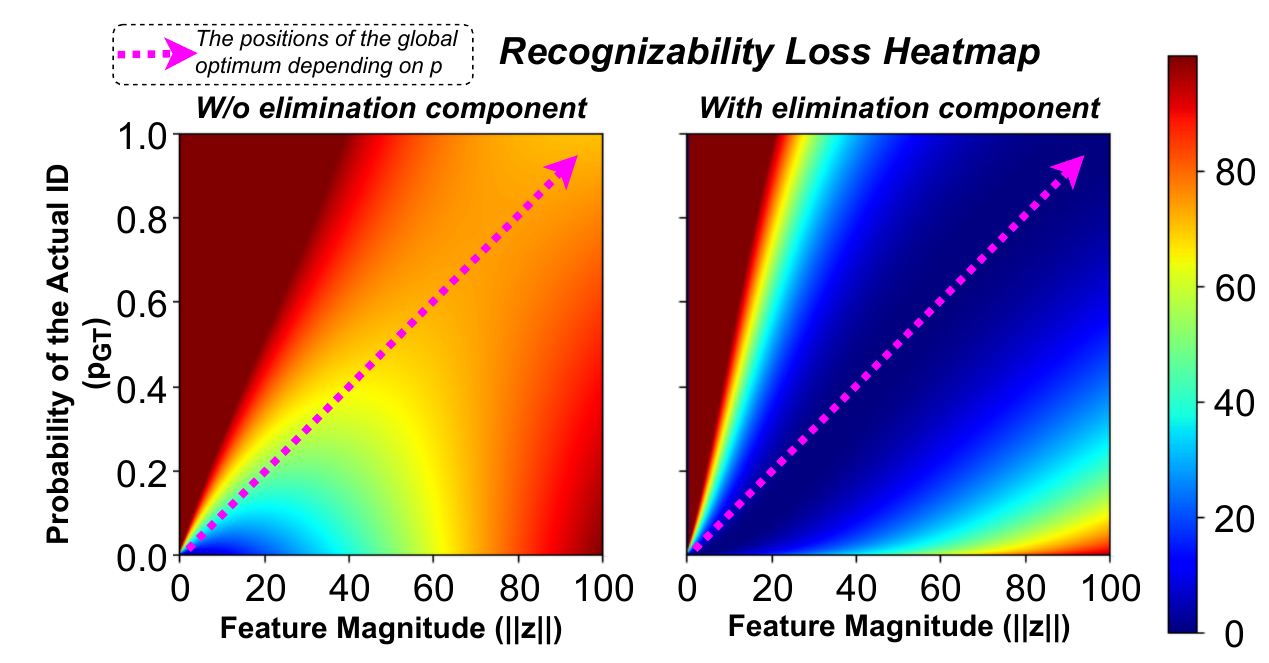}
   \caption{Two heatmaps of loss value without and with reduction offset. The \textcolor{VioletRed}{pink dashed arrow} expresses the optimum position (\textit{i.e.}, $\langle p^{*}, \|\vz\|^{*} \rangle$) where loss function has the minimum value. The color of heatmap expresses the value of regularization loss $\mathcal{L}_{reg}$.}
   \label{fig:appendix_loss_heatmap}
\end{figure}

\cref{fig:appendix_loss_heatmap} shows the heat maps of the loss value. With the reduction offset, the heat map turns out that $\mathcal{L}_{reg}$ achieves $0$ at all optimal points. Besides, the gradients from the loss function of both situations (\textit{i.e.,} with and without reduction offset) are similar to one another since $b$ is a constant value and does not contribute to the update gradient. Consequently, $b$ is only used to monitor whether $\mathcal{L}_{reg}$ reaches zero, without affecting the optimal result.


\section{Datasets and augmentation}
For evaluation datasets, AgeDB-30 \cite{moschoglou2017agedb}, CFP-FP \cite{sengupta2016frontal}, LFW \cite{huang2008labeled}, CALFW \cite{zheng2017cross}, CPLFW \cite{zheng2018cross}, XQLFW \cite{knoche2021cross}, IJB-B \cite{whitelam2017iarpa}, IJB-C \cite{maze2018iarpa} and TinyFace \cite{cheng2018low} are used as benchmark datasets, which are widely used in almost SoTA studies. They all play an important role in the comprehensive evaluation of model recognition ability and fair comparison with other methods. The details of each evaluation dataset are shown in \cref{tab:appendix_exp_evaluate_ds}. In addition, the details of augmentation, which also affect the learning improvement of a model, are shown in \cref{tab:appendix_augment}.

\begin{table}[!t]
\centering
\caption{The details of evaluation datasets.}
\label{tab:appendix_exp_evaluate_ds}
\begin{adjustbox}{width=0.9\columnwidth,center}
\begin{tabular}{l;{1pt/1pt}c;{1pt/1pt}c;{1pt/1pt}c} 
\hline
\rowcolor[rgb]{0.784,0.784,0.784} \multicolumn{1}{c|}{{\cellcolor[rgb]{0.784,0.784,0.784}}}                                    & \multicolumn{2}{c|}{\textbf{ No. Pairs}}                                        & {\cellcolor[rgb]{0.784,0.784,0.784}}                                        \\ 
\cline{2-3}
\rowcolor[rgb]{0.784,0.784,0.784} \multicolumn{1}{c|}{\multirow{-2}{*}{{\cellcolor[rgb]{0.784,0.784,0.784}}\textbf{Dataset }}} & \multicolumn{1}{c|}{\textbf{Positive}} & \multicolumn{1}{c|}{\textbf{Negative}} & \multirow{-2}{*}{{\cellcolor[rgb]{0.784,0.784,0.784}}\textbf{No. Images }}  \\ 
\hline\hline
AgeDB-30 \cite{moschoglou2017agedb}                                                                                                                      & 3k                                     & 3k                                     & 12k                                                                         \\ 
\hdashline[1pt/1pt]
CFP-FP \cite{sengupta2016frontal}                                                                                                                        & 3.5k                                   & 3.5k                                   & 12k                                                                         \\ 
\hdashline[1pt/1pt]
LFW \cite{huang2008labeled}                                                                                                                           & 3k                                     & 3k                                     & 12k                                                                         \\ 
\hdashline[1pt/1pt]
CALFW \cite{zheng2017cross}                                                                                                                         & 3k                                     & 3k                                     & 12k                                                                         \\ 
\hdashline[1pt/1pt]
CPLFW \cite{zheng2018cross}                                                                                                                         & 3k                                     & 3k                                     & 14k                                                                         \\ 
\hdashline[1pt/1pt]
XQLFW \cite{knoche2021cross}                                                                                                                         & 3k                                     & 3k                                     & 13.233k                                                                          \\ 
\hdashline[1pt/1pt]
IJB-B \cite{whitelam2017iarpa}                                                                                                                         & 10k                                    & 8M                                     & 228k                                                                        \\ 
\hdashline[1pt/1pt]
IJB-C \cite{maze2018iarpa}                                                                                                                         & 19.5k                                  & 15M                                    & 469k                                                                        \\
\hline
\end{tabular}
\end{adjustbox}
\end{table}

\begin{table}[!t]
\centering
\caption{The details of the experimental augmentations.}
\label{tab:appendix_augment}
\begin{adjustbox}{width=0.75\columnwidth,center}
    \begin{tabular}{l;{1pt/1pt}l} 
    \hline
    \rowcolor[rgb]{0.784,0.784,0.784} \multicolumn{1}{c|}{\begin{tabular}[c]{@{}>{\cellcolor[rgb]{0.784,0.784,0.784}}c@{}}\textbf{Augment}\\\textbf{Operator}\end{tabular}} & \multicolumn{1}{c}{\textbf{Details}}  \\ 
    \hline\hline
    \begin{tabular}[c]{@{}l@{}}Horizontal \&\\Vertical Flip\end{tabular} & p=0.3                                         \\ 
    \hdashline[1pt/1pt]
    Rotation                                                             & p=0.3; angle=$90^o$                           \\ 
    \hdashline[1pt/1pt]
    Blur                                                                 & p=0.3; type=\{Gaussian, Median\}              \\ 
    \hdashline[1pt/1pt]
    CLAHE                                                                & p=0.3                                         \\ 
    \hdashline[1pt/1pt]
    \begin{tabular}[c]{@{}l@{}}Brightness \&\\Contrast\end{tabular}      & p=0.3                                         \\ 
    \hdashline[1pt/1pt]
    \begin{tabular}[c]{@{}l@{}}Image \\Compression\end{tabular}          & p=1.0; quality (\%) = $\left[20, 100\right] $ \\
    \hline
    \end{tabular}
\end{adjustbox}
\end{table}

\section{Warming-up in QCFace training}
\begin{figure}[!t]
    \centering
    \includegraphics[width=\columnwidth]{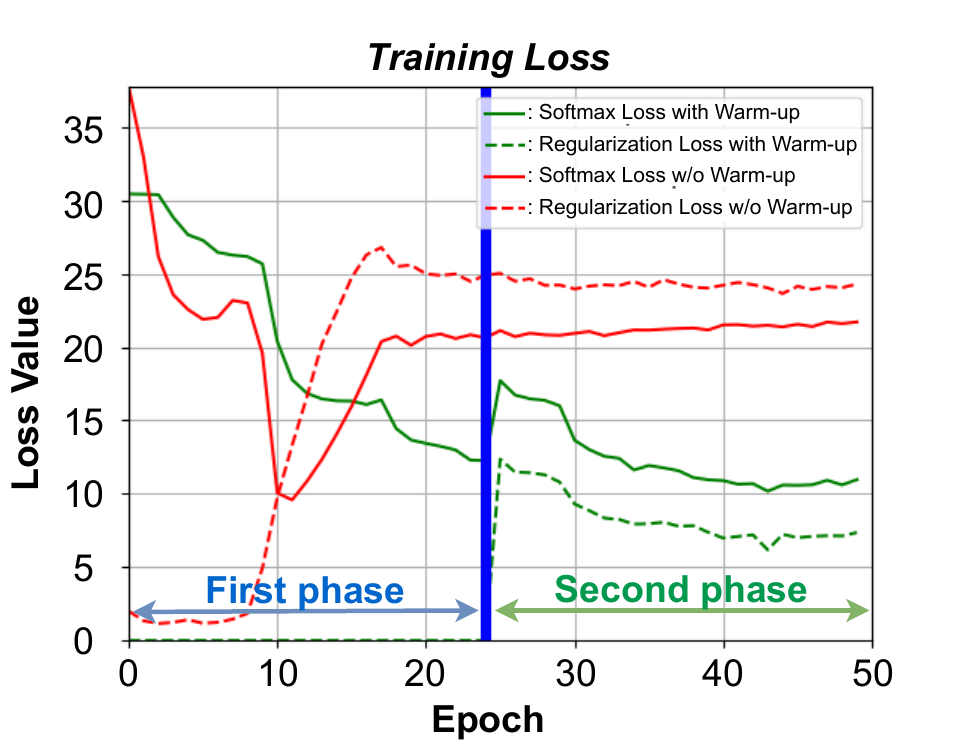}
    \caption{The training loss visualization of \textbf{\textit{QCFace}}.\protect\footnotemark[7]}
    \label{fig:supp_train_loss}
\end{figure}

Based on the evaluation with IResNet18, the negative effect of proxy oscillation readily emerges, as demonstrated by the convergence evaluation results. Specifically, the FR model must be pre-trained with ArcFace before applying \textbf{\textit{QCFace}} in order to obtain a meaningful \textbf{\textit{guidance value}}, called \textit{warm-up training phase}. \cref{fig:supp_train_loss} shows that without the \textit{warm-up phase}, $\mathcal{L}_{sm}$ and $\mathcal{L}_{rc}$ do not decrease simultaneously in the first phase; even so, both loss values increase at epoch 11, and cannot be optimized later. In contrast, with the warm-up, the convergence process is smooth in both  training phases. \cref{tab:supp_ablation_condition} further demonstrates in verification accuracy of the necessity of the \textit{warm-up phase} for the \textbf{\textit{frozen proxy}} guarantee, and provides a comparison between Cfg-1 and Cfg-2. However, as we mentioned in the main paper, this problem can be overcome by a large-size backbone and a large-scale training dataset.

\footnotetext[7]{These experimental results are done with the backbone IResNet18.}

\begin{table}[!t]
\centering
\caption{The verification performance of \textbf{\textit{QCFace-Arc}} in different training configurations. Cfg-1, Cfg-2 and Cfg-3 express non-warm-up, non-augmentation and full setting, respectively.\protect\footnotemark[7]}
\label{tab:supp_ablation_condition}
\begin{adjustbox}{width=0.9\columnwidth,center}
    \begin{tabular}{l;{1pt/1pt}c;{1pt/1pt}c;{1pt/1pt}c;{1pt/1pt}c;{1pt/1pt}c} 
    \hline
    \rowcolor[rgb]{0.784,0.784,0.784} \multicolumn{1}{c|}{\textbf{\textbf{Config}}} & \multicolumn{1}{c|}{\textbf{AgeDB-30}} & \multicolumn{1}{c|}{\textbf{CFP-FP}} & \multicolumn{1}{c|}{\textbf{LFW}} & \multicolumn{1}{c|}{\textbf{CALFW}} & \textbf{CPLFW}  \\ 
    \hline\hline
    Cfg-1                                                                           & 49.900                                 & 53.700                               & 81.600                            & 70.650                              & 53.900          \\ 
    \hdashline[1pt/1pt]
    Cfg-2                                                                           & 91.217                                 & 88.457                               & 98.783                            & 91.683                              & 82.200          \\ 
    \hdashline[1pt/1pt]
    Cfg-3                                                                           & 92.050                                 & 91.429                               & 99.117                            & 92.150                              & 84.700          \\
    \hline
    \end{tabular}
\end{adjustbox}
\end{table}

\vspace{1cm}

\section{Augmentation Effect} 
In \cref{tab:supp_ablation_condition}, the use of augmentation in \textbf{\textit{QCFace}} yields performance gains of $0.740\%$, $4.003\%$, $0.304\%$, $0.709\%$, and $3.211\%$ on AgeDB-30, CFP-FP, LFW, CALFW, and CPLFW, respectively. This substantial effect appears when utilizing small-scale training datasets (\textit{e.g.}, Casia-WebFace \cite{deng2019lightweight}) and small-size backbones (\textit{i.e.}, IResNet18).


\section{Extended visualization \& Experiments}
In this section, we extend the visualization of the geometrical representation of the embedded feature (see \cref{subsec:appendix_emb}). Moreover, we show the comprehensive evaluation of verification and identification with two backbone scales of IResNet \cite{duta2021improved}, \textit{i.e.} IResNet18 and IResNet100 (see \cref{subsec:appendix_veri_iden}).

\begin{figure*}[!t]
  \centering
  \begin{subfigure}{0.25\linewidth}
    \includegraphics[width=\textwidth]{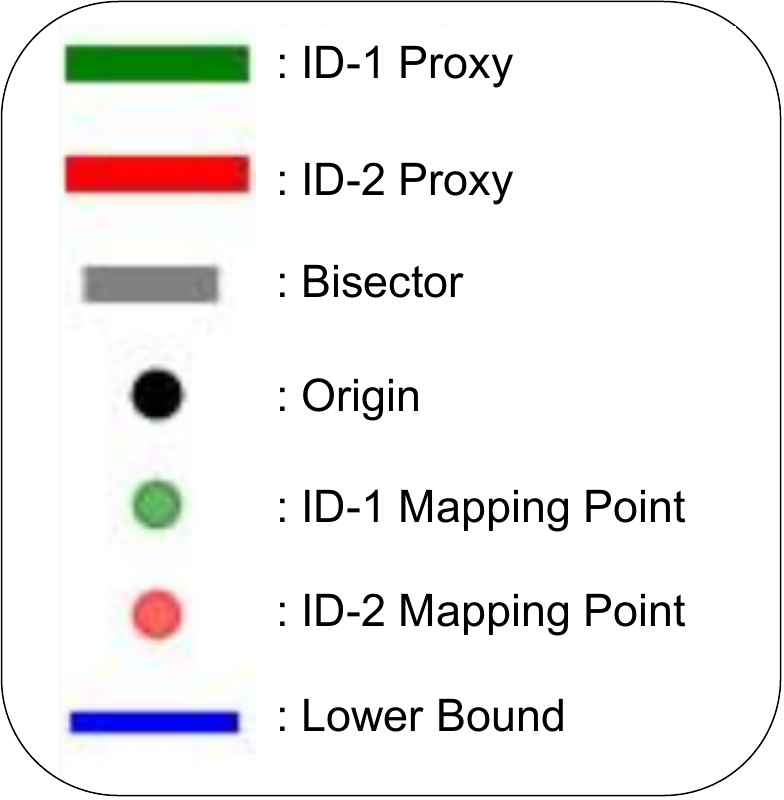}
    \vspace{20pt}
  \end{subfigure}
  \hfill
  \begin{subfigure}{0.32\linewidth}
    \includegraphics[width=\textwidth]{images/Mapping-Point/MP-Exp-ArcFace.pdf}
    \caption{ArcFace \cite{deng2019arcface}}
    \label{fig:appendix_mp_arc}
  \end{subfigure}
  \hfill
  \begin{subfigure}{0.32\linewidth}
    \includegraphics[width=\textwidth]{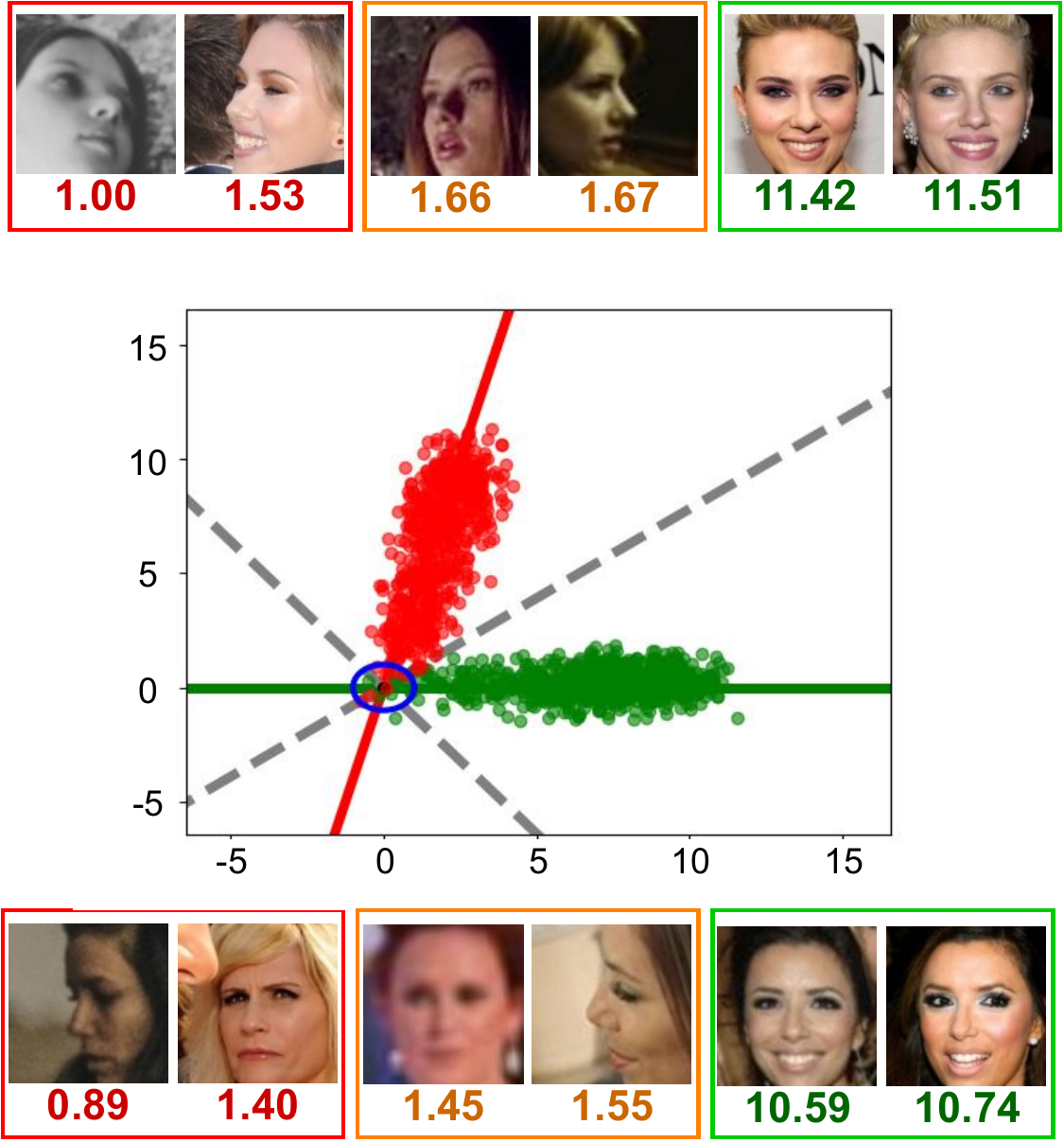}
    \caption{Curricular Face \cite{huang2020curricularface}}
    \label{fig:appendix_mp_curricular}
  \end{subfigure}
  \hfill
  \begin{subfigure}{0.32\linewidth}
    \includegraphics[width=\textwidth]{images/Mapping-Point/MP-Exp-MagFace.pdf}
    \caption{MagFace \cite{meng2021magface}}
    \label{fig:appendix_mp_mag}
  \end{subfigure}
  \hfill
  \begin{subfigure}{0.32\linewidth}
    \includegraphics[width=\textwidth]{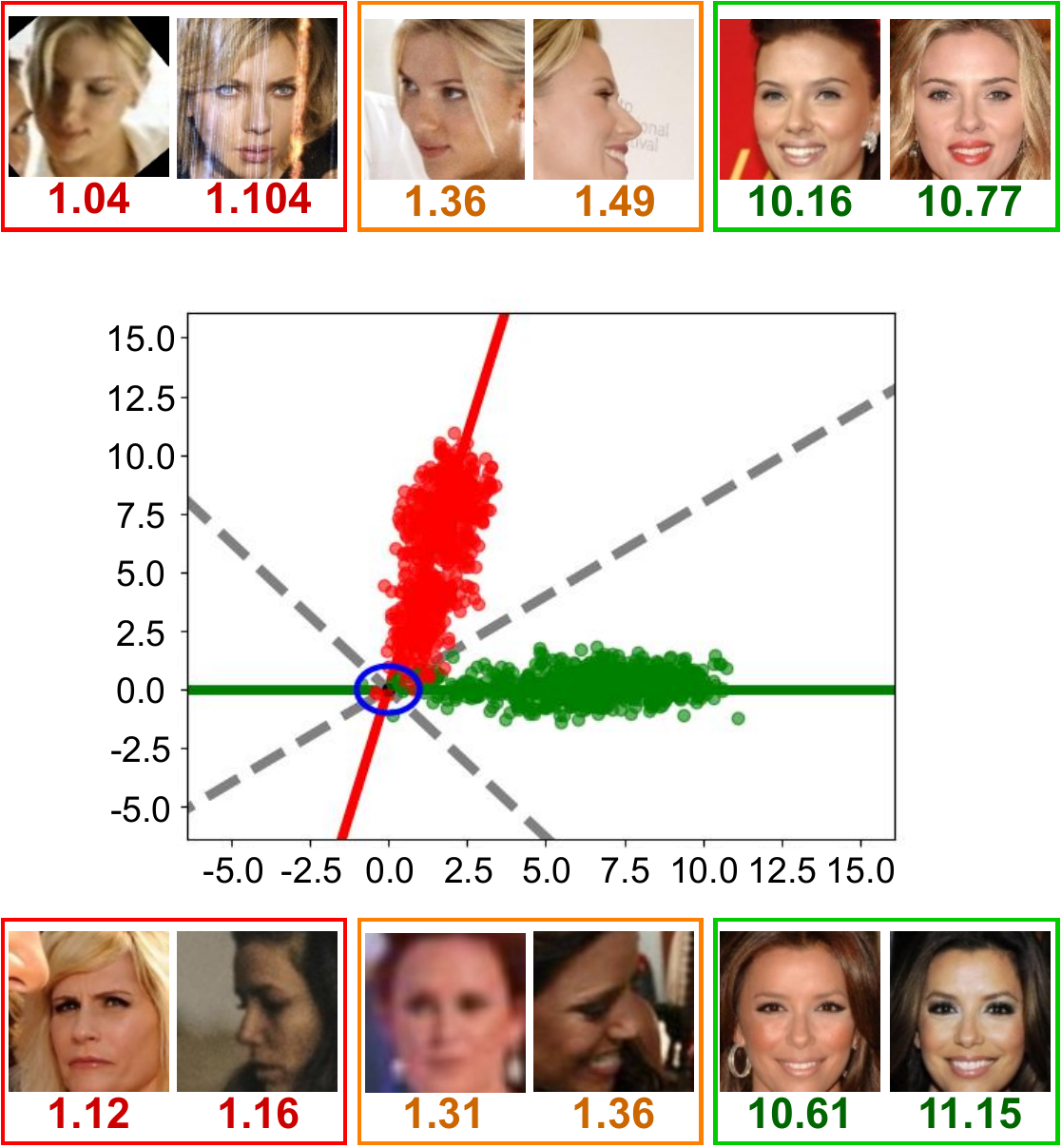}
    \caption{AdaFace \cite{kim2022adaface}}
    \label{fig:appendix_mp_ada}
  \end{subfigure}
  \hfill
  \begin{subfigure}{0.32\linewidth}
    \includegraphics[width=\textwidth]{images/Mapping-Point/MP-Exp-QCFace.pdf}
    \caption{\textbf{\textit{QCFace (Ours)}}}
    \label{fig:appendix_mp_qc}
  \end{subfigure}
  \caption{The non-normalized geometrical representation of the feature space optimized by \textbf{\textit{QCFace-Arc}} and the cutting-edge methods. The feature vectors of each image extracted by face backbones are projected to the hyperplane formed by two proxies of two identities in the CASIA-WebFace dataset by the Gram-Schmidt algorithm. The \textbf{border} \textcolor{red}{red} and \textcolor{OliveGreen}{green} lines express the proxies of two identities. The actual proxy of each mapping feature point has the same color as that of feature. The \textcolor{gray}{gray} dashed lines are the internal and external bisectors. The number below each image sample is the magnitude of the mapping feature vector.}
  \label{fig:appendix_mapping_point}
\end{figure*}

\subsection{Embedded feature distribution comparison}
\label{subsec:appendix_emb}
We add the embedded feature representations of ArcFace \cite{deng2019arcface} and CurricularFace \cite{huang2020curricularface}, whose embedded features are also uniformly distributed in the low range of magnitude value ($<15$), similar to AdaFace (\cref{fig:appendix_mp_arc,fig:appendix_mp_curricular,fig:appendix_mp_ada}). Meanwhile, \textbf{\textit{QCFace}} shows its superiority with its widest range of feature magnitude for representing the diversity of the recognizability level of face images.

\subsection{Verification and Identification evaluation}
\label{subsec:appendix_veri_iden} 
For the comprehensive evaluation, we also show the verification and identification benchmarks of both IResNet18 and IResNet100 based on high-quality (see \cref{tab:supp_comparison_hq_IR100,tab:supp_comparison_hq_IR18}), mixed-quality and low-quality benchmark datasets (see \cref{tab:supp_comparison_veri_IJB_IR100,tab:supp_comparison_veri_IJB_IR18} for verification benchmarks and \cref{tab:supp_comparison_iden_IJB_IR100,tab:supp_comparison_iden_IJB_IR18} for identification benchmarks and MegaFace Challenge \cite{kemelmacher2016megaface}). 

For high-quality benchmarks, the negative effect of proxy oscillation readily emerges, clearly shown by the evaluation results of \textbf{\textit{(Q)QCFace-Cur}} and \textbf{\textit{(Q)QCFace-MVS}} trained with IResNet18, while \textbf{\textit{(Q)QCFace-Arc}} also achieves the top-1 and top-2 verification accuracy. 

The limitation in recognizing very low-quality images is clearly demonstrated by the evaluation results of \textbf{\textit{QCFace}} using backbone IResNet18 as its feature extractor (see \textbf{Limitations} in main paper). Specifically, the recognition performance dramatically drops when benchmarking on the mixed-quality (\textit{i.e.}, IJB-B and IJB-C) and low-quality (\textit{i.e.}, TinyFace) datasets (see \cref{tab:supp_comparison_veri_IJB_IR18,tab:supp_comparison_iden_IJB_IR18}). Consequently, the large-scale backbone can enhance generalization ability, mitigating the limited impact of low-recognizability images in the presence of mislabeled data in the low-magnitude representative region.

\begin{table*}[!t]
\caption{Comparison in verification accuracy and AUC-ROC on high-quality datasets with IResNet100. The \textbf{border}, \uline{underlined} and \textit{italic} numbers correspondingly express \textbf{top-1}, \uline{top-2} and \textit{top-3} accuracy. In notation \textbf{\textit{(Q)QCFace}-X}, \textbf{\textit{(Q)}} expresses the use of \cite{terhorst2023qmagface} as a similarity score calculation instead of cosine similarity in decision making, \textbf{\textit{X}} expresses the name of softmax loss (\textit{i.e.}, Arc - ArcFace, Cur - CurricularFace, MVS - MVSoftmax). \textcolor{YellowGreen}{Geen} and \textcolor{Salmon}{red} backgrounds express the use of margin-based and misclassified softmax losses.}
\label{tab:supp_comparison_hq_IR100}
\begin{adjustbox}{width=1.85\columnwidth,center}
\begin{tabular}{c|l|c;{1pt/1pt}c|c;{1pt/1pt}c|c;{1pt/1pt}c|c;{1pt/1pt}c|c;{1pt/1pt}c|c;{1pt/1pt}c} 
\hline
\rowcolor[rgb]{0.784,0.784,0.784} {\cellcolor[rgb]{0.784,0.784,0.784}}                                & \multicolumn{1}{c|}{{\cellcolor[rgb]{0.784,0.784,0.784}}}                                  & \multicolumn{2}{c|}{\textbf{AgeDB-30}}                             & \multicolumn{2}{c|}{\textbf{CFP-FP}}                               & \multicolumn{2}{c|}{\textbf{LFW}}                                  & \multicolumn{2}{c|}{\textbf{CALFW}}                                & \multicolumn{2}{c|}{\textbf{CPLFW}}                                & \multicolumn{2}{c}{\textbf{XQLFW}}                   \\ 
\cline{3-14}
\rowcolor[rgb]{0.784,0.784,0.784} \multirow{-2}{*}{{\cellcolor[rgb]{0.784,0.784,0.784}}\textbf{Year}} & \multicolumn{1}{c|}{\multirow{-2}{*}{{\cellcolor[rgb]{0.784,0.784,0.784}}\textbf{Method}}} & \multicolumn{1}{c|}{\textbf{\textit{Ver}}} & \textbf{\textit{AUC}} & \multicolumn{1}{c|}{\textbf{\textit{Ver}}} & \textbf{\textit{AUC}} & \multicolumn{1}{c|}{\textbf{\textit{Ver}}} & \textbf{\textit{AUC}} & \multicolumn{1}{c|}{\textbf{\textit{Ver}}} & \textbf{\textit{AUC}} & \multicolumn{1}{c|}{\textbf{\textit{Ver}}} & \textbf{\textit{AUC}} & \multicolumn{1}{c|}{\textbf{Ver}} & \textbf{AUC}     \\ 
\hline\hline
\textit{2015}                                                                                         & FaceNet \cite{schroff2015facenet}                                                                                   & 58.150                                     & 61.360                & 70.314                                     & 77.734                & 79.417                                     & 87.862                & 59.567                                     & 64.095                & 62.633                                     & 64.760                & 66.650                            & 72.543           \\ 
\hdashline[1pt/1pt]
\textit{2017}                                                                                         & SphereFace \cite{liu2017sphereface}                                                                                & 96.100                                     & 98.843                & 93.729                                     & 97.783                & 99.600                                     & 99.852                & 95.400                                     & 97.983                & 88.917                                     & 93.759                & 79.883                            & 87.493           \\ 
\hdashline[1pt/1pt]
\textit{2018}                                                                                         & CosFace \cite{wang2018cosface}                                                                                   & 97.033                                     & 98.866                & 95.743                                     & 98.497                & 99.750                                     & 99.862                & 95.467                                     & 97.866                & 90.317                                     & 94.635                & 81.783                            & 90.128           \\ 
\hdashline[1pt/1pt]
\textit{2019}                                                                                         & ArcFace \cite{deng2019arcface}                                                                                   & 97.133                                     & 98.917                & 95.100                                     & 98.521                & 99.633                                     & 99.878                & 95.517                                     & 97.921                & 90.217                                     & 94.707                & 81.833                            & 90.087           \\ 
\hdashline[1pt/1pt]
\textit{2020}                                                                                         & MV-Arc-Softmax \cite{wang2020mis}                                                                            & 97.100                                     & 98.972                & 95.757                                     & 98.541                & 99.700                                     & 99.826                & 95.617                                     & 97.918                & 90.233                                     & 94.414                & 83.150                            & 90.814           \\ 
\hdashline[1pt/1pt]
\textit{2020}                                                                                         & CurricularFace \cite{huang2020curricularface}                                                                            & 97.100                                     & 98.962                & 94.686                                     & 98.418                & 99.550                                     & 99.846                & 95.700                                     & 97.929                & 90.083                                     & 94.607                & 82.650                            & 90.265           \\ 
\hdashline[1pt/1pt]
\textit{2021}                                                                                         & MagFace \cite{meng2021magface}                                                                                   & 96.867                                     & 98.797                & 95.357                                     & 98.593                & 99.600                                     & \textit{99.889}       & 95.517                                     & 97.797                & 89.583                                     & 94.105                & 81.233                            & 89.090           \\ 
\hdashline[1pt/1pt]
\textit{2021}                                                                                         & VPLFace \cite{deng2021variational}                                                                                   & 97.183                                     & 98.984                & 94.257                                     & 98.163                & 99.700                                     & 99.865                & 95.683                                     & 97.858                & 89.617                                     & 94.483                & 83.117                            & 90.677           \\ 
\hdashline[1pt/1pt]
\textit{2022}                                                                                         & SphereFace2 \cite{wen2021sphereface2}                                                                               & 96.000                                     & 98.777                & 94.029                                     & 98.210                & 99.583                                     & 99.878                & 95.283                                     & 97.949                & 90.100                                     & 94.677                & 83.933                            & 91.551           \\ 
\hdashline[1pt/1pt]
2022                                                                                                  & Elastic-Arc \cite{boutros2022elasticface}                                                                               & 97.283                                     & 98.930                & 94.329                                     & 98.004                & 99.467                                     & 99.851                & 95.500                                     & 97.799                & 89.800                                     & 94.156                & 82.467                            & 89.896           \\ 
\hdashline[1pt/1pt]
2022                                                                                                  & Elastic-Cos \cite{boutros2022elasticface}                                                                                & 97.417                                     & 98.979                & 95.357                                     & 98.432                & 99.700                                     & \textit{99.890}       & 95.733                                     & 97.837                & 89.400                                     & 94.395                & 82.350                            & 89.729           \\ 
\hdashline[1pt/1pt]
\textit{2022}                                                                                         & AdaFace \cite{kim2022adaface}                                                                                   & 96.817                                     & 98.924                & 95.614                                     & 98.723                & 99.717                                     & 99.866                & 95.733                                     & 97.790                & 90.917                                     & 94.957                & 82.017                            & 89.839           \\ 
\hdashline[1pt/1pt]
\textit{2023}                                                                                         & QAFace \cite{saadabadi2023quality}                                                                                    & 96.783                                     & 98.942                & 95.243                                     & 98.469                & 99.733                                     & 99.887                & 95.400                                     & 97.877                & 89.817                                     & 94.430                & 83.550                            & 90.946           \\ 
\hdashline[1pt/1pt]
\textit{2023}                                                                                         & QMagFace \cite{terhorst2023qmagface}                                                                                  & 97.250                                     & 98.895                & 96.014                                     & 99.119                & 99.700                                     & 99.916                & 95.617                                     & 97.801                & 90.683                                     & 96.042                & 81.483                            & 88.922           \\ 
\hdashline[1pt/1pt]
\textit{2023}                                                                                         & UniFace \cite{zhou2023uniface}                                                                                   & 97.083                                     & 98.903                & 95.686                                     & 98.749                & 99.617                                     & 99.829                & 95.667                                     & 97.856                & 90.683                                     & 95.029                & 82.700                            & 90.755           \\ 
\hdashline[1pt/1pt]
\textit{2023}                                                                                         & UniTSFace \cite{jia2023unitsface}                                                                                 & 97.217                                     & 99.008                & 95.371                                     & 98.550                & 99.667                                     & 99.889                & 95.717                                     & 97.936                & 90.450                                     & 94.835                & 83.517                            & 91.171           \\ 
\hdashline[1pt/1pt]
\textit{2024}                                                                                         & TopoFR \cite{dan2024topofr}                                                                                     & 97.500                                     & \textit{99.021}       & 95.871                                     & 99.015                & 99.750                                     & 99.876                & 95.650                                     & 97.913                & 90.383                                     & 95.490                & 83.583                            & 91.070           \\ 
\hline\hline
\rowcolor[rgb]{0.765,0.918,0.773} \textit{Now}                                                        & \textbf{\textit{QCFace-Arc}}                                                               & \uline{98.100}                             & \uline{99.129}        & \textit{98.200}                            & 99.372                & \textbf{99.800}                            & 99.872                & \textit{95.950}                            & \textit{98.031}       & \uline{92.483}                             & 95.624                & \uline{86.067}                    & \uline{93.327}   \\ 
\hdashline[1pt/1pt]
\rowcolor[rgb]{0.765,0.918,0.773} \textit{Now}                                                        & \textit{\textbf{QQCFace-Arc}}                                                              & \textbf{98.183}                            & \textbf{99.175}       & \textbf{98.457}                            & \uline{99.524}        & \textbf{99.800}                            & 99.879                & \textit{95.950}                            & \textbf{98.087}       & \textbf{92.833}                            & \textit{96.403}       & \textbf{86.267}                   & \textbf{93.634}  \\ 
\hline\hline
\rowcolor[rgb]{0.957,0.741,0.741} \textit{Now}                                                        & \textit{\textbf{QCFace-Cur}}                                                               & 97.700                                     & 98.979                & \textit{98.200}                            & \textit{99.515}       & \uline{99.783}                             & \uline{99.895}        & \textbf{96.083}                            & \textit{98.031}       & 92.383                                     & \uline{96.404}        & 83.283                            & 91.060           \\ 
\hdashline[1pt/1pt]
\rowcolor[rgb]{0.957,0.741,0.741} \textit{Now}                                                        & \textit{\textbf{QQCFace-Cur}}                                                              & \textit{97.917}                            & 99.013                & \uline{98.271}                             & \textbf{99.528}       & \textit{99.767}                            & \textbf{99.900}       & \uline{96.067}                             & \uline{98.032}        & \textit{92.467}                            & \textbf{96.718}       & 83.283                            & 91.466           \\ 
\hline\hline
\rowcolor[rgb]{0.957,0.741,0.741} \textit{Now}                                                        & \textit{\textbf{QCFace-MVS}}                                                               & 93.200                                     & 98.101                & 96.229                                     & 98.971                & 99.483                                     & 99.897                & 91.683                                     & 96.853                & 89.133                                     & 93.874                & 82.833                            & 91.058           \\ 
\hdashline[1pt/1pt]
\rowcolor[rgb]{0.957,0.741,0.741} \textit{Now}                                                        & \textit{\textbf{QQCFace-MVS}}                                                              & 93.517                                     & 98.066                & 96.186                                     & 99.001                & 99.467                                     & 99.892                & 91.833                                     & 96.757                & 88.817                                     & 93.692                & \textit{84.300}                   & \textit{92.166}  \\
\hline
\end{tabular}
\end{adjustbox}
\end{table*}

\begin{table*}[!t]
\caption{Comparison in verification accuracy and AUC-ROC on high-quality datasets with IResNet18. The \textbf{border}, \uline{underlined} and \textit{italic} numbers correspondingly express \textbf{top-1}, \uline{top-2} and \textit{top-3} accuracy. In notation \textbf{\textit{(Q)QCFace}-X}, \textbf{\textit{(Q)}} expresses the use of \cite{terhorst2023qmagface} as a similarity score calculation instead of cosine similarity in decision making, \textbf{\textit{X}} expresses the name of softmax loss (\textit{i.e.}, Arc - ArcFace, Cur - CurricularFace, MVS - MVSoftmax). \textcolor{YellowGreen}{Geen} and \textcolor{Salmon}{red} backgrounds express the use of margin-based and misclassified softmax losses.}
\label{tab:supp_comparison_hq_IR18}
\begin{adjustbox}{width=1.85\columnwidth,center}
\begin{tabular}{c|l|c;{1pt/1pt}c|c;{1pt/1pt}c|c;{1pt/1pt}c|c;{1pt/1pt}c|c;{1pt/1pt}c|c;{1pt/1pt}c} 
\hline
\rowcolor[rgb]{0.784,0.784,0.784} {\cellcolor[rgb]{0.784,0.784,0.784}}                                & \multicolumn{1}{c|}{{\cellcolor[rgb]{0.784,0.784,0.784}}}                                  & \multicolumn{2}{c|}{\textbf{AgeDB-30}}                             & \multicolumn{2}{c|}{\textbf{CFP-FP}}                               & \multicolumn{2}{c|}{\textbf{LFW}}                                  & \multicolumn{2}{c|}{\textbf{CALFW}}                                & \multicolumn{2}{c|}{\textbf{CPLFW}}                                & \multicolumn{2}{c}{\textbf{XQLFW}}                   \\ 
\cline{3-14}
\rowcolor[rgb]{0.784,0.784,0.784} \multirow{-2}{*}{{\cellcolor[rgb]{0.784,0.784,0.784}}\textbf{Year}} & \multicolumn{1}{c|}{\multirow{-2}{*}{{\cellcolor[rgb]{0.784,0.784,0.784}}\textbf{Method}}} & \multicolumn{1}{c|}{\textbf{\textit{Ver}}} & \textbf{\textit{AUC}} & \multicolumn{1}{c|}{\textbf{\textit{Ver}}} & \textbf{\textit{AUC}} & \multicolumn{1}{c|}{\textbf{\textit{Ver}}} & \textbf{\textit{AUC}} & \multicolumn{1}{c|}{\textbf{\textit{Ver}}} & \textbf{\textit{AUC}} & \multicolumn{1}{c|}{\textbf{\textit{Ver}}} & \textbf{\textit{AUC}} & \multicolumn{1}{c|}{\textbf{Ver}} & \textbf{AUC}     \\ 
\hline\hline
\textit{2015}                                                                                         & FaceNet \cite{schroff2015facenet}                                                                                   & 66.300                                     & 71.282                & 71.429                                     & 79.354                & 84.250                                     & 92.479                & 66.183                                     & 71.948                & 63.283                                     & 67.745                & 62.483                            & 67.077           \\
\hline
\end{tabular}
\end{adjustbox}
\end{table*}

\begin{table*}[!t]
\caption{Comparison in verification performance on IJB dataset with IResNet100.}
\label{tab:supp_comparison_veri_IJB_IR100}
\begin{adjustbox}{width=1.85\columnwidth,center}
\begin{tabular}{c|l|c;{1pt/1pt}c;{1pt/1pt}c;{1pt/1pt}l;{1pt/1pt}l;{1pt/1pt}l;{1pt/1pt}c|c;{1pt/1pt}c;{1pt/1pt}c;{1pt/1pt}l;{1pt/1pt}l;{1pt/1pt}l;{1pt/1pt}c} 
\hline
\rowcolor[rgb]{0.784,0.784,0.784} {\cellcolor[rgb]{0.784,0.784,0.784}}                                & \multicolumn{1}{c|}{{\cellcolor[rgb]{0.784,0.784,0.784}}}                                  & \multicolumn{7}{c|}{\textbf{IJB-B (TAR@FAR)}}                                                                                                                                                                                                                                                                                                                                       & \multicolumn{7}{c}{\textbf{IJB-C (TAR@FAR)}}                                                                                                                                                                                                                                                                                                                                         \\
\hline
\end{tabular}
\end{adjustbox}
\end{table*}

\begin{table*}[!t]
\caption{Comparison in verification performance on IJB dataset with IResNet18.}
\label{tab:supp_comparison_veri_IJB_IR18}
\begin{adjustbox}{width=1.85\columnwidth,center}
\begin{tabular}{c|l|c;{1pt/1pt}c;{1pt/1pt}c;{1pt/1pt}l;{1pt/1pt}l;{1pt/1pt}l;{1pt/1pt}c|c;{1pt/1pt}c;{1pt/1pt}c;{1pt/1pt}l;{1pt/1pt}l;{1pt/1pt}l;{1pt/1pt}c} 
\hline
\rowcolor[rgb]{0.784,0.784,0.784} {\cellcolor[rgb]{0.784,0.784,0.784}}                                & \multicolumn{1}{c|}{{\cellcolor[rgb]{0.784,0.784,0.784}}}                                  & \multicolumn{7}{c|}{\textbf{IJB-B (TAR@FAR)}}                                                                                                                                                                                                                                                                                                                                       & \multicolumn{7}{c}{\textbf{IJB-C (TAR@FAR)}}                                                                                                                                                                                                                                                                                                                                                                                                       \\
\hline
\end{tabular}
\end{adjustbox}
\end{table*}

\begin{table*}[!t]
\caption{Comparison in identification performance on IJB, TinyFace and MegaFace datasets with IResNet100. In the MegaFace benchmark, \textbf{\textit{Iden}} expresses \textbf{\textit{Rank-1}} in identification and \textbf{\textit{Veri}} is TAR@FAR=$10^{-6}$. The similarity score calculation of \cite{terhorst2023qmagface} is not supported by MegaFace devkit, expressed by the notation ``-".}
\label{tab:supp_comparison_iden_IJB_IR100}
\begin{adjustbox}{width=1.5\columnwidth,center}
\begin{tabular}{c|l|c;{1pt/1pt}c|c;{1pt/1pt}c|c;{1pt/1pt}c|c;{1pt/1pt}c} 
\hline
\rowcolor[rgb]{0.784,0.784,0.784} {\cellcolor[rgb]{0.784,0.784,0.784}}                                & \multicolumn{1}{c|}{{\cellcolor[rgb]{0.784,0.784,0.784}}}                                  & \multicolumn{2}{c|}{\textbf{IJB-B~}}                                                       & \multicolumn{2}{c|}{\textbf{IJB-C~}}                                                                         & \multicolumn{2}{c|}{\textbf{TinyFace}}                                                    & \multicolumn{2}{c}{\textbf{MegaFace}}                          \\ 
\cline{3-10}
\rowcolor[rgb]{0.784,0.784,0.784} \multirow{-2}{*}{{\cellcolor[rgb]{0.784,0.784,0.784}}\textbf{Year}} & \multicolumn{1}{c|}{\multirow{-2}{*}{{\cellcolor[rgb]{0.784,0.784,0.784}}\textbf{Method}}} & \multicolumn{1}{c|}{\textbf{\textit{Rank-1}}} & \textbf{\textbf{\textit{Rank-}}\textit{5}} & \multicolumn{1}{c|}{\textbf{\textbf{\textit{Rank-}}\textit{1}}} & \textbf{\textbf{\textit{Rank-}}\textit{5}} & \multicolumn{1}{c|}{\textbf{\textbf{\textit{Rank-1}}}} & \textbf{\textbf{\textit{Rank-5}}} & \multicolumn{1}{c|}{\textbf{\textit{Iden}}} & \textbf{Veri}    \\ 
\hline\hline
\textit{2015}                                                                                         & FaceNet \cite{schroff2015facenet}                                                                                   & 8.997                                         & 19.628                                     & 7.788                                                           & 16.240                                     & 11.534                                                 & 17.167                            & 0.724                                       & 2.209            \\ 
\hdashline[1pt/1pt]
\textit{2017}                                                                                         & SphereFace \cite{liu2017sphereface}                                                                                & 89.075                                        & 93.204                                     & 90.088                                                          & 93.411                                     & 51.261                                                 & 57.350                            & 89.597                                      & 91.387           \\ 
\hdashline[1pt/1pt]
\textit{2018}                                                                                         & CosFace \cite{wang2018cosface}                                                                                   & \textit{93.554}                               & 95.891                                     & 94.789                                                          & 96.392                                     & 60.381                                                 & 64.941                            & 95.617                                      & 96.260           \\ 
\hdashline[1pt/1pt]
\textit{2019}                                                                                         & ArcFace \cite{deng2019arcface}                                                                                   & 93.204                                        & 95.852                                     & 94.314                                                          & 96.193                                     & 57.672                                                 & 62.607                            & 95.242                                      & 95.895           \\ 
\hdashline[1pt/1pt]
\textit{2020}                                                                                         & MV-Arc-Softmax \cite{wang2020mis}                                                                            & 92.882                                        & 95.813                                     & 94.233                                                          & 96.172                                     & 59.120                                                 & 64.297                            & 94.188                                      & 94.944           \\ 
\hdashline[1pt/1pt]
\textit{2020}                                                                                         & CurricularFace \cite{huang2020curricularface}                                                                            & 93.476                                        & 95.794                                     & 94.794                                                          & 96.366                                     & 60.649                                                 & 64.941                            & 95.637                                      & 96.316           \\ 
\hdashline[1pt/1pt]
\textit{2021}                                                                                         & MagFace \cite{meng2021magface}                                                                                   & 92.911                                        & 95.803                                     & 94.187                                                          & 96.045                                     & 57.833                                                 & 62.527                            & 94.965                                      & 96.052           \\ 
\hdashline[1pt/1pt]
\textit{2021}                                                                                         & VPLFace \cite{deng2021variational}                                                                                   & 93.427                                        & \textit{96.076}                            & 94.666                                                          & 96.448                                     & 57.913                                                 & 62.446                            & 94.937                                      & 95.978           \\ 
\hdashline[1pt/1pt]
\textit{2022}                                                                                         & SphereFace2 \cite{wen2021sphereface2}                                                                               & 92.162                                        & 95.170                                     & 93.610                                                          & 95.743                                     & 57.994                                                 & 63.251                            & 92.595                                      & 94.224           \\ 
\hdashline[1pt/1pt]
2022                                                                                                  & Elastic-Arc \cite{boutros2022elasticface}                                                                               & 92.911                                        & 95.813                                     & 94.391                                                          & 96.203                                     & 57.779                                                 & 62.607                            & 95.674                                      & 96.429           \\ 
\hdashline[1pt/1pt]
2022                                                                                                  & Elastic-Cos \cite{boutros2022elasticface}                                                                               & 93.262                                        & 96.008                                     & 94.473                                                          & 96.315                                     & 60.086                                                 & 64.565                            & 94.989                                      & 95.602           \\ 
\hdashline[1pt/1pt]
\textit{2022}                                                                                         & AdaFace \cite{kim2022adaface}                                                                                   & 93.525                                        & 96.018                                     & 94.814                                                          & \textit{96.524}                            & \textit{60.837}                                        & 64.995                            & 95.888                                      & \uline{96.694}   \\ 
\hdashline[1pt/1pt]
\textit{2023}                                                                                         & QAFace \cite{saadabadi2023quality}                                                                                    & 92.687                                        & 95.764                                     & 94.018                                                          & 96.320                                     & 57.967                                                 & 62.983                            & 93.510                                      & 94.738           \\ 
\hdashline[1pt/1pt]
\textit{2023}                                                                                         & QMagFace \cite{terhorst2023qmagface}                                                                                  & 91.870                                        & 95.560                                     & 93.707                                                          & 96.213                                     & 57.833                                                 & 62.527                            & -                                           & -                \\ 
\hdashline[1pt/1pt]
\textit{2023}                                                                                         & UniFace \cite{zhou2023uniface}                                                                                   & 93.350                                        & 95.979                                     & \textit{94.886}                                                 & 96.402                                     & 60.327                                                 & 64.887                            & \uline{95.961}                              & 96.350           \\ 
\hdashline[1pt/1pt]
\textit{2023}                                                                                         & UniTSFace \cite{jia2023unitsface}                                                                                 & 93.155                                        & 95.891                                     & 94.442                                                          & 96.366                                     & 60.408                                                 & \textit{65.236}                   & 94.409                                      & 95.737           \\ 
\hdashline[1pt/1pt]
\textit{2024}                                                                                         & TopoFR \cite{dan2024topofr}                                                                                    & 93.019                                        & 95.813                                     & 94.243                                                          & 96.289                                     & 59.871                                                 & 64.029                            & \textit{95.913}                             & \textit{96.636}  \\ 
\hline\hline
\rowcolor[rgb]{0.765,0.918,0.773} \textit{Now}                                                        & \textbf{\textit{QCFace-Arc}}                                                               & \uline{94.761}                                & \uline{96.884}                             & \uline{96.157}                                                  & \uline{97.468}                             & \uline{62.634}                                         & \textbf{66.738}                   & \textbf{98.347}                             & \textbf{98.500}  \\ 
\hdashline[1pt/1pt]
\rowcolor[rgb]{0.765,0.918,0.773} \textit{Now}                                                        & \textbf{\textit{QQCFace-Arc}}                                                              & \textbf{94.898}                               & \textbf{96.923}                            & \textbf{96.172}                                                 & \textbf{97.484}                            & \textbf{64.941}                                        & \uline{68.160}                    & -                                           & -                \\
\hline
\end{tabular}
\end{adjustbox}
\end{table*}

\begin{table*}[!t]
\caption{Comparison in identification performance on IJB, TinyFace and MegaFace datasets with IResNet18. In the MegaFace benchmark, \textbf{\textit{Iden}} expresses \textbf{\textit{Rank-1}} in identification and \textbf{\textit{Veri}} is TAR at FAR=$10^{-6}$. The similarity score calculation of \cite{terhorst2023qmagface} is not supported by MegaFace devkit, expressed by the notation ``-".}
\label{tab:supp_comparison_iden_IJB_IR18}
\begin{adjustbox}{width=1.5\columnwidth,center}
\begin{tabular}{c|l|c;{1pt/1pt}c|c;{1pt/1pt}c|c;{1pt/1pt}c|c;{1pt/1pt}c} 
\hline
\rowcolor[rgb]{0.784,0.784,0.784} {\cellcolor[rgb]{0.784,0.784,0.784}}                                & \multicolumn{1}{c|}{{\cellcolor[rgb]{0.784,0.784,0.784}}}                                  & \multicolumn{2}{c|}{\textbf{IJB-B~}}                                                       & \multicolumn{2}{c|}{\textbf{IJB-C~}}                                                                         & \multicolumn{2}{c|}{\textbf{TinyFace}}                                                    & \multicolumn{2}{c}{\textbf{MegaFace}}                          \\ 
\cline{3-10}
\rowcolor[rgb]{0.784,0.784,0.784} \multirow{-2}{*}{{\cellcolor[rgb]{0.784,0.784,0.784}}\textbf{Year}} & \multicolumn{1}{c|}{\multirow{-2}{*}{{\cellcolor[rgb]{0.784,0.784,0.784}}\textbf{Method}}} & \multicolumn{1}{c|}{\textbf{\textit{Rank-1}}} & \textbf{\textbf{\textit{Rank-}}\textit{5}} & \multicolumn{1}{c|}{\textbf{\textbf{\textit{Rank-}}\textit{1}}} & \textbf{\textbf{\textit{Rank-}}\textit{5}} & \multicolumn{1}{c|}{\textbf{\textbf{\textit{Rank-1}}}} & \textbf{\textbf{\textit{Rank-5}}} & \multicolumn{1}{c|}{\textbf{\textit{Iden}}} & \textbf{Veri}    \\ 
\hline\hline
\textit{2015}                                                                                         & FaceNet \cite{schroff2015facenet}                                                                                   & 20.643                                        & 36.943                                     & 18.057                                                          & 32.813                                     & 17.543                                                 & 25.161                            & \textit{80.015}                             & \textit{84.670}  \\ 
\hdashline[1pt/1pt]
\textit{2017}                                                                                         & SphereFace \cite{liu2017sphereface}                                                                                & 52.561                                        & 69.961                                     & 50.094                                                          & 67.187                                     & 15.397                                                 & 23.095                            & 18.430                                      & 20.975           \\ 
\hdashline[1pt/1pt]
\textit{2018}                                                                                         & CosFace \cite{wang2018cosface}                                                                                   & 69.903                                        & 75.385                                     & 70.944                                                          & 75.746                                     & 44.769                                                 & 51.180                            & 71.948                                      & 76.819           \\ 
\hdashline[1pt/1pt]
\textit{2019}                                                                                         & ArcFace \cite{deng2019arcface}                                                                                   & 69.903                                        & 75.959                                     & 71.015                                                          & 76.185                                     & 41.550                                                 & 48.471                            & 71.215                                      & 75.920           \\ 
\hdashline[1pt/1pt]
\textit{2020}                                                                                         & MV-Arc-Softmax \cite{wang2020mis}                                                                            & 67.887                                        & 75.706                                     & 68.392                                                          & 75.170                                     & 48.095                                                 & 54.533                            & 54.523                                      & 57.891           \\ 
\hdashline[1pt/1pt]
\textit{2020}                                                                                         & CurricularFace \cite{huang2020curricularface}                                                                            & 70.156                                        & 75.871                                     & 71.464                                                          & 76.446                                     & 45.681                                                 & 52.173                            & 72.807                                      & 74.928           \\ 
\hdashline[1pt/1pt]
\textit{2021}                                                                                         & MagFace \cite{meng2021magface}                                                                                   & 71.782                                        & 77.897                                     & 72.659                                                          & 78.064                                     & 44.260                                                 & 50.402                            & 71.616                                      & 74.771           \\ 
\hdashline[1pt/1pt]
\textit{2021}                                                                                         & VPLFace \cite{deng2021variational}                                                                                   & \textit{83.846}                               & \textit{90.224}                            & \textit{85.638}                                                 & \textit{90.726}                            & 42.623                                                 & 49.329                            & \textit{79.485}                             & 83.521           \\ 
\hdashline[1pt/1pt]
\textit{2022}                                                                                         & SphereFace2 \cite{wen2021sphereface2}                                                                               & \uline{85.268}                                & \uline{90.691}                             & \uline{86.077}                                                  & \uline{91.027}                             & 43.750                                                 & 49.893                            & \textbf{80.574}                             & \textbf{84.818}  \\ 
\hdashline[1pt/1pt]
2022                                                                                                  & Elastic-Arc \cite{boutros2022elasticface}                                                                               & 82.210                                        & 88.802                                     & 83.050                                                          & 88.950                                     & 41.229                                                 & 47.720                            & 74.005                                      & 79.325           \\ 
\hdashline[1pt/1pt]
2022                                                                                                  & Elastic-Cos \cite{boutros2022elasticface}                                                                               & 82.960                                        & 89.026                                     & 84.163                                                          & 89.501                                     & 46.057                                                 & 52.387                            & 73.298                                      & 77.391           \\ 
\hdashline[1pt/1pt]
\textit{2022}                                                                                         & AdaFace \cite{kim2022adaface}                                                                                   & 73.087                                        & 78.793                                     & 73.853                                                          & 78.732                                     & 47.639                                                 & 54.211                            & 70.251                                      & 74.663           \\ 
\hdashline[1pt/1pt]
\textit{2023}                                                                                         & QAFace \cite{saadabadi2023quality}                                                                                    & 0.341                                         & 0.818                                      & 0.311                                                           & 0.658                                      & 3.782                                                  & 5.392                             & 0.069                                       & 0.022            \\ 
\hdashline[1pt/1pt]
\textit{2023}                                                                                         & QMagFace \cite{terhorst2023qmagface}                                                                                  & 66.095                                        & 71.694                                     & 65.467                                                          & 70.745                                     & 44.260                                                 & 50.402                            & -                                           & -                \\ 
\hdashline[1pt/1pt]
\textit{2023}                                                                                         & UniFace \cite{zhou2023uniface}                                                                                   & 78.296                                        & 85.209                                     & 78.850                                                          & 85.056                                     & \textbf{49.249}                                        & \textbf{56.330}                   & 70.749                                      & 74.664           \\ 
\hdashline[1pt/1pt]
\textit{2023}                                                                                         & UniTSFace \cite{jia2023unitsface}                                                                                 & \textbf{85.287}                               & \textbf{90.740}                            & \textbf{86.480}                                                 & \textbf{91.089}                            & \uline{48.686}                                         & \uline{55.660}                    & 79.008                                      & 82.475           \\ 
\hdashline[1pt/1pt]
\textit{2024}                                                                                         & TopoFR \cite{dan2024topofr}                                                                                    & 74.839                                        & 81.831                                     & 75.047                                                          & 81.514                                     & 47.452                                                 & 53.943                            & 65.542                                      & 0.048            \\ 
\hline\hline
\rowcolor[rgb]{0.765,0.918,0.773} \textit{Now}                                                        & \textbf{\textit{QCFace-Arc}}                                                               & 80.253                                        & 85.910                                     & 81.305                                                          & 86.021                                     & \textit{48.578}                                        & \textit{55.123}                   & \uline{80.017}                              & \uline{84.681}   \\ 
\hdashline[1pt/1pt]
\rowcolor[rgb]{0.765,0.918,0.773} \textit{Now}                                                        & \textbf{\textit{QQCFace-Arc}}                                                              & 79.231                                        & 84.732                                     & 79.824                                                          & 84.556                                     & \textit{48.578}                                        & \textit{55.123}                   & -                                           & -                \\
\hline
\end{tabular}
\end{adjustbox}
\end{table*}

\end{document}